\documentclass[pdflatex,sn-mathphys-ay]{sn-jnl}

\usepackage{graphicx}
\usepackage{multirow}
\usepackage{amsmath,amssymb,amsfonts}
\usepackage{amsthm}
\usepackage[cal=rsfs,scr=rsfs]{mathalfa}
\usepackage[title]{appendix}
\usepackage{xcolor}
\usepackage{textcomp}
\usepackage{manyfoot}
\usepackage{booktabs}
\usepackage{algorithm}
\usepackage{algorithmicx}
\usepackage{algpseudocode}
\usepackage{listings}
\usepackage{multicol}

\usepackage[utf8]{inputenc}
\usepackage[T1]{fontenc}
\usepackage{pifont}

\usepackage{rotating} 
\usepackage{colortbl}
\usepackage{hyperref}
\hypersetup{bookmarksdepth=2}

\usepackage{footnote}
\usepackage{pifont}
\usepackage[caption=false]{subfig}
\usepackage{tikz}
\usepackage{wrapfig}
\usepackage{lmodern}
\usetikzlibrary{arrows.meta, positioning, calc, shapes.symbols, shapes.callouts}
\makesavenoteenv{tabular}
\usepackage{float}
\usepackage{setspace}
\usepackage{placeins}
\usepackage{pdflscape}
\usepackage{svg}

\theoremstyle{thmstyleone}

\theoremstyle{thmstyletwo}

\theoremstyle{thmstylethree}

\newcommand{\pharrulesize}{\footnotesize}

\newcommand{\pharrule}[3]{
  {
  \pharrulesize
  \[
  \setlength{\arraycolsep}{2pt}
  \begin{aligned}
  #1:\;
  \left\{
  \begin{aligned}
  #2
  \end{aligned}
  \right.
  \\[0.2em]
  \hspace{0.25em}\Rightarrow\;
  \begin{aligned}
  #3
  \end{aligned}
  \end{aligned}
  \]
  }
}

\usepackage{silence}
\begin{document}

\title[Article Title]{Explaining Time Series Classifiers with PHAR: Rule Extraction and Fusion from Post-hoc Attributions}

\author*[1,2]{\fnm{Maciej} \sur{Mozolewski}}\email{m.mozolewski@uj.edu.pl}

\author[2]{\fnm{Szymon} \sur{Bobek}}\email{szymon.bobek@uj.edu.pl}

\author[2]{\fnm{Grzegorz J.} \sur{Nalepa}}\email{grzegorz.j.nalepa@uj.edu.pl}

\affil[1]{
      \orgdiv{Jagiellonian Human-Centered AI Lab, Mark Kac Center for Complex Systems Research},
      \orgname{Jagiellonian University},
      \orgaddress{\street{Łojasiewicza 11}, \city{Krakow}, \postcode{30-348}, \country{Poland}}
    }

\affil[2]{\orgdiv{Department of Human-Centered Artificial Intelligence}, \orgname{Institute of Applied Computer Science, Jagiellonian University}, \orgaddress{\street{Łojasiewicza 11}, \city{Krakow}, \postcode{30-348}, \country{Poland}}}

\abstract{
Explaining machine learning (ML) models for time series (\emph{TS}) classification remains challenging due to the difficulty of interpreting raw time series and the high dimensionality of the input space. 
We introduce \emph{PHAR}---\emph{Post-hoc Attribution Rules}---a unified framework that transforms numeric feature attributions from post-hoc, instance-wise explainers (e.g. \emph{LIME}, \emph{SHAP}) into structured, human-readable rules. 
These rules define human-readable intervals that indicate where and when decision-relevant segments occur and can enhance model transparency by localizing threshold-based conditions on the raw series. 
\emph{PHAR} performs comparably to native rule-based methods, such as \emph{Anchor}, while scaling more efficiently to long \emph{TS} sequences and achieving broader instance coverage. 
A dedicated \emph{rule fusion} step consolidates rule sets using strategies like \emph{weighted selection} and \emph{lasso-based refinement}, balancing key quality metrics: \emph{coverage}, \emph{confidence}, and \emph{simplicity}. 
This fusion ensures each instance receives a concise and unambiguous rule, improving both explanation fidelity and consistency. 
We further introduce visualization techniques to illustrate specificity-generalization trade-offs in the derived rules. 
\emph{PHAR} resolves conflicting and overlapping explanations---a common effect of the \emph{Rashomon phenomenon}---into coherent, domain-adaptable insights. 
Comprehensive experiments on \emph{UCR/UEA Time Series Classification Archive} demonstrate that \emph{PHAR} may improve interpretability, decision transparency, and practical applicability for \emph{TS} classification tasks by providing concise, human-readable rules aligned with model predictions. 
}


\maketitle

\section{Introduction}\label{sec:introduction}
Explaining time series classifiers presents unique challenges compared to models for images or tabular data. 
Time series often lack obvious semantic features. 
Without domain expertise, even simple attribution outputs can be difficult to interpret, which complicates the use of standard \emph{Explainable AI (XAI)} tools~\citep{theissler2022explainable}. 
Temporal dependencies, phase shifts, and high dimensionality further worsen the explanations. 
When non-specialized methods are applied, results can become unstable or overly generic~\citep{gu2024explainable}. 
These issues reveal a key gap in current XAI research. 
Unlike non-temporal models built on intuitive, semantically rich features, time series classifiers demand explanations that both capture temporal dynamics and deliver meaningful, context-aware insights tailored to sequential data. 
In addition, \emph{Deep Learning (DL)} is widely used for (\emph{Time Series (TS)}) classification, particularly in medicine and industry, because of its ability to model complex temporal patterns for tasks such because disease diagnosis and predictive maintenance. 
Despite strong predictive performance, these models remain difficult to interpret in practice.

To address the challenges of XAI for state-of-the-art models such as {DL}, common approaches include numeric attribution methods such as \emph{permutation importance}~\citep{altmann2010permutation}, \emph{SHAP}~\citep{lundberg2017unifiedapproachinterpretingmodel} and \emph{LIME}~\citep{ribeiro2016why}, as well as gradient-based techniques such as \emph{saliency maps}~\citep{simonyan2013deep} and \emph{Layerwise Relevance Propagation (LRP)}~\citep{bach2015pixel}, all of which quantify feature contributions and trace relevance through the model’s computations.
These approaches facilitate feature ranking, sensitivity analysis, and model debugging by producing developer-oriented outputs like importance scores or heatmaps. 
However, studies show that such numeric explanations often highlight broad regions of input without providing clear, context-specific insights~\citep{arsenault2024surveyexplainableartificialintelligence,jin2021deep}. 
In sequential data, this tendency gives rise to \emph{"ambiguous rules"}, where multiple time steps or variables appear equally influential and cannot be easily distinguished. 
Consequently, these explanations are difficult for non-experts to interpret \citep{jeyakumar2020empirical,rojat2021explainable}, and the challenge is amplified in \emph{TS} classifiers, where temporal dependencies and the high dimensionality of multivariate sequences further obscure meaningful patterns \citep{schlegel2019towards,theissler2022explainable}. 

Anchor rule-based explanations have demonstrated improved interpretability by extracting succinct \emph{"IF–THEN"} conditions that align closely with model behavior~\citep{Ribeiro_Singh_Guestrin_2018}. 
However, such methods are seldom applied to {TS} data. 
Furthermore, rule-based methods become computationally prohibitive as the sequence length increases~\citep{arshad2024investor}. 
An exhaustive  search for stable \emph{"IF--THEN"} conditions could be computationally infeasible, leading to a combinatorial explosion of candidate rules and temporal instability. 
Beyond computational infeasibility, \emph{Cognitive Load Theory}, summarized by~\cite{Skulmowski2022}, emphasizes reducing extraneous demands on human working memory. 
Analogously, simpler and unambiguous decision rules referencing a limited number of time steps minimize cognitive effort for experts, 
supporting interpretability and visualization 
by exposing time-localized, threshold-based conditions that can be inspected on the original signal. 

Our motivation stems from three persistent gaps in explaining deep time series classifiers. 
First, both numeric attributions and native rule-based methods remain difficult to interpret on high-dimensional time series, especially when many time steps and channels contribute simultaneously. 
Second, most studies assess each explainer in isolation, without fusing their outputs or systematically comparing rule quality in terms of coverage, confidence, and simplicity. 
Third, existing work rarely overlays explanations directly on time series plots, missing an opportunity to show where and when decision-relevant segments occur.

PHAR addresses these gaps by transforming instance-level numeric attributions into compact, human-readable rules for time series classification. 
Rather than relying on global XAI approaches based on clustering or surrogate models~\citep{mekonnen_2024_parameterized_event_primitives}, we build on local post-hoc attribution methods such as SHAP~\citep{lundberg2017unifiedapproachinterpretingmodel} and LIME~\citep{ribeiro2016why}. 
PHAR preserves their main strengths -- precise feature ranking, post-hoc applicability, and local fidelity---while adapting them to sequential data through explicit temporal conditions. 
Using semi-factual reasoning, we identify minimal temporal segments in which perturbing feature values does not change the model prediction and turn these stable segments into ``IF--THEN'' rules. 
We adopt Anchor~\citep{Ribeiro_Singh_Guestrin_2018} as a baseline based on native rules and systematically compare it to rules derived from numeric attributions.

Beyond single explainers, PHAR includes a dedicated fusion step that consolidates rules from LIME, SHAP, Anchor, and potentially expert-defined rules. 
Fusion methods use coverage, confidence, and rule length to select a single optimized rule per instance, reducing ambiguity caused by overlapping or conflicting explanations. 
Finally, PHAR visualizes the resulting rules as value intervals overlaid on the original time series, allowing domain experts to inspect temporal decision boundaries directly on the signal. 
Although we focus on time series, the same numeric-to-rule transformation and fusion principles can be extended to other structured data, such as tabular datasets.

\begin{figure*}[h]
    \centering
    \resizebox{0.95\textwidth}{!}{
        \includegraphics{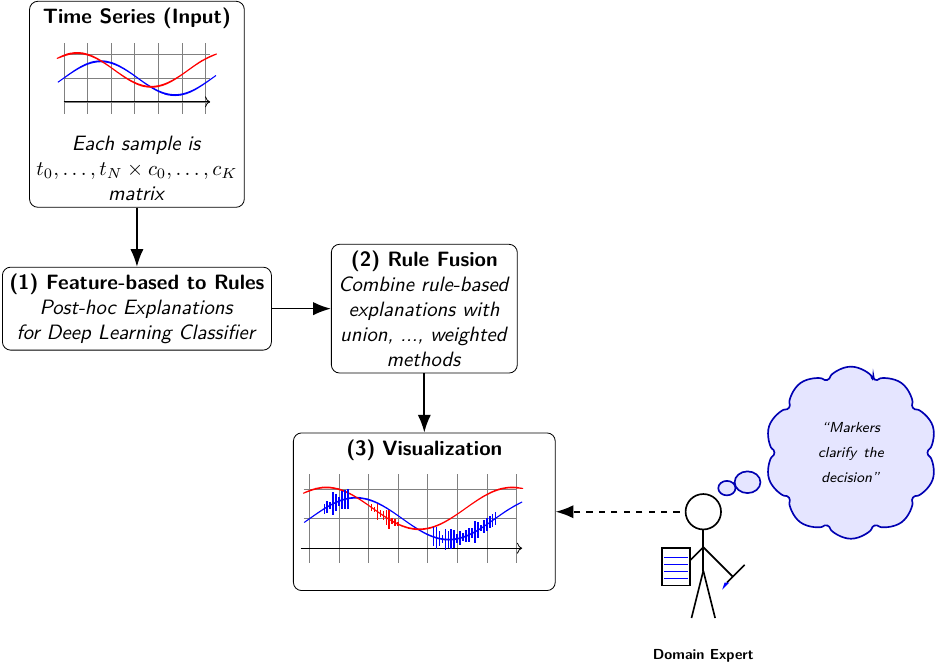}
    }
    \caption{Rules derived from post-hoc model explanations like SHAP.}
    \label{fig:pipeline}
\end{figure*}

\emph{PHAR} aims to improve trust and transparency in \emph{TS} classification with DL through three key contributions:  
(1) a pipeline that converts numeric post-hoc attributions into expert-level rule representations for \emph{TS};  
(2) a \emph{fusion mechanism} that unifies rules from different explainers for better coherence and quality;  
(3) visualization of rules, as outlined in Figure~\ref{fig:pipeline}.
Each input time series is represented as a \(T \times C\) matrix—\(T\) time steps \(t_{1},\dots,t_{T}\) by \(C\) channels \(c_{1},\dots,c_{C}\)—and is processed by a \emph{DL classifier}. 
Post-hoc explainers such as \emph{SHAP} and \emph{LIME} compute numerical attributions. 
In the first stage, these outputs are converted into interval rules for each instance. 
Rule-based explanations, such as \emph{Anchor} , can be used directly. 
In the second stage, rule sets derived from different explainers are fused using methods such as \emph{union}, \emph{weighted selection}, and \emph{lasso-based} refinement to yield a single coherent rule per instance. 
Finally, each rule is visualized as vertical interval markers overlaid on the original time-series plot, allowing domain experts to inspect and validate the model’s decision logic. 
Crucially, the visualization conveys information beyond the textual rule: it shows rule–signal alignment in time, overlap across channels and explainers,
which supports faster error checking and expert validation. 

The remainder of this paper is structured as follows: 
Section~\ref{sec:related} reviews related work and explains our motivation.
Section~\ref{sec:method} details our framework, covering the transformation from numeric-to-rule (Section~\ref{sec:numeric_to_rules}) and the \emph{fusion explanations} (Section~\ref{sec:rule_ensemble}).  
Section~\ref{sec:evaluation} evaluates coverage, confidence, and other metrics.  
Section~\ref{sec:summary} summarizes findings and future directions.  
The Appendix~\ref{sec:appendix} 
provides supplementary tables, figures, and extended results, 
while "Supplementary Materials" (Section~\ref{sec:supplement})  
contains statical results for Critical Difference Diagrams.  
\FloatBarrier

\section{Related work} \label{sec:related}
In this section, we review the state of the art in time series explainability that informs our \emph{PHAR} framework. 
Existing methods follow two main lines. 
First, surrogate models, such as decision trees, approximate black-box predictors, and induce rule-based intervals over time. 
Second, post-hoc attribution methods such as LIME~\citep{ribeiro2016why}, SHAP~\citep{lundberg2017unifiedapproachinterpretingmodel}, and saliency maps provide numeric importance scores but often fail to enforce temporal contiguity or capture long-range dependencies in time series. 
\emph{Dynamic time warping (DTW)} is a classical similarity measure that aligns sequences of different lengths or phases. 
DTW-based clustering has been used to identify prototypical patterns, offering structural insight into recurrent temporal shapes. 
In parallel, fuzzy rules, decision tree splits, and statistical impact rules have been proposed to improve the interpretability of sequential models, typically trading off the complexity of the rules, local fidelity, and temporal coherence. 
However, most of these approaches study each explainer in isolation and emphasize either local or global views, with limited support for integrating multiple explainers into a coherent description of model behavior. 
In what follows, we first examine rule-based explanation methods tailored to sequential data and then survey strategies for fusing outputs from multiple explainers. 

Explaining TS classifiers is challenging, especially for complex models.  
~\cite{mekonnen_2024_parameterized_event_primitives} proposed a post-hoc method using \emph{Parameterized Event Primitives (PEPs)} to explain TS models.  
Their approach detects key events (trends, extrema), clusters them, and derives rules via decision trees.  
These rules capture sequential patterns, 
supporting fidelity to model predictions and human interpretability, while acknowledging that such rules are proxies of model behavior rather than causal explanations. 
This method extracts meaningful patterns while maintaining temporal dependencies
and lays the groundwork for hybrid explanations that address temporal dependencies.  

~\cite{SpinnatoFrancesco2023UATS} propose a TS explanation framework combining saliency maps, example-based examples, and rule-based methods.  
Saliency maps highlight key regions, instance-based methods provide prototypes and counterfactuals, and rule-based methods define logical conditions on subsequences.  
Integrating these methods bridges local and global explanations.  
This hybrid approach can improve the interpretability and temporal coherence of the resulting explanations by constraining rules to compact time intervals and simple inequalities on the raw series. 

Decision rules improve interpretability in \emph{TS} forecasting, especially in finance. 
~\cite{arsenault2024surveyexplainableartificialintelligence} review decision trees, fuzzy trees, and \emph{Mamdani-type rules}.  
~\cite{silva2021c45fuzzy} introduced \emph{FDT-FTS}, which combines fuzzy logic and decision trees for multivariate \emph{TS} forecasting.  
\emph{Mamdani-type rules} use fuzzy logic and linguistic variables for human-readable explanations.  
~\cite{xie2021neuralfuzzy} integrated them into the \emph{Neural Fuzzy Hammerstein-Wiener network}.  
These methods balance predictive accuracy with human-readable simplicity of rules (few conditions, short intervals), supporting \emph{TS} analysis. 
Next, we discuss the decision rules for progressive decision-making.  

The Belief–Rule–Based (BRB) framework~\citep{NimmySoniaFarhana2023} overcomes key weaknesses of post-hoc methods such as SHAP~\citep{lundberg2017unifiedapproachinterpretingmodel}, LIME~\citep{ribeiro2016why}, and LINDA-BN~\citep{Moreira2021LINDA} by structuring explanations into human-readable rules. 
In the Knowledge Graph Module (KGM), expert input is formalized into a causal graph that captures both temporal order and inter-feature dependencies, addressing SHAP’s and LIME’s tendency to treat each timestep or feature independently and thus lose important sequence information. 
The Knowledge Propagation Module (KPM) then propagates belief distributions through this graph to extrapolate feature trajectories, inherently managing missing data and uncertainty without relying on expensive local sampling that can render SHAP and LIME at times unstable. 
Finally, the Feature Evaluation Module (FEM) synthesizes these propagated beliefs into concise \emph{"IF–THEN"} rules that highlight the most influential intervals, removing the need to interpret raw attribution weights or fuzzy inference states as in LINDA-BN and enabling straightforward expert validation.

LIMREF~\citep{RajapakshaDilini2022LLIM} provides local model-agnostic explanations for global TS forecasting models by selecting
 similar TS via DTW and generating synthetic instances via bootstrapping.  
Predictions guide the derivation of the impact rule using~\citep{Webb2001}.  
Rules fall into four types, including counterfactual and supportive, which aid in forecast adjustments.  
By focusing on local TS neighborhoods, LIMREF ensures that rules represent global model behavior.  
Its focus on local fidelity complements global and progressive methods.  

~\cite{Veerappa_2021_Maritime} apply the interpretation of the association rule to the classification of TS in maritime settings.  
They evaluate surrogate models for rule-based explanations of a neural network classifier.  
Each method generates decision rules, balancing fidelity and complexity.  
They analyze how support thresholds and discretization bins affect explanation quality.  
The authors highlight trade-offs between rule complexity, accuracy, and interpretability, showing the role of hyperparameter tuning in domain-specific rule optimization.  

\cite{RuleXAI_MACHA2022101209} introduced RuleXAI\footnote{\url{https://github.com/adaa-polsl/RuleXAI}}, an open-source rule-based explainability library for machine learning tasks. 
It supports classification, regression, and survival analysis by generating \emph{elementary rule conditions} — simple threshold-based splits on individual features. 
RuleXAI provides both global and local explanations by directly learning interpretable models from data, without relying on black-box feature attribution methods. 
It transforms input data into binary feature conditions, reducing preprocessing complexity and ensuring high transparency. 
Although not designed for time series data, RuleXAI’s symbolic rule extraction aligns with decision table logic and rule-based decision systems, facilitating straightforward human interpretability. 

~\cite{hryniewskaguzik2024normensemblexaiunveilingstrengthsweaknesses} explore the combination of XAI methods based on numeric characteristics in an ensemble framework, although not for TS.  
They propose \emph{NormEnsembleXAI}, which normalizes and aggregates explanations (e.g. feature-importance, local attributions) for consistent interpretability across models.  
Integrating LIME, SHAP, and Integrated Gradients, they show that uniform normalization reduces attribution variability and reveals coherent explanation patterns across models.

TSProto by~\cite{sbk2025tsproto} offers a complementary approach to explaining time series classifiers by converting feature attributions (e.g. SHAP or LIME) into high-level prototypes, which are then integrated into interpretable decision trees. However, it relies on a single XAI method to identify important regions and builds rules using a separate model based on SHAP-extracted features. As a result, it remains susceptible to the Rashomon effect, since in TSProto SHAP only indicates potentially relevant regions, not how the model actually uses them in decision-making.

\section{Method: Post-hoc Attribution Rules for time series} \label{sec:method}

\subsection{Overview of the \emph{PHAR} Framework}
The \emph{PHAR (Post-hoc Attribution Rules for time series)} framework addresses the demand for semantically meaningful, stable, and cognitively efficient explanations in time series classification. 
In \emph{PHAR}, we transform \emph{post-hoc}, \emph{instance-level numeric feature attributions} into concise, human-readable \emph{rule-based explanations} and integrate outputs from multiple explainers enabling assessment of the quality of explanations and their visualization. 
First, \emph{PHAR} selects the most salient temporal intervals---and, in multivariate settings, the most informative channels---so that rules focus on key patterns. 
Next, these intervals are translated into interpretable \emph{“IF–THEN”} rules that preserve the model’s predicted class. 
An optional \emph{fusion step} then merges rules obtained from numeric attributions, Anchor explanations~\citep{Ribeiro_Singh_Guestrin_2018}, and potentially expert rules, optimizing for \emph{coverage}, \emph{confidence}, and \emph{simplicity}. 
\emph{Fusing} explanations via techniques such as \emph{intersection}, \emph{union}, and \emph{weighted aggregation} ensures faithfulness to DL classifier predictions and usability for end-users.   
Finally, to reduce extraneous cognitive load, \emph{PHAR} visualizes each rule with \emph{semi-factual} plots that contrast actual and hypothetical behaviors, highlighting how small changes would affect the model’s decision. 
Our framework is structured around 4 Research Questions that guide the transformation, fusion, and visualization of rule-based explanations for time series classification.

\texttt{RQ1:} How can numeric attributions from \emph{LIME} or \emph{SHAP} be transformed into precise, interpretable \emph{rule-based explanations} for \emph{TS}? 

\texttt{RQ2:} 
How can metrics like \emph{coverage}, \emph{confidence}, \emph{feature count per rule}, and \emph{explained instance percentage} 
help select a single, unambiguous rule representation and reveal explanation strengths and weaknesses? 

\texttt{RQ3:} How can \texttt{fusion methods}, combining rules from \emph{LIME}, \emph{SHAP}, and \emph{Anchor} or from a single explainer, produce an optimized rule per instance that improves \emph{coverage}, \emph{confidence}, and \emph{explanation ratio} while controlling rule complexity? 

\texttt{RQ4:} How can visualizations support the interpretability of rule-based explanations in \emph{TS} models?  
Rules may span multiple time steps and features (in \emph{multivariate TS}). 
We explore whether overlaying rule-based explanations on \emph{TS} plots helps to illustrate their structure. 

\begin{figure*}[h]
    \centering
    \resizebox{0.60\textwidth}{!}{
        \includegraphics{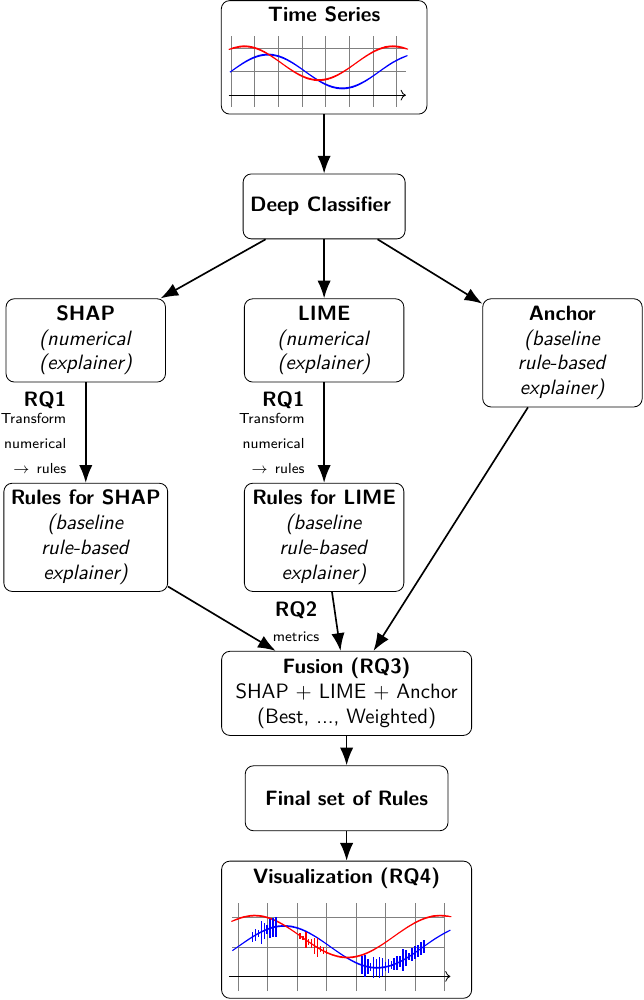}
    }
    \caption{Overview of the PHAR framework and its main stages (RQ1–RQ4). Numeric explainers (e.g. LIME, SHAP) and rule-based methods (e.g. Anchor) are shown as illustrative modules that can be replaced by other compatible explainers.}
    \label{fig:pipeline-detailed}
\end{figure*}

\singlespacing \noindent Figure~\ref{fig:pipeline-detailed} presents \emph{PHAR} as a modular framework aligned with our Research Questions. 
A deep classifier first assigns each input time series, univariate or multivariate, to a predicted class. 
Local post-hoc explainers, such as \emph{SHAP} and \emph{LIME}, produce numeric attributions that we convert into interval rules of the form
\(
r: \; x_c^{(t)} \in (l_c^{(t)}, h_c^{(t)}] \;\Rightarrow\; \hat y,\quad 
\text{conf}(r)=\gamma,\;\text{cov}(r)=\kappa,
\) 
where $x_c^{(t)}$ denotes channel $c$ (for multivariate \emph{TS}) at time step $t$, $(l_c^{(t)},h_c^{(t)}]$ is the selected interval and $\gamma,\kappa$ are the confidence and coverage of the rule. 
This transformation from numeric-to-rule addresses \textbf{RQ1} (Section~\ref{sec:numeric_to_rules}).  

In parallel, a native rule-based explanation, such as \emph{Anchor}, generates rules directly from the classifier. 
All candidate rules, regardless of origin, are then evaluated using metrics such as \emph{coverage}, \emph{confidence}, and \emph{feature count}, which addresses \textbf{RQ2} (Section~\ref{sec:new_rule_conf_cov}). 

Next, \emph{PHAR} applies multiple fusion strategies (\texttt{Intersection}, \texttt{Union}, \texttt{Weighted}, \texttt{Lasso}, \texttt{Lasso global}, and \texttt{Best}) to these rule sets, using the same metrics to select a single optimized rule per instance (\textbf{RQ3}, Section~\ref{sec:rule_ensemble}).  

Finally, to support \textbf{RQ4}, the resulting rules are visualized as vertical intervals overlayed on the original time series, highlighting which channels and timesteps support the prediction (Section~\ref{sec:qual_eval} and Appendix~\ref{app:visual_appendix}). 

Although Figure~\ref{fig:pipeline-detailed} instantiates the framework with \emph{SHAP}, \emph{LIME}, and \emph{Anchor}, the same pipeline can integrate other numeric or rule-based explanations as configurable elements. 
\FloatBarrier

\subsection{Rule-based explanations} \label{sec:logical_rules}
Converting scores from explainers that produce raw \emph{post hoc numerical attributions}, such as \emph{SHAP} and \emph{LIME}, into a structured rule format makes it possible to compare and merge rules from different explainers, resolve conflicts when features are correlated, and engage \emph{Domain Experts} in validating and refining the symbolic rules. 
Our approach employs \emph{rule fusion} to aggregate rule sets produced by \emph{post hoc explainers} and additionally allows us to incorporate \emph{native rule-based explainers} like \emph{Anchor} or even rules developed directly by \emph{Domain Experts}. 
Inspired by ensemble methods \citep{dietterich2000ensemble}, we call this process \texttt{fusion} because it goes beyond classical bagging or boosting. 
\emph{Fusion} refines and unifies rules by weighting them according to \emph{coverage}, \emph{precision}, and \emph{simplicity}, and by merging redundant or overlapping conditions. 
The resulting \emph{fused rules} conform to the same \emph{expert system format}, allowing them to be easily evaluated, adjusted, visualized, and aligned with expert knowledge to support robust domain reasoning. 

In \citet{Davis1977}, rules are expressed in the form \emph{"IF–THEN"} augmented with a certainty factor. 
The premise is a conjunction of predicate evaluations \(p_j(o_j,a_j)=v_j\) for \(j=1,\dots,n\), and the conclusion \(q(o',a')=v'_{\mathrm{out}}\) is drawn with a strength given by multiplying the aggregated premise confidence by \(\mathrm{CF}_i\):
\begin{equation}
R_i:\;\bigl(\bigwedge_{j=1}^n p_j(o_j,a_j)=v_j\bigr)
\;\xrightarrow{\times\,\mathrm{CF}_i}\;
q(o',a')=v'_{\mathrm{out}}
\end{equation}
Here, \(p_j\) denotes a predicate function applied to object \(o_j\) and attribute \(a_j\), \(v_j\) is its value, and the certainty factor \(\mathrm{CF}_i\) quantifies the reliability of the inference.
Building on this idea, we define our \emph{PHAR} rule format as in~\citep{ligeza_rule_based,Niederlinski2013}, where each rule \(R_i\) comprises a set of logical conditions \(W_1,\dots,W_n\), a conclusion \(W_{\mathrm{out}}\), and an inference semaphore \(S_i\) representing the combined confidence \(c_i\) obtained with the method described in Section~\ref{sec:new_rule_conf_cov}: 
\begin{equation}
R_i:\;\{W_1,W_2,\ldots,W_n\}
\;\xrightarrow{S_i}\;
W_{\mathrm{out}}
\end{equation}
In this schema, each \(W_j\) is a logical condition derived from feature attributions, \(W_{\mathrm{out}}\) is the predicted class or decision, and \(S_i = c_i\) signals the confidence of the rule. 

Our approach follows established rule classifications in \emph{knowledge-based systems}~\citep{Nalepa2018}. 
The \emph{PHAR} rules are best described as \emph{derivation} and \emph{classification rules}.  
\emph{Derivation rules} define logical conditions under which a conclusion is inferred from given premises, in our case \emph{TS} feature conditions leading to model predictions.  
\emph{Classification rules} assign instances to classes based on these conditions.  
From the perspective of logical inference, our rules are \emph{probabilistic deductive rules}.  
Deductive rules infer conclusions from premises. 
If the premises hold, the conclusion follows deductively.  
Given a set of features and its constraints, our rules classify instances under a probabilistic confidence threshold rather than absolute certainty.  
Unlike classical deduction, our rules incorporate \emph{confidence}, applying deterministically within probability constraints.  
Furthermore, our rules belong to \emph{attributive rule languages}~\citep{Nalepa2018}.  
These languages encode \emph{knowledge} through attributes and values, which are common in decision tables, rule-based systems, and databases.  
They enable intuitive representation and efficient manipulation.  
For example, the rule \emph{if \(\mathrm{temperature}(t=10)>50\) and \(\mathrm{pressure}(t=11)<10\), then class = failure [\(c=0.85\)]}, where \(t=11\) denotes the timestep index, provides a human-readable decision boundary that replaces raw numeric attribution scores. 

We illustrate a generic rule pattern for an arbitrary instance \(i\) as follows:
\begin{equation}
\label{eq:rule_pattern}
R_i:\quad
\left\{
\begin{aligned}
& t_{j_1}c_{j_1} \in (\ell_{j_1},u_{j_1}] \\
& t_{j_2}c_{j_2} \in (\ell_{j_2},u_{j_2}] \\
& \;\;\vdots \\
& t_{j_k}c_{j_k} \in (\ell_{j_k},u_{j_k}]
\end{aligned}
\right.
\;\Rightarrow\;
\begin{aligned}
\hat y &= "A", \\
\mathrm{CONF}(R_i) &= \gamma,\;\gamma \in [0,1],\\
\mathrm{COV}(R_i) &= \kappa,\;\kappa \in [0,1].
\end{aligned}
\end{equation}
Here, each pair \(t_{j_m}c_{j_m}\) denotes \emph{“channel”} \(c_{j_m}\) (in multivariate series) at the time step \(t_{j_m}\). 
The indices \(j_1,\dots,j_k\) are arbitrarily selected from the complete set of \(T\) time steps and \(C\) \emph{"channels"}, choosing the pairs \(k+1\) with the greatest impact on model predictions. 
Each interval \((\ell_{j_m},u_{j_m}]\) has the lower bound \(\ell_{j_m}\) and the upper bound \(u_{j_m}\). 
\(\mathrm{CONF}(R_i)\) is the proportion of instances that satisfy the rule that are correctly classified as class \("A"\), and \(\mathrm{COV}(R_i)\) is the fraction of all instances that satisfy the rule. 
The formal derivation of the rules and their \emph{CONF} and \emph{COV} measures is presented in Sections~\ref{sec:numeric_to_rules} and~\ref{sec:new_rule_conf_cov}, while Section~\ref{sec:qual_eval} provides real-world examples of the rules in this notation for a selected dataset.

\subsection{Conversion of numeric importance values to rules} \label{sec:numeric_to_rules}
We consider a TS dataset with \(N\) samples and a DL model that produces multi-class predictions. 
In the univariate case, each feature \(f\) is indexed only by timestep \(\mathrm{t}_i\), whereas in the multivariate case, the features are indexed by timestep and channel, for example \(\mathrm{t}_i\mathrm{c}_v\). 
Let \(\{X_n\}_{n=1}^N\) denote the input instances with model predictions \(y_n\). 
In this work, we explain the behavior of the trained model, so the predicted classes \(y_n\) act as a ground truth for the explanation stage. 

Feature-based explainers, such as \emph{LIME} and \emph{SHAP}, assign a numeric importance value to each feature of each instance. 
Let \(e_{n,f}\) denote the importance of the feature \(f\), for instance \(n\). 
To extract rules, we first identify a threshold \(T\) that marks \emph{important} features based on \(|e_{n,f}|\). 
In \emph{PHAR}, \(T\) is percentile-based and can be computed per feature (\emph{local} threshold) or across all features (\emph{global} threshold). 
Other thresholding strategies are also possible, for example, selecting the top-\(k\) features per instance, using a fixed absolute cutoff on \(|e_{n,f}|\), or defining \(T\) from distributional statistics such as \(\mu + \lambda \sigma\). 
However, in this study, we do not explore these alternatives.

We first compute absolute explanation values for all training instances and features. Let \( E \) be the multiset of absolute values:
\(
E = \{|e_{n,f}| : n=1,\dots,N,\; f=1,\dots,F \},
\)
where \( e_{n,f} \) is the explanation value for the feature \( f \) (e.g. \( \text{t}_{i}\text{c}_{v} \) for multivariate or \( \text{t}_{i} \) for univariate TS) at instance \( n \). 
We then define a global threshold \( T_{global} \) by selecting a percentile \( p \):
\(
T_{global} = \text{Percentile}(E, p).
\)
If we apply thresholds per feature, we instead compute a separate threshold \( T_{f} \) for each feature \( f \):
\(
E_{f} = \{|e_{n,f}| : n=1,\dots,N\},
\quad
T_{f} = \text{Percentile}(E_{f}, p).
\)
During inference in test instances, for each feature \( f \), let \( e_{n,f} \) be the explanation value for instance \( n \):
\begin{equation}
\begin{aligned}
|e_{n,f}| &\geq T_{global} \quad \text{(if using global thresholding; \(\delta = True\) )}, \\
\text{or} \quad |e_{n,f}| &\geq T_{f} \quad \text{(if using per-feature thresholding; \(\delta = False\)).}
\end{aligned}
\end{equation}

Features that exceed their threshold are \emph{important}.  
For \emph{important features}, we generate intervals that preserve the prediction of the model.  
Stable intervals \((l_{f}, u_{f}]\) are derived by controlled perturbation of each \emph{important feature} and sampling around \( x_{n,f} \).  
Let \(F_{*}\) be \emph{important features} for an instance, identified after thresholding.  
For each \( f \in F_{*}\), we take the original value \( x_{n,f} \) and generate perturbed samples:  
\(
x_{n,f}^{(m)} \sim \text{Uniform}(x_{n,f}-\Delta_{f},\; x_{n,f}+\Delta_{f}),
\)
where \(\Delta_{f}\) is a perturbation range based on training explanation statistics (std. deviation × constant \(\sigma_p\)).  
Joint sampling over \( f \in F_{*} \) captures nonlinear feature dependencies.  
For each perturbed sample \((x_{n,f}^{(m)})_{f \in F_{*}}\), we calculate the prediction of the model.  
Let \( c_{ref}(n) \) be the original predicted class of instance \( n \).  
If \( c_{ref}(n) = y_n \), we determine \((l_{f}, u_{f}]\) per feature \( f \) that preserves this class.  
Each \( f \in F_{*} \) receives a stable interval:
\(
f : x_{f} \in (l_{f},u_{f}],
\)
which defines a rule \( R_{n} \) for the instance \( n \):
\begin{equation}
R_{n}: \bigwedge_{f \in F_{*}} (x_{f} \in (l_{f},u_{f}]).
\end{equation}
This converts numeric importance into rule conditions with stable feature intervals.  

\subsubsection{Confidence and coverage of newly derived rules} \label{sec:new_rule_conf_cov}
After deriving the final rule \( R_{n} \) for the instance \( n \), we evaluate its explanatory quality on the test set. Consider the test data set \(\{X_{test,k}\}_{k=1}^{N_{test}}\), where \(N_{test}\) is the number of test instances. We identify the subset of test instances that satisfy the rule:
\(
S(R_{n}) = \{ X_{test,k} : X_{test,k} \text{ satisfies } R_{n}\}.
\)
To measure coverage, we compute the following: 
\begin{equation}
COV(R_{n}) = \frac{|S(R_{n})|}{N_{test}}.
\end{equation}
This ratio indicates how many test instances fall into the region defined by the rule's intervals. A higher \( COV(R_{n}) \) means that the rule covers a larger portion of the test set.

To measure confidence, we focus on the consistency of the predictions within \( S(R_{n}) \). Let:
\(
T(R_{n}) = \{ X_{test,k} \in S(R_{n}) : y_{model}(X_{test,k}) = c_{ref}(n)\},
\)
i.e., the subset of test instances covered for which the predicted class by the model matches \( c_{ref}(n) \). 
The confidence is then:
\begin{equation}
CONF(R_{n}) = \frac{|T(R_{n})|}{|S(R_{n})|}.
\end{equation}
If no test instance satisfies the rule (\( S(R_{n}) = \emptyset \)), then \( COV(R_{n}) = 0 \) and \( CONF(R_{n}) \) are undefined, which means that there is no explanatory power. 
However, at least the initial instance \( n \) should satisfy the rule.  
If the rule covers test instances but does not match \( c_{ref}(n) \), confidence decreases.  

This procedure ensures \emph{coverage} and \emph{confidence} reflect how well derived intervals define a decision region on test data.  
High \emph{coverage} means broad generalization; high \emph{confidence} ensures alignment with model predictions. 

\subsubsection{Hyperparameter optimization in rule transformation step} \label{sec:optuna}
To ensure that our rule‐-based explanations achieve a balance between fidelity and simplicity, we employ Optuna~\footnote{Optuna: \url{https://optuna.org/}}~\citep{optuna_2019} to tune key hyperparameters of this step of our framework. 
\paragraph{Optimized hyperparameters:}
\begin{itemize}
    \item \(p\) (\emph{threshold percentile}): Feature importance threshold in terms of percentile. 
    Higher values prioritize top-scoring features and thus result in a lower number of \emph{important features}. 
    \item \(\delta\) (\emph{global importance indicator}): Boolean for global or per-feature thresholding. If \emph{True}, a single threshold \( T_{global} \) is applied; otherwise, thresholds \( T_{f} \) are computed per feature. 
    \item \(\sigma_p\) (\emph{perturbation scale}): Perturbation scale. 
    Higher values expand the search for stable intervals. 
    \item \(N_p\) (\emph{count of perturbation samples}): Number of perturbed samples for interval estimation. 
    More samples improve precision, but increase computational cost. 
\end{itemize}
Optuna searches over the hyperparameters \(p\), \(\delta\), \(\sigma_p\) and \(N_p\) to maximize the objective in a holding ‐out test set. 
During each trial, candidate configurations are evaluated by applying uniform sampling to estimate rule stability and decision region consistency, 
ensuring stable decision regions in the input space of the ML model.  
The best configuration is selected based on its performance in the test set, ensuring that the chosen thresholds generalize beyond the training data. 
The tuning of these parameters balances feature selectivity, interval granularity, and computational cost.  
Table \ref{tab:hyperparameters} (Appendix) summarizes the hyperparameter ranges used in our study.  

Specifically, we define an \emph{objective function} maximizing \emph{coverage} and \emph{confidence} while penalizing empty or complex rules during \emph{numeric explanations to the rule transformation step} which is then fed to Optuna framework. 

\paragraph{The objective function used in the optimization:}
Each instance \( n \) starts with a complex rule from an explainer (e.g. \emph{Anchor}, \emph{LIME}, \emph{SHAP}).  
Each rule defines intervals for features, e.g. timestep-channel \( \text{t}_{i}\text{c}_{v} \) in multivariate TS or timestep \( \text{t}_{i} \) in univariate TS, denoted \( F(R_n) \). 
The rules yield metrics: \emph{coverage} \( \operatorname{COV}(R_n) \), \emph{confidence} \( \operatorname{CONF}(R_n) \), and \emph{count of features that the rule specifies} \( |F(R_n)| \).  
The initial \emph{objective function} is as follows: 
\begin{equation}
M(n) = {COV}(n) \times {CONF}(n).
\end{equation}
If rule for the instance \( n \) is empty or absent, \( M(n) = 0 \). 
Otherwise, penalties are applied when certain thresholds are not met. 

Then we introduce the thresholds \( \tau_{conf} \), \( \tau_{cov} \), and \( \tau_{feat} \), which are chosen to balance the quality and interpretability of the rules. 
\( \tau_{conf} \) ensures a minimum confidence level for the rules, \( \tau_{cov} \) prevents overly narrow rules by enforcing minimum coverage, and \( \tau_{feat} \) limits the number of features in a rule to improve cognitive interpretability for end-users. 
We set \(\tau_{conf}=0.5\) because scikit-learn and many binary classifiers use 0.5 as the default probability cutoff, aligning the applicability of the rule with positive predictions above the chance level and following the calibration practices. 
The coverage threshold \(\tau_{cov}=0.01\) (1\% support) corresponds to the minimum support filter commonly adopted in association rule mining to remove infrequent patterns and ensure rule generality. 
We fix \(\tau_{feat}=10.0\) to limit the number of conjuncts per rule to at most ten, a choice motivated by default max-length settings in popular rule-mining libraries and grounded in human cognitive constraints defined by Miller’s Law~\citep{Miller1956} of \(7\pm2\) chunks for short-term memory.

\begin{equation}
M(n) :=
\begin{array}{ll}
M(n) \times \frac{{CONF}(n)}{\tau_{conf}}, & \text{if } 0<{CONF}(n)<\tau_{conf}, \\[0.5em]
M(n) \times \frac{{COV}(n)}{\tau_{cov}}, & \text{if } 0<{COV}(n)<\tau_{cov}, \\[0.5em]
M(n) \times \frac{\tau_{feat}}{Fcount(n)}, & \text{if } |F(R_n)|>\tau_{feat}.
\end{array}
\end{equation}
\vspace{10pt}
These conditions are applied sequentially. 
The overall objective function is the average over all \( N \) instances:
\(
\bar{M} = \frac{1}{N}\sum_{n=1}^{N}M(n).
\)
Maximizing \(\bar{M}\) finds hyperparameters balancing \( {COV}(n) \) (not too low) and \( {CONF}(n) \) (not too low), avoiding empty solutions and minimizing the complexity of the rules. 

\subsubsection{Resolution of rules ambiguity} \label{rules_ambiguity}
Several equally valid rules can be derived for a single instance and require subsequent resolution.
When transforming numeric explanations into symbolic rules, different valid rules may arise depending on training data and hyperparameters. 
Consider three distinct timestep–channel pairs \(t_{j_1}c_{j_1}\), \(t_{j_2}c_{j_2}\), and \(t_{j_3}c_{j_3}\). 
The following interval rules might be derived for an arbitrary instance \(i\): 
\begin{align}
R_{i,1}:&\;\bigl\{\,t_{j_1}c_{j_1}\in(\ell_{1},u_{1}]\bigr\}\;\Rightarrow\;\hat y = A,\;\mathrm{CONF}(R_{i,1})=\gamma_{1},\;\mathrm{COV}(R_{i,1})=\kappa_{1}, \label{eq:R_i1}\\
R_{i,2}:&\;\bigl\{\,t_{j_2}c_{j_2}\in(\ell_{2},u_{2}]\bigr\}\;\Rightarrow\;\hat y = A,\;\mathrm{CONF}(R_{i,2})=\gamma_{2},\;\mathrm{COV}(R_{i,2})=\kappa_{2}, \label{eq:R_i2}\\
R_{i,3}:&\;\bigl\{\,t_{j_3}c_{j_3}\in(\ell_{3},u_{3}]\bigr\}\;\Rightarrow\;\hat y = A,\;\mathrm{CONF}(R_{i,3})=\gamma_{3},\;\mathrm{COV}(R_{i,3})=\kappa_{3}, \label{eq:R_i3}
\end{align}
Here, each \(t_{j_m}c_{j_m}\) denotes \emph{"channel"} \(c_{j_m}\) (in multivariate series) at timestep \(t_{j_m}\), each interval \((\ell_{m},u_{m}]\) has a lower bound \(\ell_{m}\) and an upper bound \(u_{m}\), and \(\gamma_{m},\kappa_{m}\in[0,1]\). 
Although \(R_{i,1}\), \(R_{i,2}\), and \(R_{i,3}\) yield equivalent predictions due to temporal correlations in \emph{TS}, rule ambiguity arises because explainers (e.g. \emph{SHAP}, \emph{LIME}) combined with hyperparameter-driven optimization select different variables. 
This optimization, which maximizes an objective defined in Section~\ref{sec:optuna}, resolves the ambiguity within a single explainer by choosing, for example, \(R_{i,\mathrm{shap}}\) for \emph{SHAP-based}-based attribution. 
However, the same procedure on \emph{LIME} may produce \(R_{i,\mathrm{lime}}\). 
To handle these divergent explanations, in Section~\ref{sec:rule_ensemble} we propose fusing the rule set
\(
\mathcal{R}_i = \{\,R_{i,\mathrm{shap}},\;R_{i,\mathrm{lime}},\;R_{i,\mathrm{anchor}}\,\}
\)
and selecting a single optimized rule per instance.

\subsection{Rule fusion across explainers} \label{sec:rule_ensemble}
In rule-based systems theory~\citep{ligeza_rule_based}, combining rules is called \emph{rule aggregation}.  
Our framework calls this \emph{rule fusion}, as it \emph{consolidates multiple rules into exactly one} whenever possible.  
Unlike standard aggregation, our method also applies to a \emph{single explainer}, assigning each instance a rule refined via metric optimization (e.g. \texttt{Lasso} method discussed later).   

Let \(\mathcal{F}\) be the feature set and \(M\) are various explanations methods for \emph{TS classification model} (e.g. rules obained from \emph{SHAP}).  
Features are indexed by \( f \), each representing a timestep-channel pair \((t,c)\) for multivaraite TS, or timestep \((t)\) for univariate TS. 
Explanations methods are indexed with \( m \). 
Each method starts with multiple rules \( R_{i}^{(m)} \) from different explainers \( m \).  
Each rule contains feature conditions:
\(
R_{i}^{(m)} = \{ (f, (l_{f}^{(m)}, u_{f}^{(m)}]) \mid f \in F_{i}^{(m)} \},
\)
where \((l_{f}^{(m)}, u_{f}^{(m)}]\) is an interval for feature \( f \).
\noindent
We propose the following \emph{rule fusion} approaches:

\noindent\texttt{Intersection:}
Only conditions present in \emph{all} rules per instance are kept. For each feature \( f \) that appears in all \( R_{i}^{(m)} \), we take the \emph{narrowest} intersection of intervals:
\begin{equation}
R_{i}^{\cap} = \bigcap_{m} R_{i}^{(m)}.
\end{equation}
In practice, if a feature \( f \) has intervals \((l_{f}^{(m)}, u_{f}^{(m)}]\) in each rule, we take:
\begin{equation}
l_{f}^{\cap} = \max_{m}(l_{f}^{(m)}), \quad u_{f}^{\cap} = \min_{m}(u_{f}^{(m)}).
\end{equation}

\noindent\texttt{Union:}
All conditions appearing in any rule are included (instance wise). For features shared among multiple rules, we take the \emph{broadest} union of intervals:
\begin{equation}
R_{i}^{\cup} = \bigcup_{m} R_{i}^{(m)}.
\end{equation}
If a feature \( f \) has intervals from different rules, we set:
\begin{equation}
l_{f}^{\cup} = \min_{m}(l_{f}^{(m)}), \quad u_{f}^{\cup} = \max_{m}(u_{f}^{(m)}).
\end{equation}

\noindent\texttt{Weighted:}
We define a weight \( w_{m} \) for each method \( m \), derived from a chosen metric (e.g. \emph{confidence}). We compute a weighted presence of each feature and select only those exceeding a threshold \(\tau\):
\begin{equation}
\frac{\sum_{m} w_{m}\chi_{f}^{(m)}}{\sum_{m} w_{m}} > \tau
\end{equation}
where \(\chi_{f}^{(m)}=1\) if feature \( f \) appears in \( R_{i}^{(m)} \). For the selected features, intervals are combined by union.

\noindent\texttt{Lasso and Lasso global:}
Each condition (feature-interval pair) is encoded as a binary \( Z_{n,j} \), indicating if sample \( n \) meets condition \( j \).  
We then solve:
\begin{equation}
\min_{\beta} \|y - Z\beta\|_2^2 + \lambda\|\beta\|_1.
\end{equation}
Conditions with \(\beta_{j} \approx 0\) are discarded, yielding sparser rules.  
\texttt{Lasso local}, \texttt{Lasso} for short, fits per instance; \texttt{Global lasso} finds conditions across all instances at once to find globally important conditions.  
Multiple conditions per feature are combined by union.  
If no rules exist for an observation in \texttt{Lasso global}, imputation is used.  
Rules are first collected per observation from different explainers (e.g. \emph{SHAP}, \emph{LIME}, \emph{Anchor}).  
If no rule is assigned to an observation, the \emph{most frequent rule in the predicted class} is used.  
If none exist, the \emph{most frequent rule overall} is assigned as fallback.  
In \texttt{Lasso local} (later referred as \texttt{Lasso}) and other methods, observations without rules are excluded.  

\noindent\texttt{Best:}
We select the single rule that maximizes a chosen metric (e.g. confidence):
\begin{equation}
R_{i}^{\text{best}} = \arg\max_{R_{i}^{(m)}}(\text{metric}(R_{i}^{(m)})).
\end{equation}
This approach avoids merging and simply picks the best-performing rule instance wise.

\noindent\texttt{Finalization of resultant rules:} 
The \emph{confidence} and \emph{coverage} of newly derived rules are computed as described in Section~\ref{sec:new_rule_conf_cov} and added to each rule, quantifying how well the fused rule generalizes to the data.

\section{Evaluation} \label{sec:evaluation}
\subsection{Model architecture}
All time series datasets were classified using a unified deep neural architecture based on stacked \emph{ConvLSTM1D} layers. 
The input consisted of time series samples with shape $(T, C)$, where $T$ denotes the number of time steps and $C$ the number of \emph{"channels"}. 
For univariate datasets, $C = 1$. 
To capture both local and global temporal dependencies, each sequence was segmented into $n_{\mathrm{steps}}$ temporal blocks, where $n_{\mathrm{steps}}$ was chosen as the third smallest divisor of $T$, excluding $1$ and $2$. 
The segment length was computed as $n_{\mathrm{length}} = T / n_{\mathrm{steps}}$, resulting in an input shape of $(n_{\mathrm{steps}}, n_{\mathrm{length}}, C)$. 

\begin{figure}[h!]
  \centering
  \includegraphics[width=1.0\linewidth]{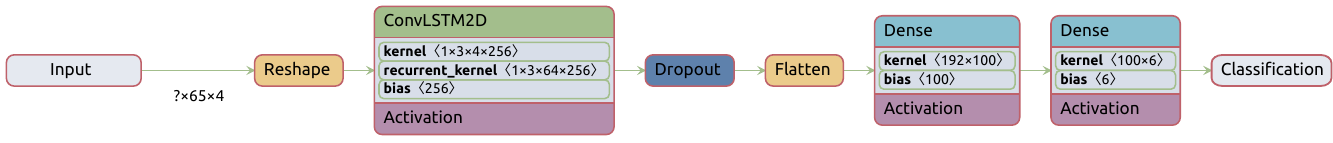}
  \caption{Overview of the architecture of the ConvLSTM1D-based model used in the experiments. 
  }
  \label{fig:model_architecture}
\end{figure}

The model architecture included two ConvLSTM1D layers with $64$ and $32$ filters, respectively, using a kernel size of $9$ and ReLU activation. 
A dropout layer with a rate of $0.5$ followed, then a flattening operation producing a fixed-length embedding. 
This was followed by two fully connected layers: a dense layer with $100$ ReLU units and an output layer with softmax activation. 
Class imbalance was handled using computed class weights. 
The model was trained with the Adam optimizer, categorical cross-entropy loss, for $25$ epochs with a batch size of $64$. 
The same architecture was used across all univariate and multivariate datasets without dataset-specific tuning. 
A random $75:25$ train-test split was applied with stratification by class label.

A graphical summary of this architecture is shown in Figure~\ref{fig:model_architecture}, illustrating the flow from segmented TS input through ConvLSTM layers to the final classification output.

\subsection{Computational complexity of PHAR} \label{sec:computational_efficiency}
The computational cost of the proposed \emph{PHAR} framework is primarily determined by two factors: the application of post-hoc explainers to time series data and the subsequent conversion of numeric attributions into symbolic rule representations and combining rules with fusion. 

To provide a clear and structured analysis of computational complexity, we divide this discussion into three dedicated subsections, each corresponding to a major component of the \emph{PHAR} pipeline. 
First, we evaluate the computational cost of the \emph{base explainers}, such as \emph{SHAP}, \emph{LIME}, and \emph{Anchor}, which generate the initial feature attributions or rule sets. 
Second, we analyze the cost of the \emph{numeric-to-rule transformation}, which converts numeric explanations into human-readable symbolic rule intervals via perturbation-based sampling and interval search. 
Finally, we assess the complexity of the \emph{rule fusion process}, which consolidates multiple rule sets into a single, optimized rule per instance. 

\subsubsection{Computational complexity of base explainers}
\textbf{SHAP} explanations, based on kernel SHAP, have a computational complexity of $O(M^2)$, where $M$ is the number of input features~\citep{lundberg2017unifiedapproachinterpretingmodel}. 
Exact SHAP computations are exponential ($O(2^M)$), but Kernel SHAP approximates Shapley values via weighted linear regression, which still scales quadratically with $M$~\footnote{\url{https://shap.readthedocs.io/en/latest/generated/shap.KernelExplainer.html}}.

\textbf{LIME} explanations involve sampling perturbed instances and fitting a local surrogate model (typically linear regression), resulting in complexity $O(S \cdot M)$, where $S$ is the number of samples~\citep{ribeiro2016why}. 
Further details are provided in the LIME documentation~\footnote{\url{https://github.com/marcotcr/lime} }.

\textbf{Anchor} explanations involve a beam search procedure over binary feature subsets, with complexity proportional to the size of the search space and sampling budget, typically $O(B \cdot S)$ where $B$ is beam width and $S$ sampling per candidate~\citep{Ribeiro_Singh_Guestrin_2018}, see appropriate github repository\footnote{\url{https://github.com/marcotcr/anchor}}.

\subsubsection{Computational complexity of numeric-to-rule transformation}
The \emph{PHAR} pipeline performs two core steps: \emph{thresholding} and \emph{interval derivation}; and these are wrapped inside a \emph{hyperparameter optimization} loop. 
PHAR repeatedly applies thresholding and interval generation to evaluate different parameter configurations, and once the best hyperparameters are selected, it executes those two steps one final time to produce the final rule set. 
In the following subsections, we first estimate the computational cost of hyperparameter optimization (i.e., the number of repeated runs), and then analyze the individual complexities of the thresholding and interval derivation steps.

\textbf{Hyperparameter optimization} guides the selection of optimal threshold and perturbation settings. 
The search tunes four key parameters: the percentile threshold \( p \in [50,99] \), a binary flag \( \emph{global} \in \{\text{true},\text{false}\} \), perturbation scale \( \sigma \in [0.01,1.0] \), and number of perturbation samples \( S_p \in [1000,10000] \) (see Table~\ref{tab:hyperparameters}, Appendix). 
We use Optuna, a Bayesian optimization framework that adaptively explores this space and prunes unpromising trials early~\citep{optuna_2019}. 
In the worst-case scenario: if no pruning occurs—the cost scales with the number of evaluated trials \( T \), which can be as high as the total number of possible combinations in the Cartesian product of parameter ranges. 
However, due to pruning and adaptive sampling, the actual number of full evaluations is observed to be a small fraction of \( T \). 
For practical use, PHAR allows bypassing this stage entirely by applying default settings, e.g. \( p = 90 \), \( \emph{global} = \text{true} \), making the optimization step optional.

\textbf{Thresholding} requires a linear pass over the explanation values with complexity $O(N \cdot F)$, where $N$ is the number of instances and $F$ the number of features. 

\textbf{Interval derivation} constitutes the most computationally intensive step, with complexity scaling as $O(N \cdot F' \cdot S_{p})$, where $F' \ll F$ denotes the average number of important features selected after thresholding, and $S_{p}$ is the number of perturbation samples generated per instance. 
In practice, $S_{p}$ is user-defined (typically $\leq 10000$), and both the perturbation sampling and subsequent model evaluations are highly parallelizable. 
This keeps the overall runtime feasible for time series datasets of moderate size, such as those commonly found in the \emph{UCR/UEA archives}, where $N$ usually ranges from several hundred to a few thousand instances, and $F$ corresponds to tens or low hundreds of input features~\cite{Dau2019UCRArchive,Bagnall2018UEAMultivariate}. 
This design ensures that PHAR remains computationally tractable on standard multi-core hardware.

\subsubsection{Computational complexity of rule fusion}
\textbf{Rule fusion} in PHAR uses lightweight aggregation strategies and adds negligible computational overhead compared to the previous steps. 

\texttt{Intersection}, \texttt{Union}, and \texttt{Weighted} methods select the final rule for each instance by aggregating multiple candidate rules originating from different explainability sources. 
These methods rely on precomputed metrics embedded within the candidate rules (e.g. confidence, coverage) and do not require re-evaluation of the predictive model. 
Their computational cost scales linearly with the number of rule sources $M$ and the number of explained instances $N$, resulting in an overall complexity of $O(N \cdot M)$. 
Since the number of candidate rule sources per instance is typically low (e.g. $M = 2$ or $3$), this step introduces negligible overhead relative to earlier phases.

\texttt{Lasso} involves solving a sparse linear regression problem, scaling as $O(N \cdot F_{r}^{2})$ for each instance or globally across the dataset. 
However, the rule matrix is highly sparse, allowing efficient optimization via coordinate descent~\citep{Tibshirani1996}. 

Overall, fusion contributes marginally to total computation time, with the primary bottleneck being the initial attribution computation.

\subsection{Evaluation metrics and statistical testing}
We compared the \texttt{baseline explainers} and \texttt{fusions} of two or three explainers.  
\(\bar{M}\) was applied as described in Section~\ref{sec:optuna}.  
The metric \(ER\) counts instances where the explainer generated rules in proportion to all cases.  
\(\overline{\text{CONF}}\) and \(\overline{\text{COV}}\), as in Section~\ref{sec:optuna}, were averaged over all instances.  
\(\overline{F(n)}\) and \(Med(F(n))\) represent the \emph{mean} and \emph{median} number of features in generated rules (for instances with rules).  
Furthermore, we defined composite metrics \(\overline{\text{CONF}} \cdot \text{ER}\) and \(\overline{\text{CONF}} \cdot \overline{\text{COV}} \cdot ER\), which combine \emph{confidence}, \emph{coverage}, and rule generation rates.  
To compare \texttt{explainers} and \texttt{fusion methods} statistical evaluation uses non-parametric \emph{Friedman} test and post-hoc analysis with \emph{Nemenyi} tests.  
Metrics are visualized for univariate and multivariate datasets to highlight key differences.  
Package \emph{scikit-posthocs}~\citep{Terpilowski2019} was used.  

\subsection{Setup: time series, explainers, and rule fusion}
The datasets used in this study come from the \emph{UCR/UEA Time Series Classification Archive}. 
Some multivariate datasets (e.g. \emph{Libras}, \emph{PenDigits}) originate from the \emph{UCI} ML Repository but are used via their curated UCR/UEA\footnote{UCR/UEA archive: \url{https://www.cs.ucr.edu/~eamonn/time_series_data_2018/} and \url{https://www.timeseriesclassification.com/}. 
Selected datasets (e.g. \emph{PenDigits}) originate from UCI: \url{https://archive.ics.uci.edu/}.} versions~\citep{Dau2019UCRArchive,Bagnall2018UEAMultivariate}. 
Our code is available at GitHub\footnote{\url{https://github.com/mozo64/papers/tree/main/phar/notebooks}} in the following notebooks: \emph{UCR-explainers-lime-shap-anchor.ipynb}, \emph{UCR-num2rules-optuna.ipynb}, \emph{UCR-fusions-results.ipynb}.

We evaluated our methods on 43 datasets, including 81.4\% univariate and 18.6\% multivariate TS. 
These datasets cover diverse domains.
For each dataset, a TensorFlow model classified the TS samples, followed by explanations generated using Anchor, SHAP, and LIME at the variable (\(>1\) for multivariate) and time point levels. 
SHAP and LIME were converted into rule-based explanations using the process in~\ref{sec:optuna}. 
A separate test set, unseen during training, was used for evaluation. 
Table~\ref{tab:dataset_summary} summarizes the datasets, including the number of features used in each dataset, explainers, fusion methods, and resulting fused rule sets.

Across the benchmark, Anchor, LIME, and SHAP served as primary baselines and provided the rule sets on which fusion methods were built. 
Due to the computational cost of running some explainers and fusion procedures on long, high-dimensional time series, not every explainer could be executed on all 43 datasets within our runtime budget. 
Consequently, Anchor, LIME, and SHAP contributed baseline rules in 58.1\%, 65.1\%, and 62.8\% of the datasets, respectively. 

Fusion techniques such as \texttt{Lasso}, \texttt{Lasso global}, and \texttt{Weighted} were applied to all datasets (100.0\%), since they can refine rules even when only a single explainer is available. 
In contrast, \texttt{Best}, \texttt{Intersection}, and \texttt{Union} were only used when at least two distinct baseline explainers were available, otherwise, they would trivially replicate the underlying baseline. 
As a result, these three fusion methods were applied in 62.8\% of the datasets. 
By construction, the \texttt{Baseline} method appears in all datasets (100.0\%) as a reference. 
Finally, fused rule sets combining two and three explainers were applicable in 62.8\% and 23.3\% of the datasets, respectively, reflecting the fraction of problems for which two or all three baseline explainers were computed.

\subsection{Evaluation of \emph{fusion methods}}
\subsubsection{Pairwise and aggregated comparisons} 

\noindent \textbf{Performance of baseline explainers:}  
We evaluated \emph{Anchor} and rules derived from \emph{LIME} and \emph{SHAP} under the \texttt{baseline} condition, representing their original, untransformed explanations.  
Unlike fusion methods that optimize rule structures, \texttt{baselines} reflect each explainer's native performance.  
To compare them, we applied the \emph{Wilcoxon signed-rank test} on matched datasets, ensuring robustness for non-normal data.  
Table~\ref{tab:wilcoxon_single_explainers} presents the statistical results, while 
Figure~\hyperlink{fig:single_explainers_plots__all}{A1} visualizes the distribution of key metrics. 

\noindent \textbf{Comparison of fusions to \emph{Anchor} baseline:}  
We applied the \emph{Wilcoxon signed-rank test} to compare fusion methods with individual explainers, considering two perspectives.  
Tables~\ref{tab:wilcoxon_rules_combination_final_metric}--\ref{tab:wilcoxon_rules_combination_median_features_count} analyze fused rule sets, such as \emph{Anchor+LIME} or \emph{LIME+SHAP}, against the \emph{Anchor baseline}.  
This evaluation determines whether combining multiple explainers improves key metrics. 

\noindent \textbf{Comparison of fusion methods to their respective baselines:}  
Tables~\ref{tab:wilcoxon_method_final_metric}--\ref{tab:wilcoxon_method_median_features_count} compare different fusion strategies, including \texttt{Lasso}, \texttt{Weighted}, and \texttt{Intersection}, to individual explainers like \emph{Anchor}, \emph{LIME}, and \emph{SHAP}. 

\noindent \textbf{Visualization of comparisons:}  
Figure~\hyperlink{fig:merged_boxplots_1}{A2} visualizes metric distributions, contrasting single explainers (orange) with fusion methods (blue) for \(\bar{M}\), \(\overline{\text{CONF}} \cdot ER\), \(\overline{\text{COV}}\), and \(F(n)\).  
Each boxplot includes mean, standard deviation, and median values.  
Additional plots detail results contrasting single explainers baseline with fusion explainers: Figure~\hyperlink{fig:ensemble_vs_single_objective_fun}{A3} (\(\bar{M}\)), Figure~\hyperlink{fig:ensemble_vs_single_conf_er}{A4} (\(\overline{\text{CONF}} \cdot ER\)), Figure~\hyperlink{fig:ensemble_vs_single_cov}{A5} (\(\overline{\text{COV}}\)), and Figure~\hyperlink{fig:ensemble_vs_single_avg_features_count}{A6} (\(F(n)\)).  

\noindent \textbf{Median values of metrics:} 
Table~\ref{tab:median_summary_rules_combination_all_metrics} reports the summary of median values, comparing rule sets derived from different explainers and theirs \texttt{fusions}, such as \emph{Anchor} and \emph{LIME+SHAP}, regardless of \texttt{fusion method}. 
Table~\ref{tab:median_summary_methods_all_metrics} presents median values for different \emph{fusion strategies} applied to explainers and theirs combinations, e.g. \texttt{Intersection}, \texttt{Weighted}, .... 

\subsubsection{Ranking-based and post-hoc analysis}  

\begin{figure*}[ht]
  \centering

  \subfloat[\(\bar{M}\)\label{fig:cd_methods_final_metric}]{
    \includegraphics[width=0.45\textwidth]{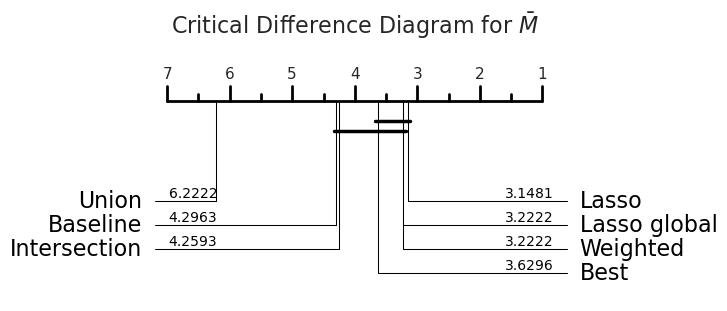}
  }\hfill
  \subfloat[\(\bar{\text{CONF}} \cdot ER\)\label{fig:cd_methods_conf_er}]{
    \includegraphics[width=0.45\textwidth]{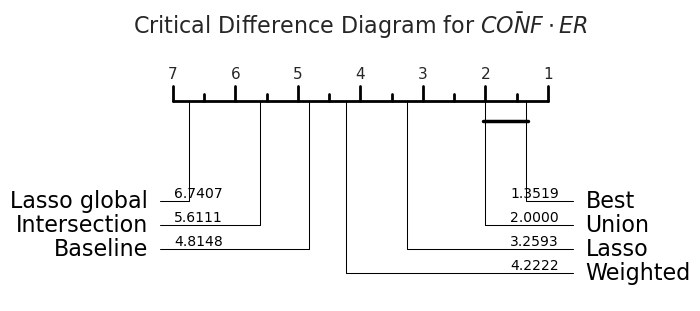}
  }

  \medskip

  \subfloat[\(\bar{\text{CONF}}\)\label{fig:cd_methods_conf}]{
    \includegraphics[width=0.45\textwidth]{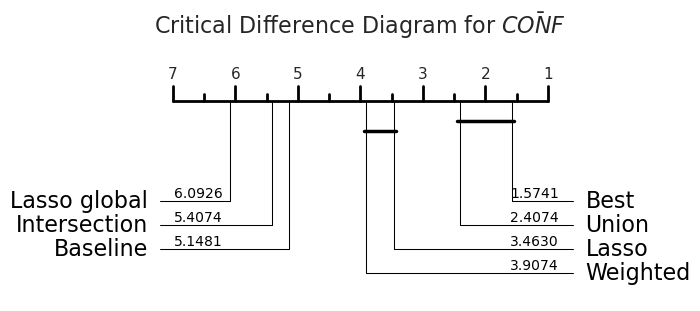}
  }\hfill
  \subfloat[\(\bar{\text{COV}}\)\label{fig:cd_methods_cov}]{
    \includegraphics[width=0.45\textwidth]{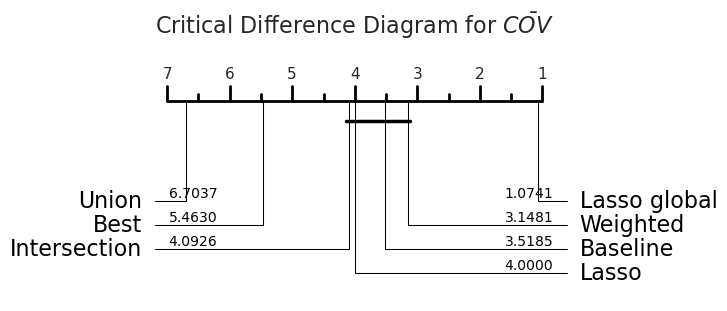}
  }

  \medskip

  \subfloat[\(\bar{F(n)}\)\label{fig:cd_methods_avg_features}]{
    \includegraphics[width=0.45\textwidth]{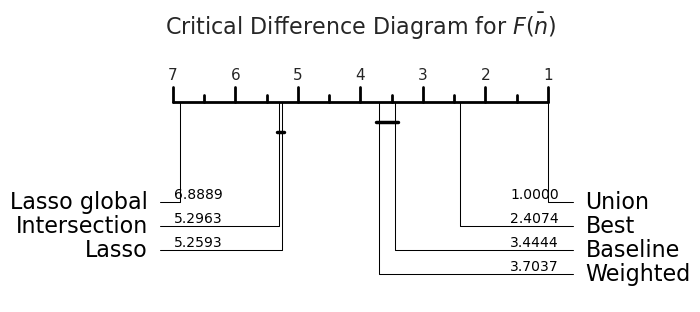}
  }\hfill
  \subfloat[\(ER\)\label{fig:cd_methods_explained_ratio}]{
    \includegraphics[width=0.45\textwidth]{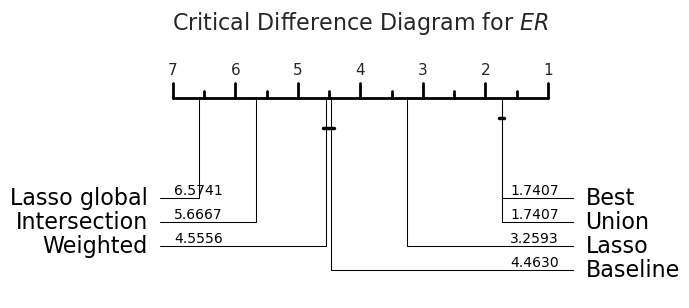}
  }

  \caption{Critical Difference Diagrams for selected metrics:
  (a) objective function \(\bar{M}\),
  (b) \(\bar{\text{CONF}} \times ER\),
  (c) \(\bar{\text{CONF}}\),
  (d) \(\bar{\text{COV}}\),
  (e) average feature count \(\bar{F(n)}\),
  (f) explained ratio \(ER\).
  Methods joined by horizontal lines do not differ significantly at \(\alpha=0.05\).}
  \label{fig:critical_diff_selection}
\end{figure*}

\begin{figure*}[ht]
  \centering

  \subfloat[\(\bar{M}\)\label{fig:cd_ensemble_objective_fun}]{
    \includegraphics[width=0.45\textwidth]{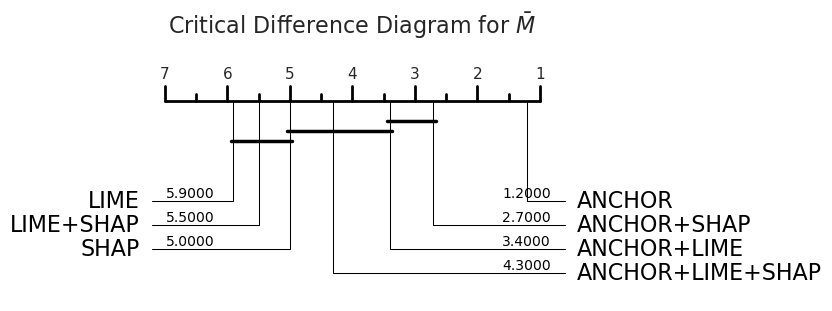}
  }\hfill
  \subfloat[\(\bar{\text{CONF}} \cdot ER\)\label{fig:cd_ensemble_conf_er}]{
    \includegraphics[width=0.45\textwidth]{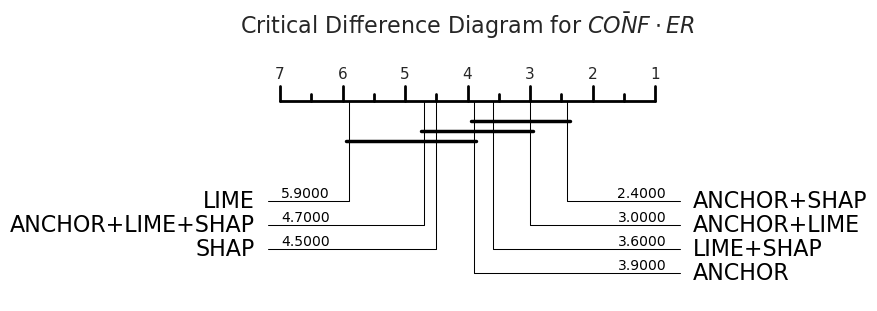}
  }

  \medskip

  \subfloat[\(\bar{\text{CONF}}\)\label{fig:cd_ensemble_conf}]{
    \includegraphics[width=0.45\textwidth]{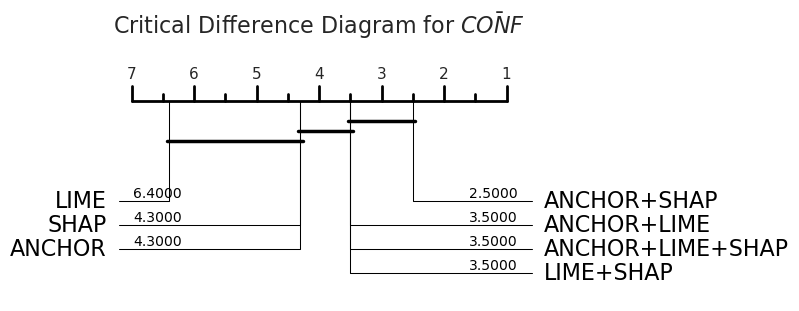}
  }\hfill
  \subfloat[\(\bar{\text{COV}}\)\label{fig:cd_ensemble_cov}]{
    \includegraphics[width=0.45\textwidth]{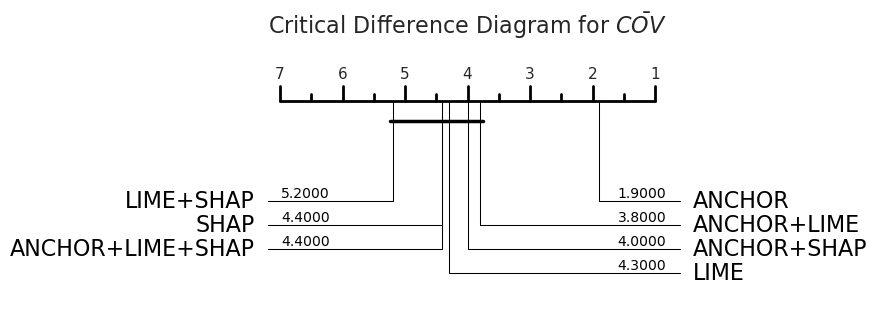}
  }

  \medskip

  \subfloat[\(\bar{F(n)}\)\label{fig:cd_ensemble_avg_features}]{
    \includegraphics[width=0.45\textwidth]{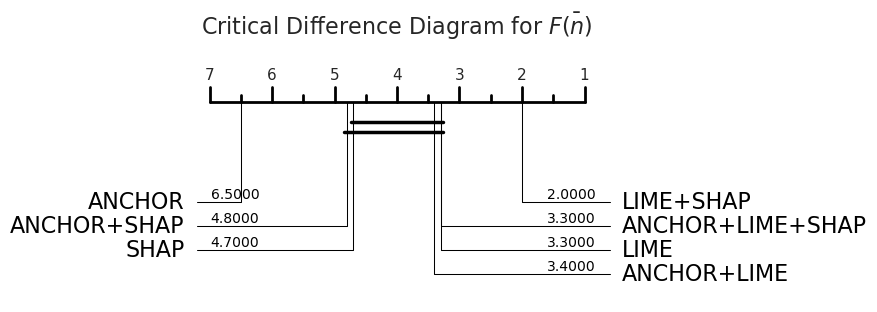}
  }\hfill
  \subfloat[\(ER\)\label{fig:cd_ensemble_explained_ratio}]{
    \includegraphics[width=0.45\textwidth]{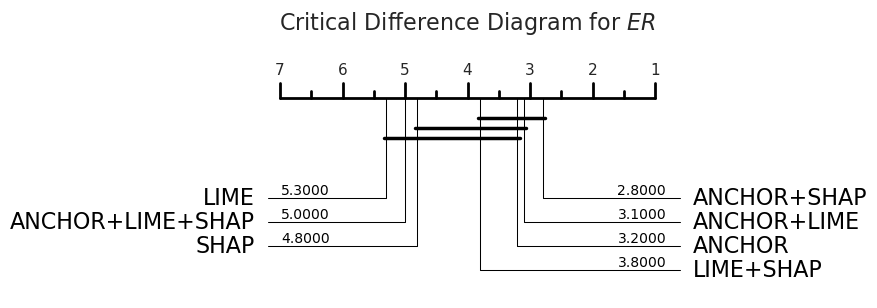}
  }

  \caption{Critical Difference Diagrams for selected metrics using fusion methods:
  (a) objective function \(\bar{M}\),
  (b) \(\bar{\text{CONF}} \times ER\),
  (c) \(\bar{\text{CONF}}\),
  (d) \(\bar{\text{COV}}\),
  (e) average feature count \(\bar{F(n)}\),
  (f) explained ratio \(ER\).
  Horizontal lines connect methods with no statistically significant differences (\(\alpha=0.05\)).}
  \label{fig:critical_diff_ensemble}
\end{figure*}

\noindent \textbf{Critical Difference Diagrams (CDD):}  
CDD rank methods across datasets based on their average performance.  
A lower rank generally indicates better results, as it corresponds to higher values of the evaluated metric, except for \(\bar{F(n)}\), where lower values are preferred, making higher ranks more desirable.  
Horizontal bars connect methods that do not differ significantly at \(\alpha = 0.05\), while gaps denote statistically significant differences.  
Figure~\ref{fig:critical_diff_selection} presents CDD for different fusion methods and baselines, ranking them according to their effectiveness across multiple metrics.  
Figure~\ref{fig:critical_diff_ensemble} extends the comparison to fused rule sets, showing how different explainers contribute to overall performance when used in combination.   
Additional figures in the appendix (Figures~\ref{fig:cd_final_metric_ens_and_m}--\ref{fig:cd_avg_features_ens_and_m}) provide detailed comparisons for \(\bar{M}\) , \(ER\) , and \(\bar{F(n)}\) , incorporating both explainer-level analysis (including individual explainers and their combinations) and specific fusion methods applied to these rule sets.

Detailed tables with Friedman and Nemenyi post-hoc comparisons are provided in "Supplementary Materials" (Section~\ref{sec:supplement}) in Tables~\ref{tab:friedman_method_all_metrics}–~\ref{tab:nemenyi_method_median_features_count}. 

\subsubsection{Compound metrics \texorpdfstring{\boldmath$\bar{M}$}{M}, \texorpdfstring{$\overline{\text{CONF}} \cdot ER$}{CONF ER} and \texorpdfstring{$\overline{\text{CONF}} \cdot \overline{\text{COV}} \cdot ER$}{CONF COV ER}}
Using Wilcoxon signed-rank tests, we first evaluated the baseline performance of \emph{Anchor}, \emph{LIME}, and \emph{SHAP}, confirming slight advantage of \emph{Anchor}. 
Among \emph{LIME} and \emph{SHAP}, the latter performed better for $\overline{\text{CONF}} \cdot ER$. 
Next, we compared fusion methods to the \emph{Anchor} baseline. 
For $\bar{M}$ and $\overline{\text{CONF}} \cdot \overline{\text{COV}} \cdot ER$, fusions generally showed no significant gains, indicating that combining multiple explainers without considering the fusion method does not inherently improve results. 
However, when each fusion strategy was compared to its respective single-explainer baseline, we observed consistent improvements across all three compound metrics, with the best results from \texttt{Lasso}, \texttt{Weighted}, and \texttt{Best}. 
Other methods yielded comparable, worse, or statistically insignificant results.
Friedman tests confirmed significant differences ($p < 0.05$) among both \texttt{fusion methods} and \emph{explainer combinations}. 
Nemenyi post-hoc analysis further highlighted differences in performance, particularly for the best-performing methods. 
{Critical Difference Diagrams} (CDD) reinforced these findings, ranking \texttt{Lasso}, \texttt{Weighted}, and \texttt{Best} as the top fusion strategies. 
Among explainer combinations, \emph{Anchor} and its fusions performed best, with \emph{LIME+SHAP} achieving comparable results.

\subsubsection{Confidence \texorpdfstring{\boldmath$\overline{\text{CONF}}$}{CONF}, Coverage \texorpdfstring{\boldmath$\overline{\text{COV}}$}{COV}, and Explained Ratio \texorpdfstring{\boldmath$ER$}{ER}}
Wilcoxon signed-rank tests showed \emph{SHAP} outperforming \emph{LIME} in $\overline{\text{CONF}}$, while \emph{Anchor+SHAP} exceeded \emph{Anchor}. 
Most multi-explainer fusions improved $\overline{\text{COV}}$, but no significant differences were found for $ER$. 
Fusion methods \texttt{Best}, \texttt{Lasso}, \texttt{Union}, and \texttt{Weighted} consistently improved all three metrics over baselines. 
Friedman tests confirmed significant differences ($p < 0.05$), and Nemenyi post-hoc analysis validated the superior performance of these fusion methods. 
CDD illustrate these trends, ranking \texttt{Best}, \texttt{Lasso}, \texttt{Union}, and \texttt{Weighted} highest for $\overline{\text{CONF}}$, $\overline{\text{COV}}$, and $ER$. 
\texttt{Lasso global} excelled in $\overline{\text{COV}}$. 
Among explainer combinations, \emph{Anchor}-based fusions ranked highest, with \emph{LIME+SHAP} performing well for $\overline{\text{CONF}}$ and $ER$. 

\subsubsection{Features Count \texorpdfstring{\boldmath$\overline{F(n)}$}{F(n)} and \texorpdfstring{\boldmath$\mathrm{Med}(F(n))$}{Median F(n)}}
Wilcoxon signed-rank tests showed \emph{Anchor} using the fewest features, with \emph{LIME} generating fewer than \emph{SHAP}. 
Against the \emph{Anchor} baseline, all fused explainers had higher $\overline{F(n)}$ and $\mathrm{Med}(F(n))$. 
Among fusion methods, \texttt{Intersection} and both \texttt{Lasso} variants minimized feature count, while \texttt{Union} ($42.6$ vs. $23.6$) and \texttt{Best} ($27.5$ vs. $23.6$) increased it. 
Friedman and Nemenyi tests confirmed significant differences across methods and explainer combinations. 
CDD ranked \texttt{Lasso}, \texttt{Lasso global}, and \texttt{Intersection} as the most efficient in minimizing $\overline{F(n)}$, while \emph{Anchor}, \emph{SHAP}, and \emph{Anchor+SHAP} ranked best in that regard among explainer combinations. 

\subsection{Summary of evaluation results}
\emph{Anchor} consistently performed best in $\bar{M}$ and minimizing feature count, but was not dominant in $\overline{\text{CONF}}$, $\overline{\text{COV}}$, and $ER$. 
Among explainers, \emph{SHAP} outperformed \emph{LIME} in $\overline{\text{CONF}}$, and \emph{Anchor+SHAP} surpassed \emph{Anchor}. 
Most multi-explainer fusions improved $\overline{\text{COV}}$, while no significant differences were found for $ER$.
Fusion methods had mixed effectiveness and choosing the right method is crucial. 
In compound metrics, \texttt{Lasso}, \texttt{Weighted}, and \texttt{Best} consistently improved performance. 
Some methods, such as \texttt{Union}, increased feature count significantly, while \texttt{Lasso} and \texttt{Intersection} minimized it. 
\emph{CDD} confirmed that fusions often outperformed individual explainers. 
The best methods depended on the metric: \texttt{Lasso}, \texttt{Weighted}, and \texttt{Best} excelled in $\bar{M}$ and compound metrics, while \texttt{Lasso global} was particularly effective in $\overline{\text{COV}}$. 
Among explainer combinations, \emph{Anchor}-based fusions ranked highest, but \emph{LIME+SHAP} also performed well in $\overline{\text{CONF}}$ and $ER$. 

\subsection{Qualitative evaluation} \label{sec:qual_eval}
We illustrate PHAR on the \emph{ECG200} dataset (two classes) using \emph{observation~21}. 
Because the data are \emph{univariate}, the channel index is omitted and rules refer directly to \emph{time-step} indices. 
For LIME and SHAP we convert per–time-step attributions into contiguous interval constraints using the procedure in Sec.~\ref{sec:numeric_to_rules}; Anchor provides a native rule set. 
We then form rule sets by three fusion operators (Sec.~\ref{sec:rule_ensemble}): 
\emph{union} (all intervals from the inputs are pooled into one conjunctive rule), 
\emph{lasso} (a sparse re–selection of intervals under an $\ell_1$ penalty), 
and \emph{best} (a single rule is chosen by a deterministic ranking: confidence has top priority; in a tie a fixed explainer order breaks ties).\footnote{Other tie–breakers, e.g.\ adding a sparsity term, are possible but not used here.} 
All values are standardised \(z\)-scores relative to the training set. 
Thus thresholds can be read as deviations from the mean, which makes the intervals actionable: one can see how much the signal would need to move to break or satisfy a condition. 
In practice, the interval conditions (\(t_i \in [\ell,u]\)) are contextualised as characteristic segments, such as an initial dip, a mid‑sequence plateau or a terminal return to baseline—which makes the explanation more natural than a bare list of thresholds. 

We recommend a two–step reading path. 
First, inspect the visualised intervals to understand \emph{where} in the trace the decision is anchored. 
Second, examine the explicit rule statements to see the exact inequalities. 
We follow this order below. 

\begin{figure*}[ht]
  \centering
  \subfloat[Anchor (benchmark)]{
    \includegraphics[width=.48\textwidth]{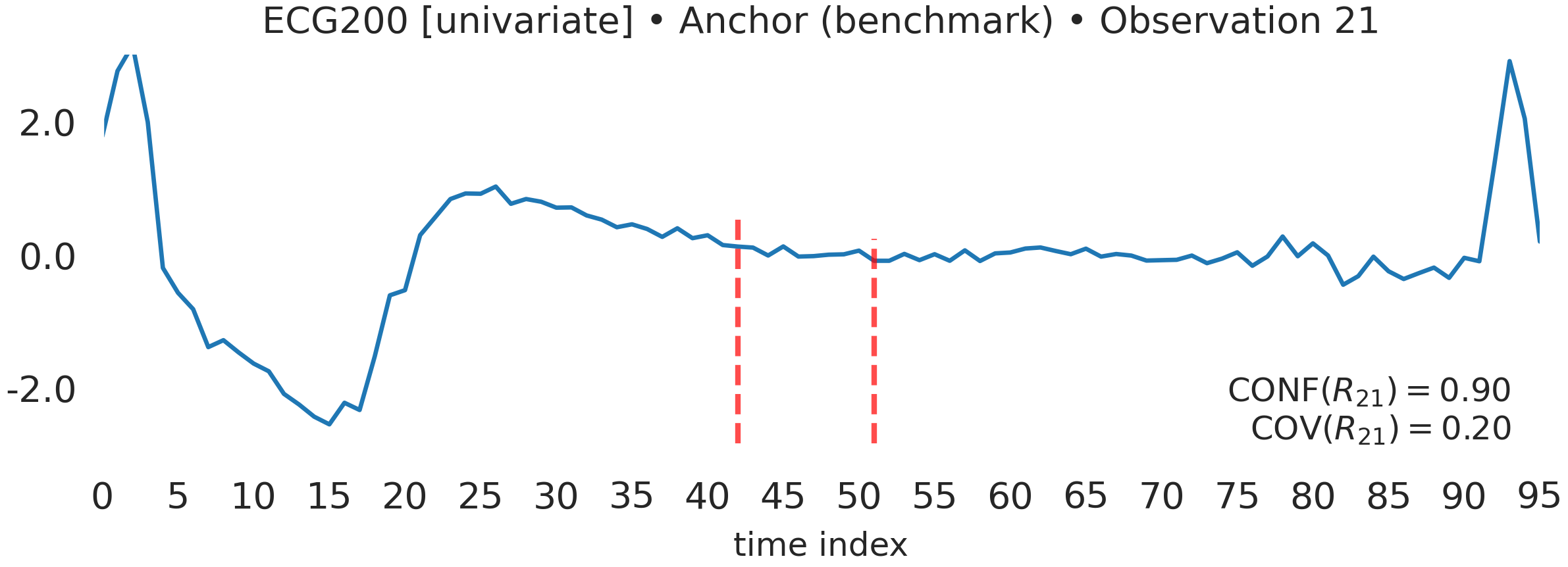}
  }\hfill
  \subfloat[LIME (benchmark)]{
    \includegraphics[width=.48\textwidth]{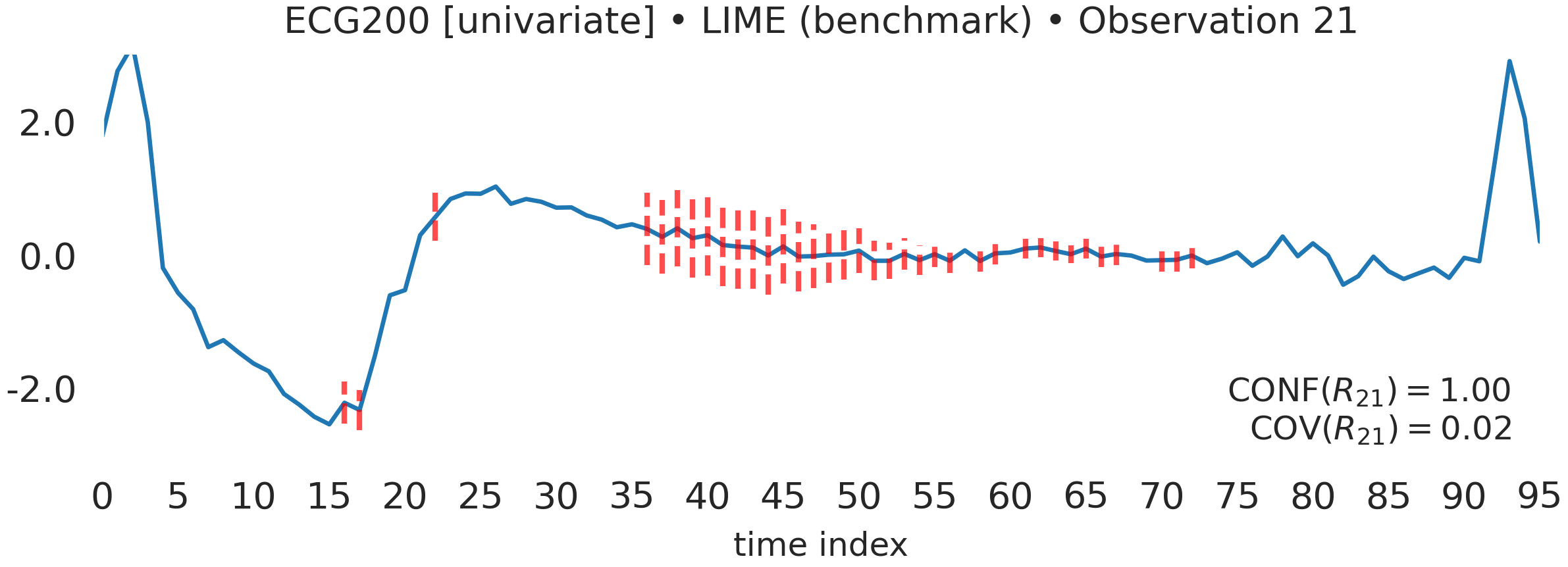}
  }\\[2mm]
  \subfloat[SHAP (benchmark)]{
    \includegraphics[width=.48\textwidth]{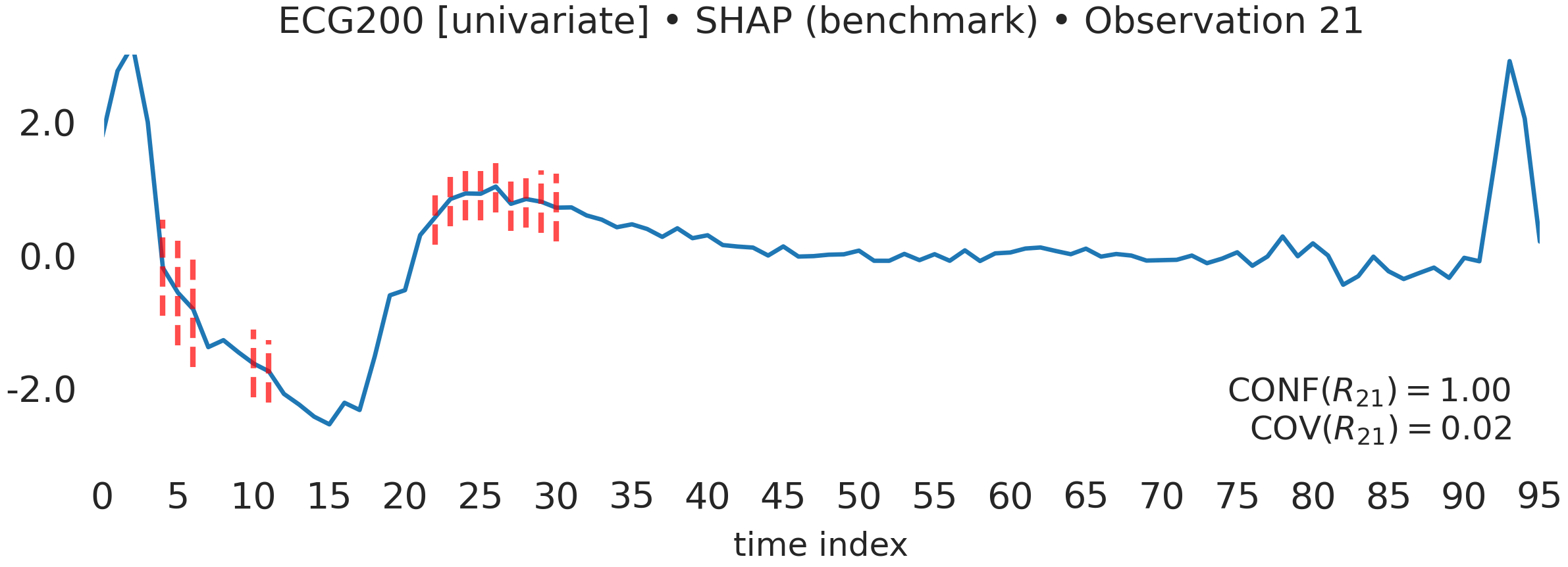}
  }\hfill
  \subfloat[Fusion (union): Anchor + SHAP]{
    \includegraphics[width=.48\textwidth]{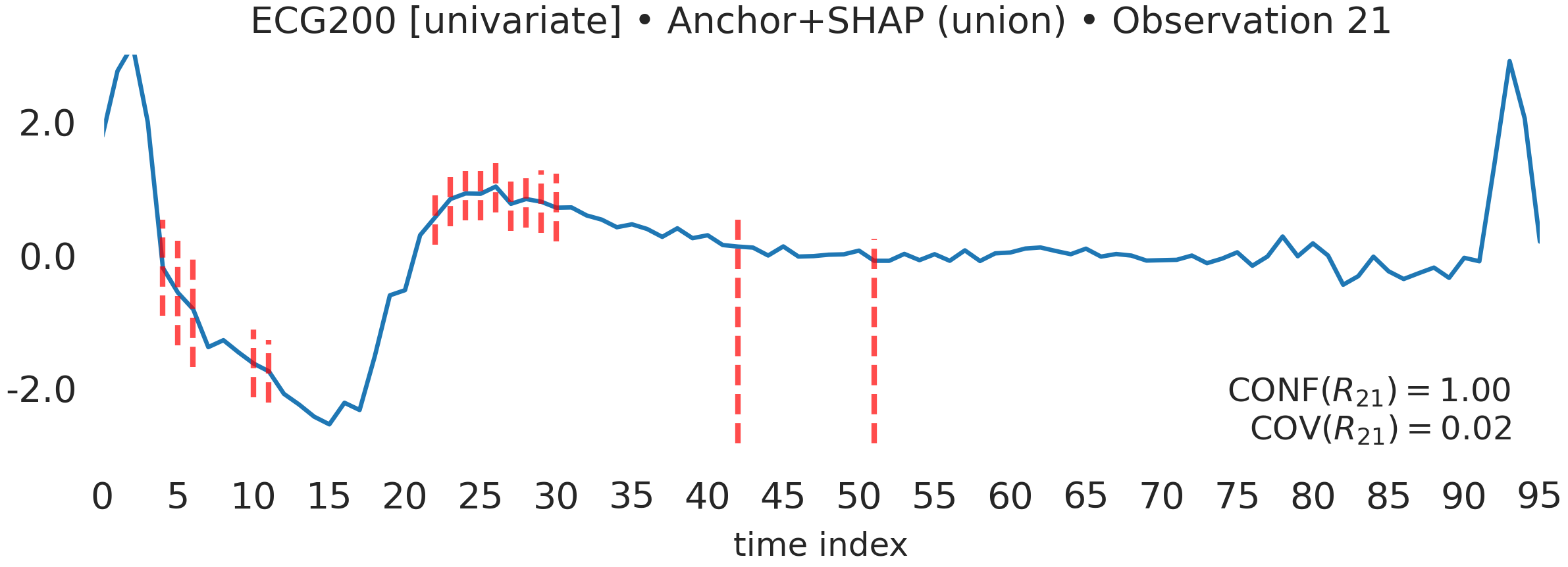}
  }\\[2mm]
  \subfloat[Fusion (lasso): LIME + SHAP]{
    \includegraphics[width=.48\textwidth]{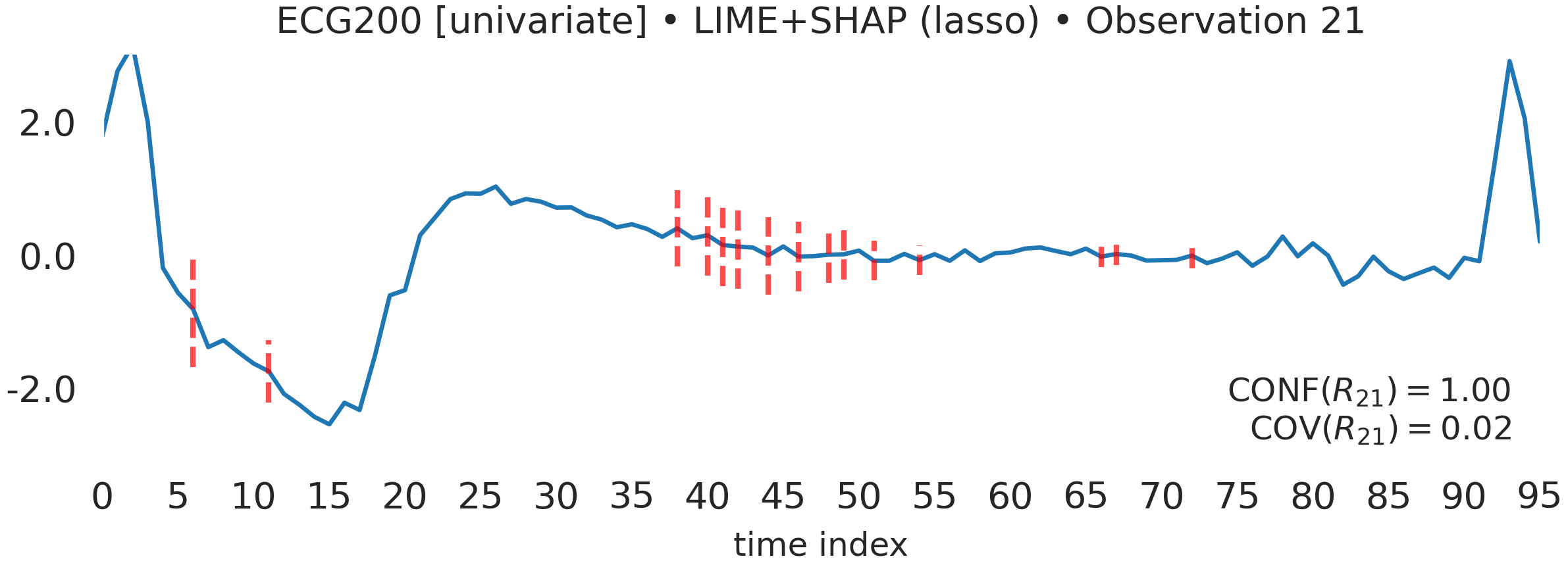}
  }\hfill
  \subfloat[Fusion (union): LIME + SHAP]{
    \includegraphics[width=.48\textwidth]{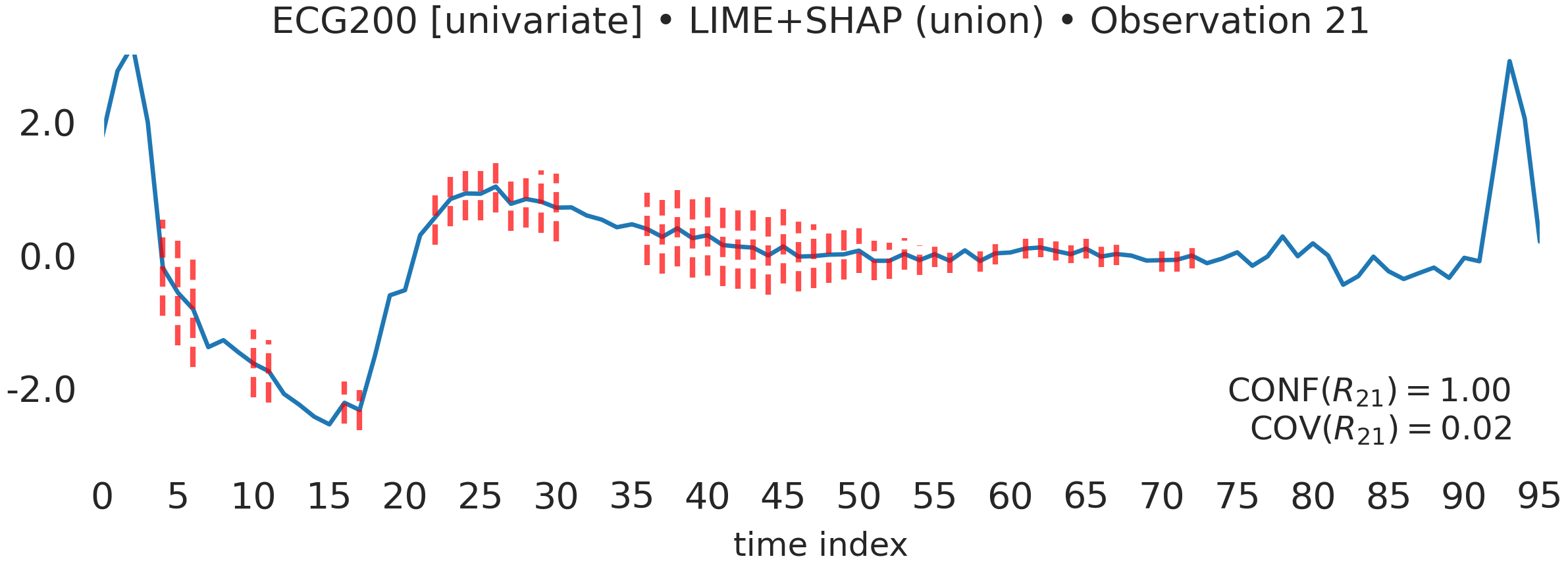}
  }\\[2mm]
  \subfloat[Fusion (best): Anchor + LIME + SHAP]{
    \includegraphics[width=.48\textwidth]{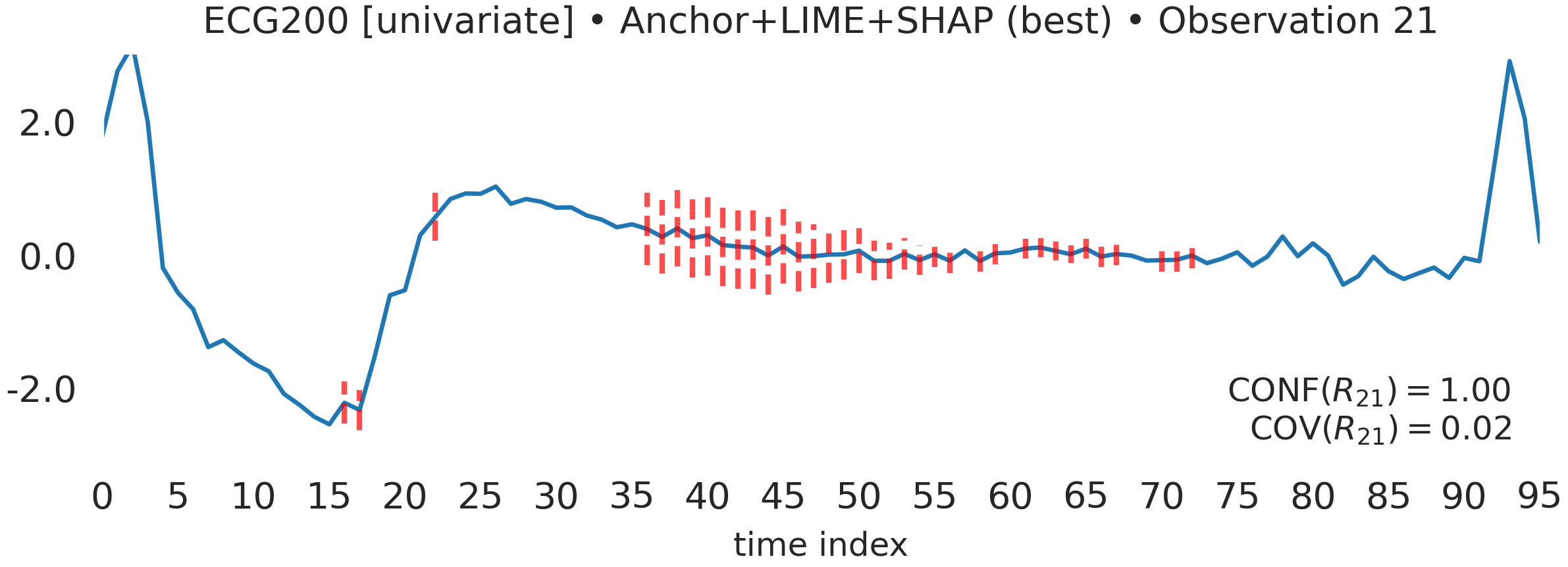}
  }\hfill
  \subfloat[Fusion (lasso): Anchor + LIME + SHAP]{
    \includegraphics[width=.48\textwidth]{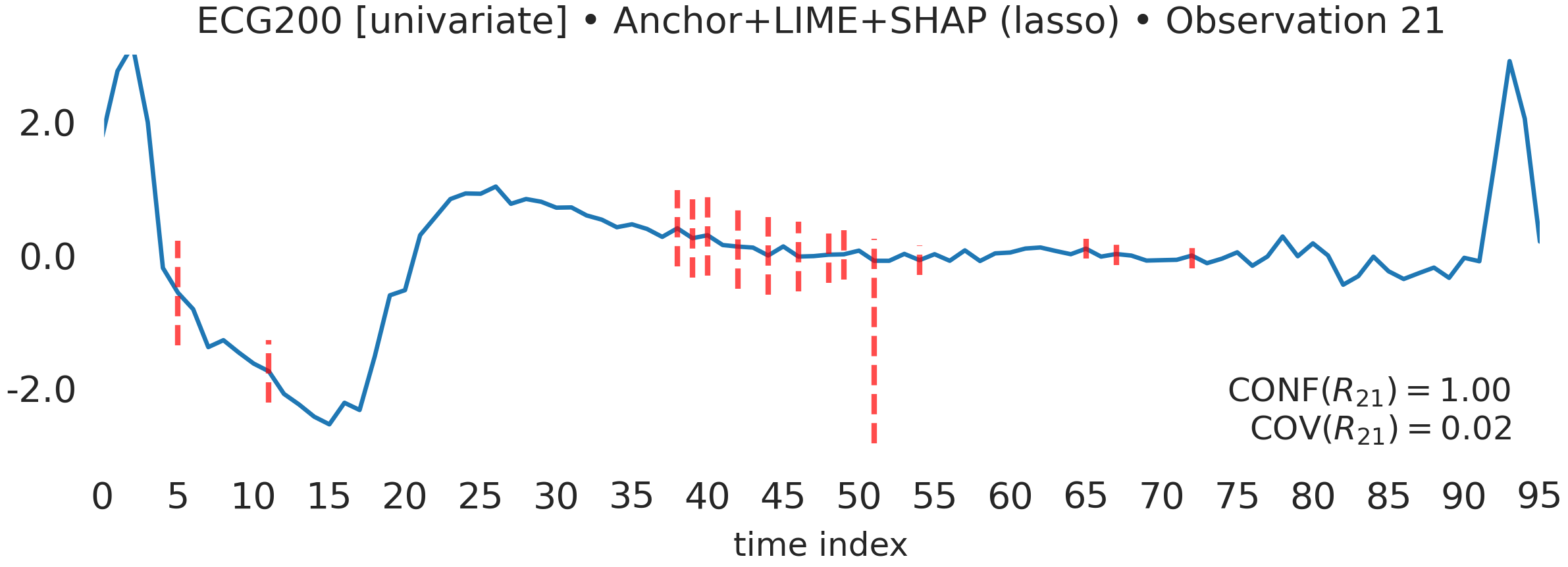}
  }

  \caption{Observation~21 from ECG200 dataset (univariate). 
  Rule intervals produced by single explainers (Anchor, LIME, SHAP) and by various fusion strategies. 
  The blue curve is the signal; red dashed bands mark the value intervals used by rule $R_{21}$ for each method. 
  Confidence and coverage are printed within each panel. 
  Definitions of fusion methods are provided in Sec.~\ref{sec:rule_ensemble}.}
  \label{fig:ecg200_idx21_rulepanels}
\end{figure*}

Figure~\ref{fig:ecg200_idx21_rulepanels} shows the rule intervals for \emph{observation~21}. 
The blue curve is the signal. 
Red dashed segments mark the time intervals required by each rule \(R_{21}\). 
The numbers in each panel report confidence and coverage on the test set. 
Panels~(a)--(c) present single-explainer benchmarks: \emph{Anchor}, \emph{LIME}, and \emph{SHAP}. 
Panels~(d) and~(f) show \emph{union} fusions: \emph{Anchor+SHAP} and \emph{LIME+SHAP}. 
Panels~(e) and~(h) show \emph{lasso} fusions: \emph{LIME+SHAP} and \emph{Anchor+LIME+SHAP}. 
Panel~(g) reports the \emph{best} selection over all explainers. 
We next discuss each panel in turn.

\emph{(a) Anchor (benchmark).} Two semi-infinite thresholds at \(t_{42}\) and \(t_{51}\) localise the decision in the mid/late part of the trace. 
The rule is extremely compact (high confidence, moderate coverage), but the one-sided nature of the thresholds provides little temporal extent, which limits shape specificity. 

\emph{(b) LIME (benchmark).} Many short intervals concentrate in the mid section (\(\approx\!t_{36}\)–\(t_{56}\)) and late flat portion, and LIME also flags the early inflection around \(t_{16}\)–\(t_{17}\) and another change near \(t_{22}\). 
This pattern closely tracks local shape variations and attains full confidence at low coverage.

\emph{(c) SHAP (benchmark).} SHAP emphasises the initial drop (\(t_{4}\)–\(t_{6}\), \(t_{10}\)–\(t_{11}\)) and the local peak/plateau around \(t_{22}\)–\(t_{30}\). 
It therefore complements LIME by assigning more weight to the early and early-mid morphology.

\emph{(d) Fusion (union): Anchor+SHAP.} The union operator pools all conditions from both inputs into a single conjunctive rule. 
In this instance the union has the same confidence and coverage as SHAP because every test case that satisfies SHAP’s intervals already satisfies Anchor’s mid/late thresholds, so the added Anchor conditions are redundant and the covered set is unchanged. 
Since adding conjunctive conditions cannot increase coverage and does not decrease it here, confidence also remains the same. 
The extra Anchor intervals mainly make the mid section explicit in the panel. 

\emph{(e) Fusion (lasso): LIME+SHAP.} Lasso reduces cognitive load by pruning overlapping or lower-priority intervals under the \(\ell_1\) selection criterion. 
The retained set remains balanced: it keeps informative early (from SHAP), mid (from both), and late segments (from LIME), yielding a concise yet representative rule.

\emph{(f) Fusion (union): LIME+SHAP.} Pooling both explainers yields the most extensive set of intervals along the trace—dense mid/late coverage from LIME together with SHAP’s early emphasis. 
This gives a simple reading: early evidence is provided by SHAP and mid/late evidence by LIME, allowing a practitioner to probe \emph{"WHAT-IF"} edits across substantial part of waveform. 
Adding intervals in a conjunctive rule can maintain or reduce coverage, but not inflate it.

\emph{(g) Fusion (best): Anchor+LIME+SHAP.} For this observation, the \emph{best} selector returns exactly the \emph{LIME (benchmark)} rule. 
LIME achieves the highest confidence at the attained coverage. 
In this case SHAP attains the same metrics, but our deterministic tie-break (fixed explainer order) selects LIME, so SHAP is not chosen.

\emph{(h) Fusion (lasso): Anchor+LIME+SHAP.} Lasso selects a sparse subset drawn from \emph{all three} sources, without favouring a single explainer. 
The resulting rule captures dispersed early, mid and late segments while remaining sparse (compact). 

To sum up, Anchor provides the shortest and most readable rule, but it is less useful in visual analysis: its one-sided, semi-infinite thresholds highlight isolated time indices rather than contiguous segments, so they convey little about waveform duration or shape. 
LIME and SHAP emphasise different phases (mid/late vs.\ early/early-mid). 
Union fusions offer the broadest spatial map of supporting intervals, whereas lasso fusions distil them to a compact subset. 
The \emph{best} fusion method returns a single rule according to the stated priorities; in this case the selected rule is LIME, tied with SHAP on the metrics. 

\subsubsection{Counterfactuals: testing rule boundaries}
Counterfactuals help assess whether an instance-level rule genuinely separates classes in ways that are meaningful to practitioners. 
In our setting, a \emph{counterfactual (CF)} is defined operationally: it is a \emph{test} observation drawn from the \emph{opposite} class and overlaid as a dotted trace. 
This construction is diagnostic rather than causal: it provides a direct visual contrast that lets experts see where and how the rule would be violated by an instance that should be rejected. 

We use CFs to clarify what the rule demands and which time segments carry the decision. 
All series are shown in the same normalised scale. 
Red dashed segments indicate the admissible value ranges at specific, constrained time indices~$t$; these may be two-sided intervals or one-sided thresholds. 
The blue curve (target observation) satisfies all marked ranges. 
The grey dotted CF does not: whenever the dotted trace exits a red segment at a constrained~$t$, the rule is broken and the CF remains in the opposite class. 
A small box on each panel reports the rule’s $\mathrm{CONF}$ and $\mathrm{COV}$, which contextualise how reliable and how frequent the rule is across the test set. 
This combination—contrast against a CF, explicit intervals, and summary metrics—yields a compact, expert-friendly reading of the decision boundary. 
\begin{figure*}[t]
  \centering

  \subfloat[Anchor (benchmark): obs.\ 21 vs.\ CF 38]{
    \includegraphics[width=0.72\textwidth]{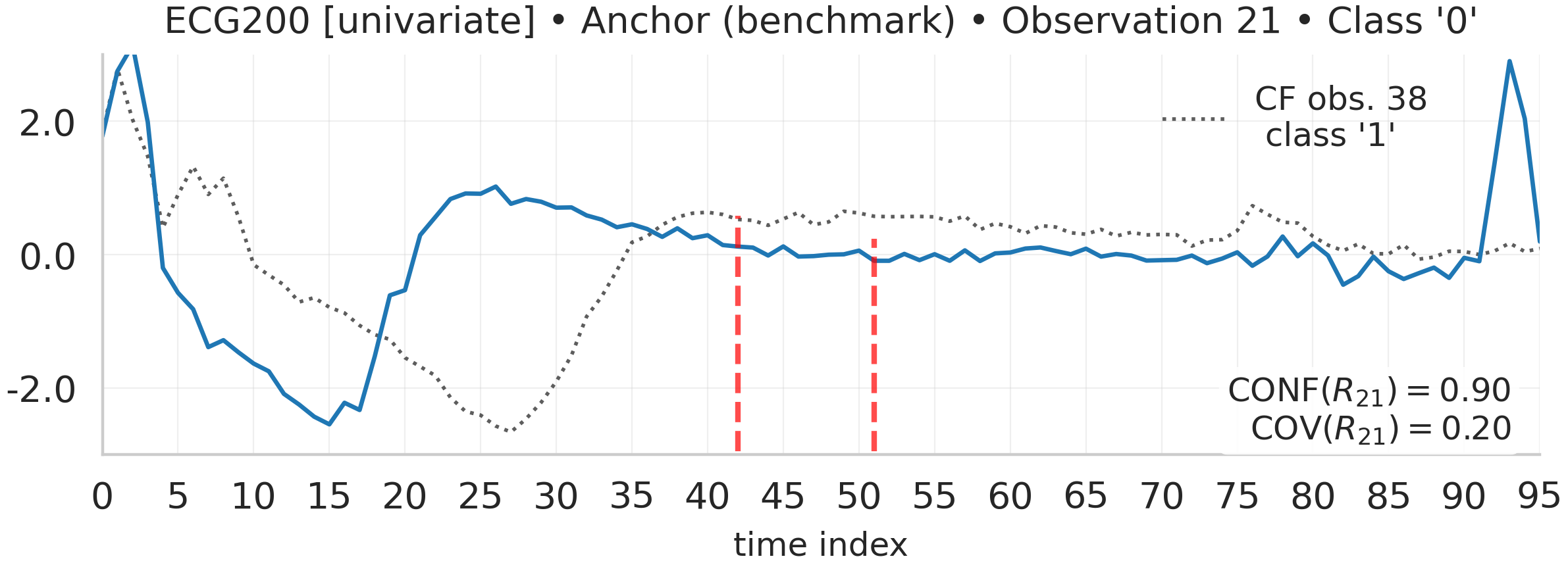}
  }\\[+0.25em]

  \subfloat[SHAP (benchmark): obs.\ 21 vs.\ CF 3]{
    \includegraphics[width=0.72\textwidth]{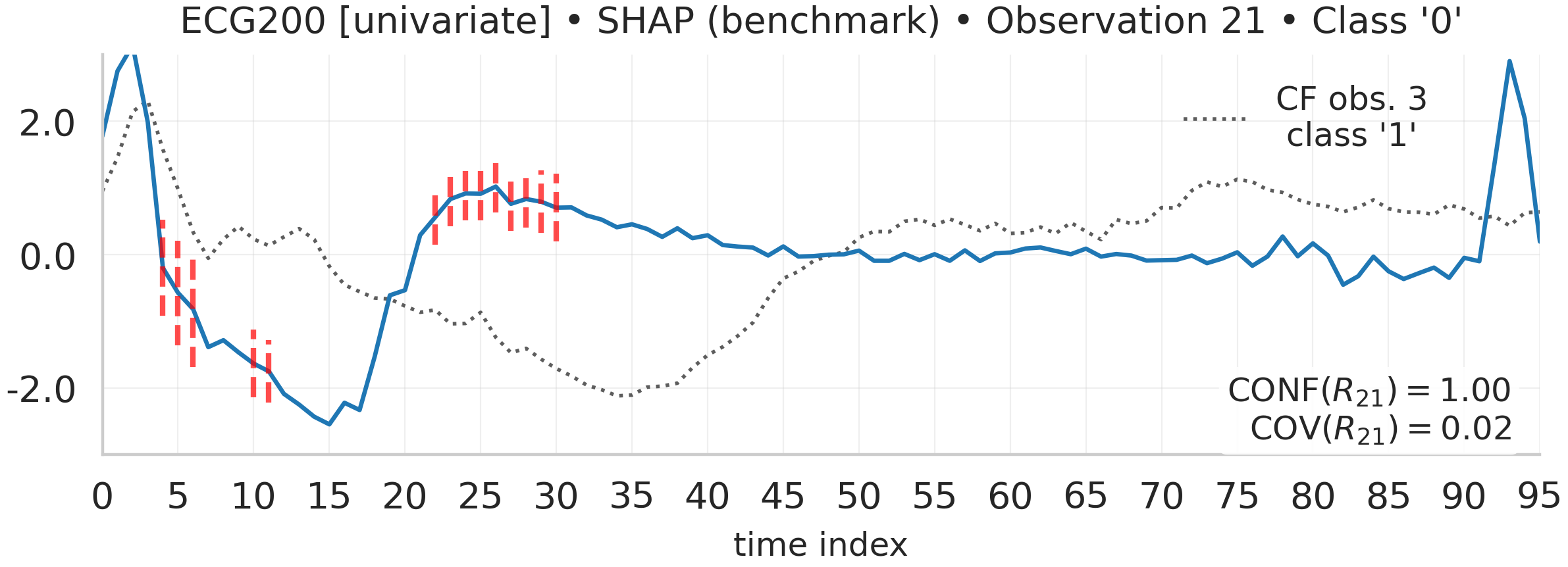}
  }\\[+0.25em]

  \subfloat[LIME+SHAP (lasso): obs.\ 21 vs.\ CF 47]{
    \includegraphics[width=0.72\textwidth]{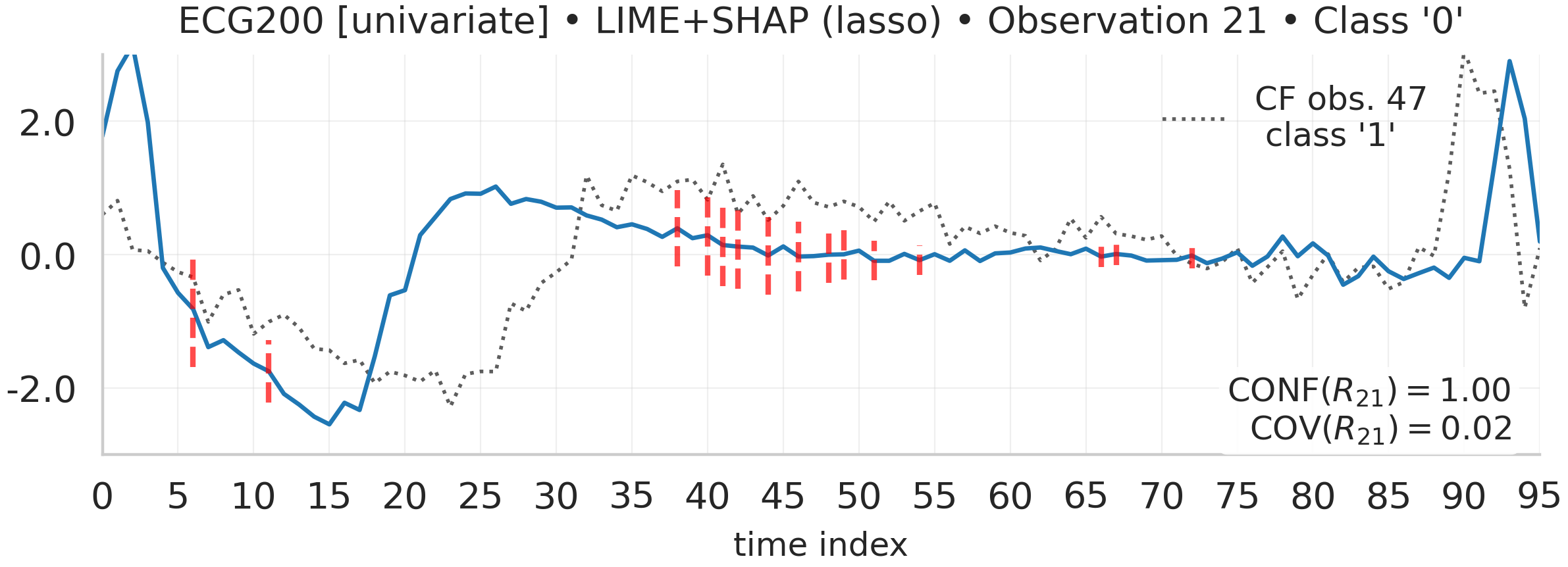}
  }

  \caption{Examples illustrate three Counterfactuals for instance~21: (a) Anchor (benchmark) with CF instance~38; (b) SHAP (benchmark) with CF instance~3; and (c) LIME+SHAP with lasso fusion with CF instance~47.}
  \label{fig:cf-ecg200}
\end{figure*}

The three panels tell a coherent story: CF leaves the admissible bands, which results in a class flip. 
In panel (a) \emph{Anchor (benchmark)} imposes two mid-series intervals, around \(t\!\approx\!43\) and \(t\!\approx\!50\), with narrow bands slightly above zero; the target trace remains inside both, whereas the CF crosses them from above and thus violates the rule. 
In panel (b) \emph{SHAP (benchmark)} concentrates on the early rebound after the initial dip (\(t\!\approx\!20\text{–}27\)) and requires a short positive excursion; the target exhibits this burst, while the CF stays lower and repeatedly exits the band.  
In panel (c) \emph{LIME+SHAP (lasso)} selects a compact sequence of mid-section intervals (\(t\!\approx\!39\text{–}60\)) centred near zero with a slight positive offset; the target stays within these tight corridors, whereas the CF drifts outside at several constrained times.

\subsubsection{Rule listings for the panels}
Guided by the rule intervals in Fig.~\ref{fig:ecg200_idx21_rulepanels} and Fig.~\ref{fig:cf-ecg200}, we now give the corresponding rule statements for panels (a)–(h). 
Each listing follows the same notation and uses standardised \(z\)-scores. 
Presenting visuals before equations aids time-series interpretation: the reader can first locate the relevant phases of the waveform and then verify the precise inequalities.  
Showing the inequalities after the visuals keeps the focus on \emph{where} the decision is anchored and then makes it precise enough to audit or run simple \emph{"WHAT-IF"} perturbations. 
This also facilitates counterfactual checks—one can perturb the signal within a highlighted window to test whether the rule would be violated. 
Panel labels are retained so the reader can cross-reference each rule with its subfigure. 

\begin{multicols}{2}
\raggedcolumns
\setlength{\columnsep}{0.8em}

\begingroup
\scriptsize
\setlength{\abovedisplayskip}{0.35em}
\setlength{\belowdisplayskip}{0.35em}
\setlength{\jot}{0.25em}

\begin{scriptsize}
\paragraph{a) Anchor (benchmark)}
\pharrule{R_{21}}{
& t_{42} \le 0.58\\
& t_{51} \le 0.23
}{
\hat y = \text{"0"},\\
\mathrm{CONF}(R_{21}) = 0.90,\\
\mathrm{COV}(R_{21}) = 0.20.
}
\end{scriptsize}

\begin{scriptsize}
\paragraph{c) SHAP (benchmark)}
\pharrule{R_{21}}{
& t_{4} \in (>-0.92,\, \le 0.52] \\
& t_{5} \in (>-1.36,\, \le 0.21] \\
& t_{6} \in (>-1.69,\, \le 0.04] \\
& t_{10} \in (>-2.15,\, \le -1.13] \\
& t_{11} \in (>-2.22,\, \le -1.28] \\
& t_{22} \in (>0.15,\, \le 0.97] \\
& t_{23} \in (>0.43,\, \le 1.23] \\
& t_{24} \in (>0.51,\, \le 1.31] \\
& t_{25} \in (>0.51,\, \le 1.31] \\
& t_{26} \in (>0.62,\, \le 1.41] \\
& t_{27} \in (>0.35,\, \le 1.17] \\
& t_{28} \in (>0.40,\, \le 1.26] \\
& t_{29} \in (>0.32,\, \le 1.26] \\
& t_{30} \in (>0.20,\, \le 1.21] \\
}{
\hat y = \text{"0"},\\
\mathrm{CONF}(R_{21}) = 1.00,\\
\mathrm{COV}(R_{21}) = 0.02.
}
\end{scriptsize}

\begin{scriptsize}
\paragraph{d) Union of Anchor + SHAP}
\pharrule{R_{21}}{
& t_{4} \in (>-0.92,\, \le 0.52] \\
& t_{5} \in (>-1.36,\, \le 0.21] \\
& t_{6} \in (>-1.69,\, \le 0.04] \\
& t_{10} \in (>-2.15,\, \le -1.13] \\
& t_{11} \in (>-2.22,\, \le -1.28] \\
& t_{22} \in (>0.15,\, \le 0.97] \\
& t_{23} \in (>0.43,\, \le 1.23] \\
& t_{24} \in (>0.51,\, \le 1.31] \\
& t_{25} \in (>0.51,\, \le 1.31] \\
& t_{26} \in (>0.62,\, \le 1.41] \\
& t_{27} \in (>0.35,\, \le 1.17] \\
& t_{28} \in (>0.40,\, \le 1.26] \\
& t_{29} \in (>0.32,\, \le 1.26] \\
& t_{30} \in (>0.20,\, \le 1.21] \\
& t_{42} \le 0.58 \\
& t_{51} \le 0.23 \\
}{
\hat y = \text{"0"},\\
\mathrm{CONF}(R_{21}) = 1.00,\\
\mathrm{COV}(R_{21}) = 0.02.
}
\end{scriptsize}

\begin{scriptsize}
\paragraph{b) LIME (benchmark)}
\pharrule{R_{21}}{
& t_{16} \in (>-2.54,\, \le -1.91] \\
& t_{17} \in (>-2.63,\, \le -2.03] \\
& t_{22} \in (>0.20,\, \le 0.93] \\
& t_{36} \in (>-0.16,\, \le 0.92] \\
& t_{37} \in (>-0.29,\, \le 0.82] \\
& t_{38} \in (>-0.18,\, \le 0.97] \\
& t_{39} \in (>-0.35,\, \le 0.83] \\
& t_{40} \in (>-0.32,\, \le 0.89] \\
& t_{41} \in (>-0.48,\, \le 0.76] \\
& t_{42} \in (>-0.52,\, \le 0.75] \\
& t_{43} \in (>-0.51,\, \le 0.72] \\
& t_{44} \in (>-0.60,\, \le 0.56] \\
& t_{45} \in (>-0.44,\, \le 0.68] \\
& t_{46} \in (>-0.55,\, \le 0.49] \\
& t_{47} \in (>-0.51,\, \le 0.46] \\
& t_{48} \in (>-0.43,\, \le 0.42] \\
& t_{49} \in (>-0.37,\, \le 0.38] \\
& t_{50} \in (>-0.28,\, \le 0.40] \\
& t_{51} \in (>-0.39,\, \le 0.20] \\
& t_{52} \in (>-0.36,\, \le 0.17] \\
& t_{53} \in (>-0.23,\, \le 0.25] \\
& t_{54} \in (>-0.30,\, \le 0.13] \\
& t_{55} \in (>-0.18,\, \le 0.19] \\
& t_{56} \in (>-0.28,\, \le 0.08] \\
& t_{58} \in (>-0.26,\, \le 0.06] \\
& t_{59} \in (>-0.15,\, \le 0.18] \\
& t_{61} \in (>-0.06,\, \le 0.23] \\
& t_{62} \in (>-0.04,\, \le 0.25] \\
& t_{63} \in (>-0.09,\, \le 0.19] \\
& t_{64} \in (>-0.13,\, \le 0.13] \\
& t_{65} \in (>-0.06,\, \le 0.23] \\
& t_{66} \in (>-0.19,\, \le 0.12] \\
& t_{67} \in (>-0.16,\, \le 0.17] \\
& t_{70} \in (>-0.26,\, \le 0.09] \\
& t_{71} \in (>-0.26,\, \le 0.10] \\
& t_{72} \in (>-0.21,\, \le 0.18] \\
}{
\hat y = \text{"0"},\\
\mathrm{CONF}(R_{21}) = 1.00,\\
\mathrm{COV}(R_{21}) = 0.02.
}
\end{scriptsize}

\begin{scriptsize}
\paragraph{f) Union of LIME + SHAP}
\pharrule{R_{21}}{
& t_{4} \in (>-0.92,\, \le 0.52] \\
& t_{5} \in (>-1.36,\, \le 0.21] \\
& t_{6} \in (>-1.69,\, \le 0.04] \\
& t_{10} \in (>-2.15,\, \le -1.13] \\
& t_{11} \in (>-2.22,\, \le -1.28] \\
& t_{16} \in (>-2.54,\, \le -1.91] \\
& t_{17} \in (>-2.63,\, \le -2.03] \\
& t_{22} \in (>0.15,\, \le 0.97] \\
& t_{23} \in (>0.43,\, \le 1.23] \\
& t_{24} \in (>0.51,\, \le 1.31] \\
& t_{25} \in (>0.51,\, \le 1.31] \\
& t_{26} \in (>0.62,\, \le 1.41] \\
& t_{27} \in (>0.35,\, \le 1.17] \\
& t_{28} \in (>0.40,\, \le 1.26] \\
& t_{29} \in (>0.32,\, \le 1.26] \\
& t_{30} \in (>0.20,\, \le 1.21] \\
& t_{36} \in (>-0.16,\, \le 0.92] \\
& t_{37} \in (>-0.29,\, \le 0.82] \\
& t_{38} \in (>-0.18,\, \le 0.97] \\
& t_{39} \in (>-0.35,\, \le 0.83] \\
& t_{40} \in (>-0.32,\, \le 0.89] \\
& t_{41} \in (>-0.48,\, \le 0.76] \\
& t_{42} \in (>-0.52,\, \le 0.75] \\
& t_{43} \in (>-0.51,\, \le 0.72] \\
& t_{44} \in (>-0.60,\, \le 0.56] \\
& t_{45} \in (>-0.44,\, \le 0.68] \\
& t_{46} \in (>-0.55,\, \le 0.49] \\
& t_{47} \in (>-0.51,\, \le 0.46] \\
& t_{48} \in (>-0.43,\, \le 0.42] \\
& t_{49} \in (>-0.37,\, \le 0.38] \\
& t_{50} \in (>-0.28,\, \le 0.40] \\
& t_{51} \in (>-0.39,\, \le 0.20] \\
& t_{52} \in (>-0.36,\, \le 0.17] \\
& t_{53} \in (>-0.23,\, \le 0.25] \\
& t_{54} \in (>-0.30,\, \le 0.13] \\
& t_{55} \in (>-0.18,\, \le 0.19] \\
& t_{56} \in (>-0.28,\, \le 0.08] \\
& t_{58} \in (>-0.26,\, \le 0.06] \\
& t_{59} \in (>-0.15,\, \le 0.18] \\
& t_{61} \in (>-0.06,\, \le 0.23] \\
& t_{62} \in (>-0.04,\, \le 0.25] \\
& t_{63} \in (>-0.09,\, \le 0.19] \\
& t_{64} \in (>-0.13,\, \le 0.13] \\
& t_{65} \in (>-0.06,\, \le 0.23] \\
& t_{66} \in (>-0.19,\, \le 0.12] \\
& t_{67} \in (>-0.16,\, \le 0.17] \\
& t_{70} \in (>-0.26,\, \le 0.09] \\
& t_{71} \in (>-0.26,\, \le 0.10] \\
& t_{72} \in (>-0.21,\, \le 0.18] \\
}{
\hat y = \text{"0"},\\
\mathrm{CONF}(R_{21}) = 1.00,\\
\mathrm{COV}(R_{21}) = 0.02.
}
\end{scriptsize}

\begin{scriptsize}
\paragraph{g) Best of Anchor + LIME + SHAP}
\pharrule{R_{21}}{
& t_{16} \in (>-2.54,\, \le -1.91] \\
& t_{17} \in (>-2.63,\, \le -2.03] \\
& t_{22} \in (>0.20,\, \le 0.93] \\
& t_{36} \in (>-0.16,\, \le 0.92] \\
& t_{37} \in (>-0.29,\, \le 0.82] \\
& t_{38} \in (>-0.18,\, \le 0.97] \\
& t_{39} \in (>-0.35,\, \le 0.83] \\
& t_{40} \in (>-0.32,\, \le 0.89] \\
& t_{41} \in (>-0.48,\, \le 0.76] \\
& t_{42} \in (>-0.52,\, \le 0.75] \\
& t_{43} \in (>-0.51,\, \le 0.72] \\
& t_{44} \in (>-0.60,\, \le 0.56] \\
& t_{45} \in (>-0.44,\, \le 0.68] \\
& t_{46} \in (>-0.55,\, \le 0.49] \\
& t_{47} \in (>-0.51,\, \le 0.46] \\
& t_{48} \in (>-0.43,\, \le 0.42] \\
& t_{49} \in (>-0.37,\, \le 0.38] \\
& t_{50} \in (>-0.28,\, \le 0.40] \\
& t_{51} \in (>-0.39,\, \le 0.20] \\
& t_{52} \in (>-0.36,\, \le 0.17] \\
& t_{53} \in (>-0.23,\, \le 0.25] \\
& t_{54} \in (>-0.30,\, \le 0.13] \\
& t_{55} \in (>-0.18,\, \le 0.19] \\
& t_{56} \in (>-0.28,\, \le 0.08] \\
& t_{58} \in (>-0.26,\, \le 0.06] \\
& t_{59} \in (>-0.15,\, \le 0.18] \\
& t_{61} \in (>-0.06,\, \le 0.23] \\
& t_{62} \in (>-0.04,\, \le 0.25] \\
& t_{63} \in (>-0.09,\, \le 0.19] \\
& t_{64} \in (>-0.13,\, \le 0.13] \\
& t_{65} \in (>-0.06,\, \le 0.23] \\
& t_{66} \in (>-0.19,\, \le 0.12] \\
& t_{67} \in (>-0.16,\, \le 0.17] \\
& t_{70} \in (>-0.26,\, \le 0.09] \\
& t_{71} \in (>-0.26,\, \le 0.10] \\
& t_{72} \in (>-0.21,\, \le 0.18] \\
}{
\hat y = \text{"0"},\\
\mathrm{CONF}(R_{21}) = 1.00,\\
\mathrm{COV}(R_{21}) = 0.02.
}
\end{scriptsize}

\begin{scriptsize}
\paragraph{e) Lasso of LIME + SHAP}
\pharrule{R_{21}}{
& t_{6} \in (>-1.69,\, \le 0.04] \\
& t_{11} \in (>-2.22,\, \le -1.28] \\
& t_{38} \in (>-0.18,\, \le 0.97] \\
& t_{40} \in (>-0.32,\, \le 0.89] \\
& t_{41} \in (>-0.48,\, \le 0.76] \\
& t_{42} \in (>-0.52,\, \le 0.75] \\
& t_{44} \in (>-0.60,\, \le 0.56] \\
& t_{46} \in (>-0.55,\, \le 0.49] \\
& t_{48} \in (>-0.43,\, \le 0.42] \\
& t_{49} \in (>-0.37,\, \le 0.38] \\
& t_{51} \in (>-0.39,\, \le 0.20] \\
& t_{54} \in (>-0.30,\, \le 0.13] \\
& t_{66} \in (>-0.19,\, \le 0.12] \\
& t_{67} \in (>-0.16,\, \le 0.17] \\
& t_{72} \in (>-0.21,\, \le 0.18] \\
}{
\hat y = \text{"0"},\\
\mathrm{CONF}(R_{21}) = 1.00,\\
\mathrm{COV}(R_{21}) = 0.02.
}
\end{scriptsize}

\begin{scriptsize}
\paragraph{h) Lasso of Anchor + LIME + SHAP}
\pharrule{R_{21}}{
& t_{5} \in (>-1.36,\, \le 0.21] \\
& t_{11} \in (>-2.22,\, \le -1.28] \\
& t_{38} \in (>-0.18,\, \le 0.97] \\
& t_{39} \in (>-0.35,\, \le 0.83] \\
& t_{40} \in (>-0.32,\, \le 0.89] \\
& t_{42} \in (>-0.52,\, \le 0.75] \\
& t_{44} \in (>-0.60,\, \le 0.56] \\
& t_{46} \in (>-0.55,\, \le 0.49] \\
& t_{48} \in (>-0.43,\, \le 0.42] \\
& t_{49} \in (>-0.37,\, \le 0.38] \\
& t_{51} \le 0.23 \\
& t_{54} \in (>-0.30,\, \le 0.13] \\
& t_{65} \in (>-0.06,\, \le 0.23] \\
& t_{67} \in (>-0.16,\, \le 0.17] \\
& t_{72} \in (>-0.21,\, \le 0.18] \\
}{
\hat y = \text{"0"},\\
\mathrm{CONF}(R_{21}) = 1.00,\\
\mathrm{COV}(R_{21}) = 0.02.
}
\end{scriptsize}

\endgroup
\end{multicols}

\subsubsection{Dataset-level visual summary (ECG200 and Plane datasets)}
\label{sec:visual_catalog_pointer}
Beyond the single-trace analysis in Sec.~\ref{sec:qual_eval}, Appendix~\ref{app:visual_appendix} presents a dataset-level visual catalogue for \emph{ECG200} (Appendix~\ref{app:ecg200_visual_catalog}) and \emph{Plane} (Appendix~\ref{app:plane_visual_catalog}). 
For each explainer (Anchor, LIME, SHAP) and their fusions (union, lasso, best), we show curated per-class exemplars (diverse instances, prototype, class mean) with the same red-window overlay used throughout. 
This complements the pointwise analysis by highlighting recurrent early/mid/late concentrations at the dataset level.

\FloatBarrier

\section{Conclusion}\label{sec:summary}
This work introduces a novel rule-based explainability framework tailored for \emph{time series (TS)} classification tasks, addressing the current gap in interpretable AI solutions for temporal data. 
To the best of our knowledge, this is the first systematic study transforming \emph{post-hoc}, \emph{instance-based numerical feature attributions} from \emph{SHAP} and \emph{LIME} explainers into structured, human-readable rule representations for \emph{TS}. 
Our approach uniquely integrates and compares these rule-based explanations with \emph{Anchor}, a native rule-based explainer, across a comprehensive benchmark of 43 \emph{TS} datasets from the \emph{UCR/UEA Time Series Classification Archive}. 
We further contribute a modular \emph{rule fusion} pipeline, combining multiple explainers into a single, optimized rule per instance while ensuring high fidelity and coverage. 
This work is accompanied by 
evaluation of multiple rule aggregation strategies, along with an open-source codebase and practical visualization techniques to facilitate adoption in real-world applications.

We transform \emph{numeric feature attribution explainers} (\emph{LIME}, \emph{SHAP}) into \emph{rule-based TS explanations}, integrating them with \emph{Anchor}. 
Our post-hoc framework explains instances post-training, crucial for TS tasks like anomaly detection and decision support (\texttt{RQ1}). 
\emph{Feature-to-rule conversion} balances complexity, improving \emph{coverage}, \emph{confidence}, and ensuring interpretability for each instance.
Optimizing thresholds and perturbations refines structured, unambiguous rules, improving interpretability. 
Reliable rules align with model decisions, balancing \emph{confidence} and \emph{coverage} (\texttt{RQ2}).

\emph{SHAP} and \emph{LIME} perform comparably to \emph{Anchor}, despite the latter being \emph{natively rule-based}. 
While \emph{Anchor} incurs high computational cost on long \emph{TS}, converting \emph{SHAP} attributions into interval rules provides a more efficient alternative that achieves competitive performance when optimized via \texttt{fusion}. 
Moreover, rules derived from our \emph{numeric-to-rules transformation} use closed intervals \((\ell, u]\) for each timestamp-channel condition, whereas \emph{Anchor} often produces one-sided bounds. 
Closed intervals are especially important in \emph{TS}, as they precisely delimit both lower and upper thresholds, ensuring that temporal segments are fully specified, reducing ambiguity in sequential explanations, and allowing for clearer visualization.  

Our framework consolidates knowledge from multiple explainers (\texttt{RQ3}). 
Unlike bagging or boosting, our \emph{post-hoc fusion} refines single or multiple explainers into a coherent rule set, aligning with \emph{expert system principles}. 
\emph{Fusion} improves rule reliability, reduces ambiguity, and enhances decision-making. 
Moreover, \emph{PHAR} supports \emph{sparsity optimization}. 
We measure sparsity with the rule-length metric \(F(n)\) and report both its mean and median. 
This is important given human working-memory limits (the 7\,$\pm$\,2 principle)~\citep{Miller1956}, as discussed earlier when defining the objective function. 
Our method enables this, when desired, by offering resolution modes that promote concise, non-overlapping rules, and by using fusion methods such as lasso and intersection. 
This is also illustrated by the ECG200 example discussed in our qualitative evaluation.

Rules enhance \emph{TS} visualization, aiding expert interpretation (\texttt{RQ4}). 
\emph{Anchor} uses fewer features but produces broad intervals, missing finer temporal details. 
\emph{SHAP} and \emph{LIME} involve more features, creating narrower, 
potentially more human-readable intervals that retain predictive alignment while avoiding redundant or overlapping conditions. 
Application-dependent, broader feature sets or refined intervals enhance explanation resolution.

Interpretability of \emph{fused rules} depends on simplicity (fewer features) and uniqueness (avoiding redundancy), optimized via key metrics. 
\emph{Rule fusion} removes redundancy, ensuring a structured, coherent set of explanations. 
Results show that both single-explainer (\emph{SHAP}, \emph{LIME}) and fusion methods achieve high \emph{confidence}, \emph{coverage}, and \emph{explanation ratio}. 
\texttt{Lasso}, \texttt{Weighted}, and \texttt{Best} refine rules, often improving clarity and consistency,  
while visualization enhances the illustration of rule applicability.

\subsection{Limitations}
\label{sec:limitations}

The \emph{PHAR} framework effectively transforms numeric feature attributions into rule-based explanations, but several limitations remain.
\begin{enumerate}
\item Rule quality within \emph{PHAR} depends on explainers such as \emph{SHAP} and \emph{LIME}, which rely on perturbation-based sampling and may produce unstable attributions, particularly in noisy or highly dynamic time series. 
\item Computational complexity represents a secondary limitation; while the transformation step in \emph{PHAR} involves hyperparameter tuning, we have shown in Section~\ref{sec:computational_efficiency} that the practical cost remains acceptable due to parallelization, the optional nature of tuning, and the complexity of the \texttt{fusion} step, with the exception of the more computationally intensive \texttt{Lasso global} variant. 
\item Our evaluation of \emph{PHAR} employed a predefined set of metrics, and further exploration of alternative rule selection criteria could enhance the flexibility of rule generation. 
\item Interpretability within \emph{PHAR} may be challenged in high-dimensional time series, where resulting rules risk becoming overly specific; however, this issue is partially mitigated by the dedicated visualization techniques developed in \emph{PHAR}, which follow a semifactual design and aim to present rule intervals in a clearer, more interpretable manner. 
Future work could further address this limitation by incorporating feature selection mechanisms or additional constraints to promote more generalizable rule representations. 
\item While our visualization experiments within \emph{PHAR} demonstrated practical applicability, we did not conduct evaluations with \emph{Domain Experts}, and the relationship between rule complexity and perceived clarity remains to be systematically investigated. 
\item Finally, although \texttt{fusion} in \emph{PHAR} balances rule importance, it does not fully eliminate potential conflicts between different explainers. 
Additionally, we did not explicitly analyze rule similarity or stability, which could be particularly relevant in time series contexts where neighboring time steps may influence model predictions in a correlated manner. 
\end{enumerate}

\subsection{Future work}
\label{sec:future_work}

While our framework effectively transforms numeric feature attributions into rule-based explanations, several directions remain open. 
\begin{enumerate}
\item One of the important directions is extending \emph{PHAR} to \emph{counterfactual explanation} generation, by identifying and presenting the nearest time series belonging to an alternative class. 
This would allow \emph{Domain Experts} to specify transitions from the current predicted class to any arbitrary target class (e.g. in predictive maintenance or clinical decision support), thereby providing actionable \emph{"WHAT-IF"} insights. 
\item Further research should assess rule stability across similar instances, determining whether minor input variations lead to consistent explanations. 
\item Integrating expert feedback could refine rule usability, bridging automated rule extraction with human interpretability. 
One approach is incorporating symbolic reasoning systems like \emph{HeaRTDroid}~\citep{heartdroid_2019}~\footnote{\url{https://heartdroid.re/}}, a lightweight rule engine for expert systems, enabling a hybrid approach that merges statistical learning with structured decision-making. 
\item For improvement of interpretability, it will be worth investigating the trade-off between feature count and interval width.  
\item Usage of advanced visualization methods, such as interactive overlays or multi-dimensional interval representations could be beneficial. 
Related direction is \emph{DeepVATS}~\citep{RODRIGUEZFERNANDEZ2023110793}~\footnote{\url{https://github.com/vrodriguezf/deepvats}}, which integrates DL with interactive visualization of latent space projections for \emph{TS}, similar to \emph{TensorFlow Embeddings Projector}, aiding in anomaly detection and structure analysis. 
\item Finally, expanding the framework beyond \emph{TS classification} to areas like sequential decision-making or tabular datasets with temporal dependencies could validate its broader applicability.
\end{enumerate}

\backmatter

\bmhead{Supplementary information}
Additional tables supporting the statistical comparison of rule-based explainers and fusion methods via Critical Difference Diagrams are provided in "Supplementary Materials" (Section~\ref{sec:supplement}).  

\bmhead{Acknowledgements} \label{sec:ack}
This article is part of a project that has received funding from the European Union’s Horizon Europe Research and Innovation Programme under Grant Agreement No.\ 101120406. 
The article reflects only the authors’ view and the European Commission is not responsible for any use that may be made of the information contained herein. 

The contribution of Maciej Mozolewski for the research for this publication has been supported by a grant from the Priority Research Area (DigiWorld) under the Mark Kac Center for Complex Systems Research Strategic Programme Excellence Initiative at Jagiellonian University. 

We thank all contributors of open-source libraries (e.g. TensorFlow, Optuna) and the maintainers of the \emph{UCR/UEA Time Series Classification Archive} for curating and providing the datasets. 
We used OpenAI's GPT to assist in writing and refining the text and generating code snippets, while all conceptual ideas and research contributions were developed by the authors.

\bmhead{Declaration of Generative AI and AI-Assisted Technologies in the Manuscript Preparation Process}\label{sec:declaration_of_generative_ai}
During the preparation of this manuscript, the authors used \emph{ChatGPT} to refine the language, improve the readability, and check grammatical consistency, and to draft selected code snippets, typing, and docstrings. 
All such suggestions and code were reviewed, tested, and, when needed, rewritten by the authors, who assume full responsibility for the final text and software. 
\emph{ChatGPT} did not generate any numerical results, metrics, tables, or figures; all results were obtained by running our own code on the described data and models. 

\bibliography{sn-bibliography}

\newpage
\begin{appendices}
\section{\texorpdfstring{}{Appendix A}}\label{sec:appendix}
\FloatBarrier
\subsection{Additional Tables}\label{sec:tables}

\subsubsection{Hyperparameter for Numeric-to-Rules Transformation}
\begin{table*}[ht]
\centering
\caption{Hyperparameter Ranges for Optimization}
\label{tab:hyperparameters}
\begin{tabular}{l p{3cm} p{7cm}}
    \toprule
    \textbf{Hyperparam.} & \textbf{Range/Values} & \textbf{Description} \\ 
    \midrule
    \(p\) & 50 to 99 (step=1) & Percentile for determining feature importance thresholds. \\ 
    \addlinespace
    \(\delta\) & \{True, False\} & Global vs. per-feature threshold application. \\ 
    \addlinespace
    \(\sigma_p\) & 0.01 to 1.0 (continuous) & Scale of perturbation for interval estimation. \\ 
    \addlinespace
    \(N_p\) & 1,000 to 10,000 (step=1,000) & Number of perturbed samples for interval estimation and confidence evaluation. \\ 
    \bottomrule
\end{tabular}
\begin{tablenotes}
\footnotesize
\item \(p\) - \emph{threshold percentile},
\(\delta\) - \emph{global importance indicator},
\(\sigma_p\) - \emph{perturbation scale},
\(N_p\) - \emph{perturbation samples count}
\end{tablenotes}
\end{table*}

\clearpage
\begin{sidewaystable}[ht]
\subsubsection{Experimental Time Series, Explainers and Fused Rule Sets}
\centering
\scriptsize
\resizebox{0.9\textwidth}{!}{
\begin{minipage}{\textwidth}
\begin{tabular}{lcccccc|cccccc|ccc|ccccccc}
\multicolumn{7}{c}{}
& \multicolumn{6}{c}{\textbf{Methods}}
& \multicolumn{3}{c}{\textbf{Baselines}}
& \multicolumn{7}{c}{\textbf{Fused rule sets}} \\
\textbf{Dataset}
& \textbf{Type}
& \textbf{Acc}
& \textbf{\#F}
& \textbf{Dim}
& \textbf{\#B}
& \textbf{\#E}
& \rotatebox{90}{Best}
& \rotatebox{90}{Union}
& \rotatebox{90}{Lasso global}
& \rotatebox{90}{Lasso}
& \rotatebox{90}{Weighted}
& \rotatebox{90}{Intersection}
& \rotatebox{90}{Anchor}
& \rotatebox{90}{SHAP}
& \rotatebox{90}{LIME}
& \rotatebox{90}{Anchor + SHAP}
& \rotatebox{90}{Anchor + LIME}
& \rotatebox{90}{LIME + SHAP}
& \rotatebox{90}{Anchor + LIME + SHAP} \\
\toprule

ArticularyWordRecognition & m & 96,53 & 1296 & $144 \times 9$ & 2 & 12 & \ding{51} & \ding{51} & \ding{51} & \ding{51} & \ding{51} & \ding{51} &  & \ding{51} & \ding{51} &  & \ding{51} & \ding{51} &  &  & \ding{51} &  \\
AtrialFibrillation & m & 37,50 & 1280 & $640 \times 2$ & 2 & 12 & \ding{51} & \ding{51} & \ding{51} & \ding{51} & \ding{51} & \ding{51} &  & \ding{51} & \ding{51} &  & \ding{51} & \ding{51} &  &  & \ding{51} &  \\
BasicMotions & m & 100,00 & 600 & $100 \times 6$ & 2 & 12 & \ding{51} & \ding{51} & \ding{51} & \ding{51} & \ding{51} & \ding{51} &  & \ding{51} & \ding{51} &  & \ding{51} & \ding{51} &  &  & \ding{51} &  \\
Cricket & m & 93,33 & 7182 & $1197 \times 6$ & 2 & 12 & \ding{51} & \ding{51} & \ding{51} & \ding{51} & \ding{51} & \ding{51} &  & \ding{51} & \ding{51} &  & \ding{51} & \ding{51} &  &  & \ding{51} &  \\
ERing & m & 98,67 & 260 & $65 \times 4$ & 2 & 12 & \ding{51} & \ding{51} & \ding{51} & \ding{51} & \ding{51} & \ding{51} &  & \ding{51} & \ding{51} &  & \ding{51} & \ding{51} &  &  & \ding{51} &  \\
Epilepsy & m & 91,30 & 618 & $206 \times 3$ & 2 & 12 & \ding{51} & \ding{51} & \ding{51} & \ding{51} & \ding{51} & \ding{51} &  & \ding{51} & \ding{51} &  & \ding{51} & \ding{51} &  &  & \ding{51} &  \\
Libras & m & 63,33 & 90 & $45 \times 2$ & 1 & 3 &  &  & \ding{51} & \ding{51} & \ding{51} &  & \ding{51} &  &  & \ding{51} &  &  &  &  &  &  \\
PenDigits & m & 99,24 & 16 & $8 \times 2$ & 1 & 3 &  &  & \ding{51} & \ding{51} & \ding{51} &  & \ding{51} &  &  & \ding{51} &  &  &  &  &  &  \\
Adiac & u & 29,59 & 176 & $176 \times 1$ & 2 & 12 & \ding{51} & \ding{51} & \ding{51} & \ding{51} & \ding{51} & \ding{51} &  & \ding{51} & \ding{51} &  & \ding{51} & \ding{51} &  &  & \ding{51} &  \\
BME & u & 86,67 & 128 & $128 \times 1$ & 2 & 12 & \ding{51} & \ding{51} & \ding{51} & \ding{51} & \ding{51} & \ding{51} &  & \ding{51} & \ding{51} &  & \ding{51} & \ding{51} &  &  & \ding{51} &  \\
Beef & u & 53,33 & 470 & $470 \times 1$ & 2 & 12 & \ding{51} & \ding{51} & \ding{51} & \ding{51} & \ding{51} & \ding{51} &  & \ding{51} & \ding{51} &  & \ding{51} & \ding{51} &  &  & \ding{51} &  \\
BeetleFly & u & 80,00 & 512 & $512 \times 1$ & 2 & 12 & \ding{51} & \ding{51} & \ding{51} & \ding{51} & \ding{51} & \ding{51} &  & \ding{51} & \ding{51} &  & \ding{51} & \ding{51} &  &  & \ding{51} &  \\
BirdChicken & u & 50,00 & 512 & $512 \times 1$ & 2 & 12 & \ding{51} & \ding{51} & \ding{51} & \ding{51} & \ding{51} & \ding{51} &  & \ding{51} & \ding{51} &  & \ding{51} & \ding{51} &  &  & \ding{51} &  \\
CBF & u & 100,00 & 128 & $128 \times 1$ & 2 & 12 & \ding{51} & \ding{51} & \ding{51} & \ding{51} & \ding{51} & \ding{51} &  & \ding{51} & \ding{51} &  & \ding{51} & \ding{51} &  &  & \ding{51} &  \\
Chinatown & u & 98,90 & 24 & $24 \times 1$ & 3 & 33 & \ding{51} & \ding{51} & \ding{51} & \ding{51} & \ding{51} & \ding{51} & \ding{51} & \ding{51} & \ding{51} & \ding{51} & \ding{51} & \ding{51} & \ding{51} & \ding{51} & \ding{51} & \ding{51} \\
CricketX & u & 61,54 & 300 & $300 \times 1$ & 1 & 3 &  &  & \ding{51} & \ding{51} & \ding{51} &  & \ding{51} &  &  & \ding{51} &  &  &  &  &  &  \\
DiatomSizeReduction & u & 100,00 & 345 & $345 \times 1$ & 2 & 12 & \ding{51} & \ding{51} & \ding{51} & \ding{51} & \ding{51} & \ding{51} &  & \ding{51} & \ding{51} &  & \ding{51} & \ding{51} &  &  & \ding{51} &  \\
DistalPhalanxOutlineAgeGroup & u & 77,78 & 80 & $80 \times 1$ & 1 & 3 &  &  & \ding{51} & \ding{51} & \ding{51} &  & \ding{51} &  &  & \ding{51} &  &  &  &  &  &  \\
DistalPhalanxOutlineCorrect & u & 78,08 & 80 & $80 \times 1$ & 1 & 3 &  &  & \ding{51} & \ding{51} & \ding{51} &  & \ding{51} &  &  & \ding{51} &  &  &  &  &  &  \\
DodgerLoopDay & u & 55,00 & 288 & $288 \times 1$ & 3 & 33 & \ding{51} & \ding{51} & \ding{51} & \ding{51} & \ding{51} & \ding{51} & \ding{51} & \ding{51} & \ding{51} & \ding{51} & \ding{51} & \ding{51} & \ding{51} & \ding{51} & \ding{51} & \ding{51} \\
DodgerLoopGame & u & 87,50 & 288 & $288 \times 1$ & 3 & 33 & \ding{51} & \ding{51} & \ding{51} & \ding{51} & \ding{51} & \ding{51} & \ding{51} & \ding{51} & \ding{51} & \ding{51} & \ding{51} & \ding{51} & \ding{51} & \ding{51} & \ding{51} & \ding{51} \\
ECG200 & u & 86,00 & 96 & $96 \times 1$ & 3 & 33 & \ding{51} & \ding{51} & \ding{51} & \ding{51} & \ding{51} & \ding{51} & \ding{51} & \ding{51} & \ding{51} & \ding{51} & \ding{51} & \ding{51} & \ding{51} & \ding{51} & \ding{51} & \ding{51} \\
ECG5000 & u & 91,52 & 140 & $140 \times 1$ & 1 & 3 &  &  & \ding{51} & \ding{51} & \ding{51} &  & \ding{51} &  &  & \ding{51} &  &  &  &  &  &  \\
ECGFiveDays & u & 100,00 & 136 & $136 \times 1$ & 3 & 33 & \ding{51} & \ding{51} & \ding{51} & \ding{51} & \ding{51} & \ding{51} & \ding{51} & \ding{51} & \ding{51} & \ding{51} & \ding{51} & \ding{51} & \ding{51} & \ding{51} & \ding{51} & \ding{51} \\
FaceFour & u & 96,43 & 350 & $350 \times 1$ & 1 & 3 &  &  & \ding{51} & \ding{51} & \ding{51} &  & \ding{51} &  &  & \ding{51} &  &  &  &  &  &  \\
GunPoint & u & 86,00 & 150 & $150 \times 1$ & 2 & 12 & \ding{51} & \ding{51} & \ding{51} & \ding{51} & \ding{51} & \ding{51} &  & \ding{51} & \ding{51} &  & \ding{51} & \ding{51} &  &  & \ding{51} &  \\
GunPointAgeSpan & u & 88,50 & 150 & $150 \times 1$ & 3 & 33 & \ding{51} & \ding{51} & \ding{51} & \ding{51} & \ding{51} & \ding{51} & \ding{51} & \ding{51} & \ding{51} & \ding{51} & \ding{51} & \ding{51} & \ding{51} & \ding{51} & \ding{51} & \ding{51} \\
GunPointMaleVersusFemale & u & 100,00 & 150 & $150 \times 1$ & 2 & 12 & \ding{51} & \ding{51} & \ding{51} & \ding{51} & \ding{51} & \ding{51} &  & \ding{51} & \ding{51} &  & \ding{51} & \ding{51} &  &  & \ding{51} &  \\
GunPointOldVersusYoung & u & 100,00 & 150 & $150 \times 1$ & 2 & 12 & \ding{51} & \ding{51} & \ding{51} & \ding{51} & \ding{51} & \ding{51} &  & \ding{51} & \ding{51} &  & \ding{51} & \ding{51} &  &  & \ding{51} &  \\
InsectWingbeatSound & u & 66,91 & 256 & $256 \times 1$ & 1 & 3 &  &  & \ding{51} & \ding{51} & \ding{51} &  &  &  & \ding{51} &  &  & \ding{51} &  &  &  &  \\
MedicalImages & u & 68,53 & 99 & $99 \times 1$ & 1 & 3 &  &  & \ding{51} & \ding{51} & \ding{51} &  & \ding{51} &  &  & \ding{51} &  &  &  &  &  &  \\
MiddlePhalanxOutlineAgeGroup & u & 78,42 & 80 & $80 \times 1$ & 1 & 3 &  &  & \ding{51} & \ding{51} & \ding{51} &  & \ding{51} &  &  & \ding{51} &  &  &  &  &  &  \\
PhalangesOutlinesCorrect & u & 67,07 & 80 & $80 \times 1$ & 1 & 3 &  &  & \ding{51} & \ding{51} & \ding{51} &  & \ding{51} &  &  & \ding{51} &  &  &  &  &  &  \\
Plane & u & 96,23 & 144 & $144 \times 1$ & 3 & 33 & \ding{51} & \ding{51} & \ding{51} & \ding{51} & \ding{51} & \ding{51} & \ding{51} & \ding{51} & \ding{51} & \ding{51} & \ding{51} & \ding{51} & \ding{51} & \ding{51} & \ding{51} & \ding{51} \\
PowerCons & u & 100,00 & 144 & $144 \times 1$ & 1 & 3 &  &  & \ding{51} & \ding{51} & \ding{51} &  & \ding{51} &  &  & \ding{51} &  &  &  &  &  &  \\
ProximalPhalanxTW & u & 48,68 & 80 & $80 \times 1$ & 3 & 33 & \ding{51} & \ding{51} & \ding{51} & \ding{51} & \ding{51} & \ding{51} & \ding{51} & \ding{51} & \ding{51} & \ding{51} & \ding{51} & \ding{51} & \ding{51} & \ding{51} & \ding{51} & \ding{51} \\
SmoothSubspace & u & 96,00 & 15 & $15 \times 1$ & 1 & 3 &  &  & \ding{51} & \ding{51} & \ding{51} &  & \ding{51} &  &  & \ding{51} &  &  &  &  &  &  \\
Strawberry & u & 77,64 & 235 & $235 \times 1$ & 1 & 3 &  &  & \ding{51} & \ding{51} & \ding{51} &  & \ding{51} &  &  & \ding{51} &  &  &  &  &  &  \\
SyntheticControl & u & 89,33 & 60 & $60 \times 1$ & 1 & 3 &  &  & \ding{51} & \ding{51} & \ding{51} &  & \ding{51} &  &  & \ding{51} &  &  &  &  &  &  \\
Trace & u & 68,00 & 275 & $275 \times 1$ & 1 & 3 &  &  & \ding{51} & \ding{51} & \ding{51} &  & \ding{51} &  &  & \ding{51} &  &  &  &  &  &  \\
TwoPatterns & u & 99,92 & 128 & $128 \times 1$ & 2 & 12 & \ding{51} & \ding{51} & \ding{51} & \ding{51} & \ding{51} & \ding{51} &  & \ding{51} & \ding{51} &  & \ding{51} & \ding{51} &  &  & \ding{51} &  \\
UMD & u & 93,33 & 150 & $150 \times 1$ & 3 & 33 & \ding{51} & \ding{51} & \ding{51} & \ding{51} & \ding{51} & \ding{51} & \ding{51} & \ding{51} & \ding{51} & \ding{51} & \ding{51} & \ding{51} & \ding{51} & \ding{51} & \ding{51} & \ding{51} \\
Wafer & u & 99,83 & 152 & $152 \times 1$ & 3 & 33 & \ding{51} & \ding{51} & \ding{51} & \ding{51} & \ding{51} & \ding{51} & \ding{51} & \ding{51} & \ding{51} & \ding{51} & \ding{51} & \ding{51} & \ding{51} & \ding{51} & \ding{51} & \ding{51} \\
\end{tabular}
\caption{Dataset summary with baselines, methods, and fusion methods. \textbf{Type} is represented as \textbf{u} (univariate) or \textbf{m} (multivariate). Column \textbf{Acc} denotes Accuracy of the ML model. Column \textbf{\#F} denotes the total number of input features, and \textbf{Dim} denotes the input time series dimensionality (time $\times$ variables). Columns \textbf{\#B} and \textbf{\#E} indicate the number of baselines and ensembles for each dataset.}
\label{tab:dataset_summary}
\end{minipage}
}
\end{sidewaystable}

\clearpage \subsubsection{Wilcoxon Signed-rank Tests}
\begin{table}[ht]
    \centering
    \caption{Wilcoxon signed-rank test results comparing baselines (single explainers: Anchor, LIME, SHAP) for each metric.
    Columns include medians for each pair, test statistic (\(W\)), p-value (\(p\)), and the number of compared pairs (\(N\)).
    Statistically significant results are marked with asterisks: \(^{*}\) (\(p < 0.05\)), \(^{**}\) (\(p < 0.01\)), \(^{***}\) (\(p < 0.001\)).}
    \label{tab:wilcoxon_single_explainers}
    \begin{tabular}{lcccccc}
        \toprule
        Metric & Explainers & \(N\) & Median1 & Median2 & \(W\) & \(p\) \\
        \midrule
\multirow{3}{*}{$\bar{M}$} & Anchor vs LIME & 10 & 16.89 & 4.95 & 2.0 &0.006$^{**}$ \\
 & Anchor vs SHAP & 10 & 16.89 & 6.86 & 4.0 &0.014$^{*}$ \\
 & LIME vs SHAP & 27 & 3.30 & 2.56 & 177.0 & 0.79 \\
\midrule
\multirow{3}{*}{$\bar{{CONF}} \cdot {ER}$} & Anchor vs LIME & 10 & 83.43 & 65.75 & 20.0 & 0.49 \\
 & Anchor vs SHAP & 10 & 83.43 & 86.99 & 24.0 & 0.77 \\
 & LIME vs SHAP & 27 & 63.70 & 89.17 & 55.0 &< 0.001$^{***}$ \\
\midrule
\multirow{3}{*}{$\bar{{CONF}} \cdot \bar{{COV}} \cdot {ER}$} & Anchor vs LIME & 10 & 17.48 & 6.03 & 5.0 &0.019$^{*}$ \\
 & Anchor vs SHAP & 10 & 17.48 & 7.82 & 6.0 &0.027$^{*}$ \\
 & LIME vs SHAP & 27 & 7.83 & 6.24 & 154.0 & 0.41 \\
\midrule
\multirow{3}{*}{$\bar{{CONF}}$} & Anchor vs LIME & 10 & 0.88 & 0.88 & 16.0 & 0.28 \\
 & Anchor vs SHAP & 10 & 0.88 & 0.94 & 24.0 & 0.77 \\
 & LIME vs SHAP & 27 & 0.82 & 0.92 & 60.0 &0.001$^{**}$ \\
\midrule
\multirow{3}{*}{$\bar{{COV}}$} & Anchor vs LIME & 10 & 0.20 & 0.12 & 12.0 & 0.13 \\
 & Anchor vs SHAP & 10 & 0.20 & 0.09 & 11.0 & 0.11 \\
 & LIME vs SHAP & 27 & 0.13 & 0.10 & 113.0 & 0.07 \\
\midrule
\multirow{3}{*}{${{ER}}$} & Anchor vs LIME & 10 & 98.03 & 88.10 & 17.0 & 0.89 \\
 & Anchor vs SHAP & 10 & 98.03 & 100.00 & 10.0 & 0.92 \\
 & LIME vs SHAP & 27 & 86.11 & 100.00 & 62.0 & 0.06 \\
\midrule
\multirow{3}{*}{$\bar{F(n)}$} & Anchor vs LIME & 10 & 1.94 & 13.46 & 1.0 &0.004$^{**}$ \\
 & Anchor vs SHAP & 10 & 1.94 & 7.12 & 1.0 &0.004$^{**}$ \\
 & LIME vs SHAP & 27 & 19.98 & 32.04 & 137.0 & 0.22 \\
\midrule
\multirow{3}{*}{${Med}(F(n))$} & Anchor vs LIME & 10 & 2.00 & 8.00 & 4.0 &0.014$^{*}$ \\
 & Anchor vs SHAP & 10 & 2.00 & 6.50 & 1.5 &0.013$^{*}$ \\
 & LIME vs SHAP & 27 & 9.50 & 21.00 & 89.0 &0.048$^{*}$ \\

        \bottomrule
    \end{tabular}
\end{table}

\begin{table}[ht]
\centering
\caption{Wilcoxon signed-rank test results comparing explainers (and their fused rule sets) to Anchor (Baseline) for $\bar{M}$.}
\label{tab:wilcoxon_rules_combination_final_metric}
\begin{tabular}{lrrrrr}
\toprule
\textbf{Explainers set} & \(\mathbf{N}\) & \(\mathbf{Median_1}\) & \(\mathbf{Median_2}\) & \(\mathbf{W}\) & \(\mathbf{p}\) \\
\midrule
Anchor+LIME vs Anchor & 10 & 9.94 & 17.29 & 0.00 &0.002$^{**}$ \\
Anchor+LIME+SHAP vs Anchor & 10 & 7.60 & 17.29 & 3.00 &0.010$^{**}$ \\
Anchor+SHAP vs Anchor & 10 & 11.03 & 17.29 & 5.00 &0.019$^{*}$ \\
LIME vs Anchor & 10 & 4.65 & 17.29 & 0.00 &0.002$^{**}$ \\
LIME+SHAP vs Anchor & 10 & 5.79 & 17.29 & 0.00 &0.002$^{**}$ \\
SHAP vs Anchor & 10 & 8.32 & 17.29 & 0.00 &0.002$^{**}$ \\
\multicolumn{6}{l}{\scriptsize *p(<0.05), **p(<0.01), ***p(<0.001)} \\
\bottomrule
\end{tabular}
\end{table}
\begin{table}[ht]
\centering
\caption{Wilcoxon signed-rank test results comparing explainers (and their fused rule sets) to Anchor (Baseline) for $\bar{{CONF}} \cdot {ER}$.}
\label{tab:wilcoxon_rules_combination_metric_con_xratio}
\begin{tabular}{lrrrrr}
\toprule
\textbf{Explainers set} & \(\mathbf{N}\) & \(\mathbf{Median_1}\) & \(\mathbf{Median_2}\) & \(\mathbf{W}\) & \(\mathbf{p}\) \\
\midrule
Anchor+LIME vs Anchor & 10 & 74.08 & 68.71 & 20.00 & 0.49 \\
Anchor+LIME+SHAP vs Anchor & 10 & 69.32 & 68.71 & 27.00 & 1.00 \\
Anchor+SHAP vs Anchor & 10 & 77.28 & 68.71 & 15.00 & 0.23 \\
LIME vs Anchor & 10 & 59.68 & 68.71 & 15.00 & 0.23 \\
LIME+SHAP vs Anchor & 10 & 73.12 & 68.71 & 26.00 & 0.92 \\
SHAP vs Anchor & 10 & 70.64 & 68.71 & 21.00 & 0.56 \\
\multicolumn{6}{l}{\scriptsize *p(<0.05), **p(<0.01), ***p(<0.001)} \\
\bottomrule
\end{tabular}
\end{table}
\begin{table}[ht]
\centering
\caption{Wilcoxon signed-rank test results comparing explainers (and their fused rule sets) to Anchor (Baseline) for $\bar{{CONF}} \cdot \bar{{COV}} \cdot {ER}$.}
\label{tab:wilcoxon_rules_combination_metric_con_cov_xratio}
\begin{tabular}{lrrrrr}
\toprule
\textbf{Explainers set} & \(\mathbf{N}\) & \(\mathbf{Median_1}\) & \(\mathbf{Median_2}\) & \(\mathbf{W}\) & \(\mathbf{p}\) \\
\midrule
Anchor+LIME vs Anchor & 10 & 11.36 & 18.64 & 4.00 &0.014$^{*}$ \\
Anchor+LIME+SHAP vs Anchor & 10 & 9.02 & 18.64 & 4.00 &0.014$^{*}$ \\
Anchor+SHAP vs Anchor & 10 & 11.86 & 18.64 & 5.00 &0.019$^{*}$ \\
LIME vs Anchor & 10 & 5.87 & 18.64 & 2.00 &0.006$^{**}$ \\
LIME+SHAP vs Anchor & 10 & 7.14 & 18.64 & 1.00 &0.004$^{**}$ \\
SHAP vs Anchor & 10 & 8.85 & 18.64 & 0.00 &0.002$^{**}$ \\
\multicolumn{6}{l}{\scriptsize *p(<0.05), **p(<0.01), ***p(<0.001)} \\
\bottomrule
\end{tabular}
\end{table}
\begin{table}[ht]
\centering
\caption{Wilcoxon signed-rank test results comparing explainers (and their fused rule sets) to Anchor (Baseline) for $\bar{{CONF}}$.}
\label{tab:wilcoxon_rules_combination_avg_confidence}
\begin{tabular}{lrrrrr}
\toprule
\textbf{Explainers set} & \(\mathbf{N}\) & \(\mathbf{Median_1}\) & \(\mathbf{Median_2}\) & \(\mathbf{W}\) & \(\mathbf{p}\) \\
\midrule
Anchor+LIME vs Anchor & 10 & 0.85 & 0.84 & 17.00 & 0.32 \\
Anchor+LIME+SHAP vs Anchor & 10 & 0.87 & 0.84 & 9.00 & 0.06 \\
Anchor+SHAP vs Anchor & 10 & 0.88 & 0.84 & 4.00 &0.014$^{*}$ \\
LIME vs Anchor & 10 & 0.81 & 0.84 & 13.00 & 0.16 \\
LIME+SHAP vs Anchor & 10 & 0.89 & 0.84 & 20.00 & 0.49 \\
SHAP vs Anchor & 10 & 0.86 & 0.84 & 26.00 & 0.92 \\
\multicolumn{6}{l}{\scriptsize *p(<0.05), **p(<0.01), ***p(<0.001)} \\
\bottomrule
\end{tabular}
\end{table}
\begin{table}[ht]
\centering
\caption{Wilcoxon signed-rank test results comparing explainers (and their fused rule sets) to Anchor (Baseline) for $\bar{{COV}}$.}
\label{tab:wilcoxon_rules_combination_avg_coverage}
\begin{tabular}{lrrrrr}
\toprule
\textbf{Explainers set} & \(\mathbf{N}\) & \(\mathbf{Median_1}\) & \(\mathbf{Median_2}\) & \(\mathbf{W}\) & \(\mathbf{p}\) \\
\midrule
Anchor+LIME vs Anchor & 10 & 0.19 & 0.26 & 1.00 &0.004$^{**}$ \\
Anchor+LIME+SHAP vs Anchor & 10 & 0.15 & 0.26 & 1.00 &0.004$^{**}$ \\
Anchor+SHAP vs Anchor & 10 & 0.19 & 0.26 & 2.00 &0.006$^{**}$ \\
LIME vs Anchor & 10 & 0.15 & 0.26 & 9.00 & 0.06 \\
LIME+SHAP vs Anchor & 10 & 0.12 & 0.26 & 6.00 &0.027$^{*}$ \\
SHAP vs Anchor & 10 & 0.16 & 0.26 & 8.00 &0.049$^{*}$ \\
\multicolumn{6}{l}{\scriptsize *p(<0.05), **p(<0.01), ***p(<0.001)} \\
\bottomrule
\end{tabular}
\end{table}
\begin{table}[ht]
\centering
\caption{Wilcoxon signed-rank test results comparing explainers (and their fused rule sets) to Anchor (Baseline) for ${{ER}}$.}
\label{tab:wilcoxon_rules_combination_explained_ratio}
\begin{tabular}{lrrrrr}
\toprule
\textbf{Explainers set} & \(\mathbf{N}\) & \(\mathbf{Median_1}\) & \(\mathbf{Median_2}\) & \(\mathbf{W}\) & \(\mathbf{p}\) \\
\midrule
Anchor+LIME vs Anchor & 10 & 81.37 & 89.87 & 27.00 & 1.00 \\
Anchor+LIME+SHAP vs Anchor & 10 & 79.71 & 89.87 & 19.00 & 0.43 \\
Anchor+SHAP vs Anchor & 10 & 87.25 & 89.87 & 22.00 & 0.62 \\
LIME vs Anchor & 10 & 70.55 & 89.87 & 16.00 & 0.28 \\
LIME+SHAP vs Anchor & 10 & 81.04 & 89.87 & 24.00 & 0.77 \\
SHAP vs Anchor & 10 & 76.81 & 89.87 & 15.00 & 0.23 \\
\multicolumn{6}{l}{\scriptsize *p(<0.05), **p(<0.01), ***p(<0.001)} \\
\bottomrule
\end{tabular}
\end{table}
\begin{table}[ht]
\centering
\caption{Wilcoxon signed-rank test results comparing explainers (and their fused rule sets) to Anchor (Baseline) for $\bar{F(n)}$.}
\label{tab:wilcoxon_rules_combination_avg_features_count}
\begin{tabular}{lrrrrr}
\toprule
\textbf{Explainers set} & \(\mathbf{N}\) & \(\mathbf{Median_1}\) & \(\mathbf{Median_2}\) & \(\mathbf{W}\) & \(\mathbf{p}\) \\
\midrule
Anchor+LIME vs Anchor & 10 & 7.34 & 1.55 & 0.00 &0.002$^{**}$ \\
Anchor+LIME+SHAP vs Anchor & 10 & 9.89 & 1.55 & 1.00 &0.004$^{**}$ \\
Anchor+SHAP vs Anchor & 10 & 4.56 & 1.55 & 1.00 &0.004$^{**}$ \\
LIME vs Anchor & 10 & 9.12 & 1.55 & 1.00 &0.004$^{**}$ \\
LIME+SHAP vs Anchor & 10 & 12.24 & 1.55 & 1.00 &0.004$^{**}$ \\
SHAP vs Anchor & 10 & 4.18 & 1.55 & 2.00 &0.006$^{**}$ \\
\multicolumn{6}{l}{\scriptsize *p(<0.05), **p(<0.01), ***p(<0.001)} \\
\bottomrule
\end{tabular}
\end{table}
\begin{table}[ht]
\centering
\caption{Wilcoxon signed-rank test results comparing explainers (and their fused rule sets) to Anchor (Baseline) for ${Med}(F(n))$.}
\label{tab:wilcoxon_rules_combination_median_features_count}
\begin{tabular}{lrrrrr}
\toprule
\textbf{Explainers set} & \(\mathbf{N}\) & \(\mathbf{Median_1}\) & \(\mathbf{Median_2}\) & \(\mathbf{W}\) & \(\mathbf{p}\) \\
\midrule
Anchor+LIME vs Anchor & 10 & 3.29 & 1.75 & 0.00 &0.002$^{**}$ \\
Anchor+LIME+SHAP vs Anchor & 10 & 5.96 & 1.75 & 1.00 &0.004$^{**}$ \\
Anchor+SHAP vs Anchor & 10 & 2.92 & 1.75 & 1.00 &0.004$^{**}$ \\
LIME vs Anchor & 10 & 4.50 & 1.75 & 4.00 &0.014$^{*}$ \\
LIME+SHAP vs Anchor & 10 & 8.50 & 1.75 & 1.00 &0.004$^{**}$ \\
SHAP vs Anchor & 10 & 3.88 & 1.75 & 5.50 &0.027$^{*}$ \\
\multicolumn{6}{l}{\scriptsize *p(<0.05), **p(<0.01), ***p(<0.001)} \\
\bottomrule
\end{tabular}
\end{table}

\begin{table}[ht]
\centering
\caption{Wilcoxon signed-rank test results comparing fusion methods (consisting of one or more explainers) to the Baseline explainer (Anchor, LIME or SHAP) for $\bar{M}$.}
\label{tab:wilcoxon_method_final_metric}
\begin{tabular}{lrrrrr}
\toprule
\textbf{Methods} & \(\mathbf{N}\) & \(\mathbf{Median_1}\) & \(\mathbf{Median_2}\) & \(\mathbf{W}\) & \(\mathbf{p}\) \\
\midrule
Best vs Baseline & 27 & 5.57 & 5.46 & 163.00 & 0.55 \\
Intersection vs Baseline & 27 & 3.92 & 5.46 & 166.00 & 0.59 \\
Lasso vs Baseline & 40 & 7.56 & 5.66 & 146.00 &< 0.001$^{***}$ \\
Lasso global vs Baseline & 43 & 6.92 & 5.40 & 377.00 & 0.25 \\
Union vs Baseline & 27 & 1.76 & 5.46 & 48.00 &< 0.001$^{***}$ \\
Weighted vs Baseline & 27 & 6.29 & 5.46 & 82.00 &0.009$^{**}$ \\
\multicolumn{6}{l}{\scriptsize *p(<0.05), **p(<0.01), ***p(<0.001)} \\
\bottomrule
\end{tabular}
\end{table}
\begin{table}[ht]
\centering
\caption{Wilcoxon signed-rank test results comparing fusion methods (consisting of one or more explainers) to the Baseline explainer (Anchor, LIME or SHAP) for $\bar{{CONF}} \cdot {ER}$.}
\label{tab:wilcoxon_method_metric_con_xratio}
\begin{tabular}{lrrrrr}
\toprule
\textbf{Methods} & \(\mathbf{N}\) & \(\mathbf{Median_1}\) & \(\mathbf{Median_2}\) & \(\mathbf{W}\) & \(\mathbf{p}\) \\
\midrule
Best vs Baseline & 27 & 95.59 & 70.97 & 0.00 &< 0.001$^{***}$ \\
Intersection vs Baseline & 27 & 50.46 & 70.97 & 62.00 &0.002$^{**}$ \\
Lasso vs Baseline & 40 & 78.49 & 73.31 & 114.00 &< 0.001$^{***}$ \\
Lasso global vs Baseline & 43 & 28.73 & 70.97 & 141.00 &< 0.001$^{***}$ \\
Union vs Baseline & 27 & 94.21 & 70.97 & 9.00 &< 0.001$^{***}$ \\
Weighted vs Baseline & 27 & 76.89 & 70.97 & 79.00 &0.007$^{**}$ \\
\multicolumn{6}{l}{\scriptsize *p(<0.05), **p(<0.01), ***p(<0.001)} \\
\bottomrule
\end{tabular}
\end{table}
\begin{table}[ht]
\centering
\caption{Wilcoxon signed-rank test results comparing fusion methods (consisting of one or more explainers) to the Baseline explainer (Anchor, LIME or SHAP) for $\bar{{CONF}} \cdot \bar{{COV}} \cdot {ER}$.}
\label{tab:wilcoxon_method_metric_con_cov_xratio}
\begin{tabular}{lrrrrr}
\toprule
\textbf{Methods} & \(\mathbf{N}\) & \(\mathbf{Median_1}\) & \(\mathbf{Median_2}\) & \(\mathbf{W}\) & \(\mathbf{p}\) \\
\midrule
Best vs Baseline & 27 & 10.00 & 11.26 & 152.00 & 0.39 \\
Intersection vs Baseline & 27 & 6.93 & 11.26 & 105.00 &0.044$^{*}$ \\
Lasso vs Baseline & 40 & 10.38 & 11.02 & 245.00 &0.026$^{*}$ \\
Lasso global vs Baseline & 43 & 9.85 & 10.02 & 459.00 & 0.87 \\
Union vs Baseline & 27 & 5.64 & 11.26 & 82.00 &0.009$^{**}$ \\
Weighted vs Baseline & 28 & 12.15 & 11.31 & 95.00 &0.013$^{*}$ \\
\multicolumn{6}{l}{\scriptsize *p(<0.05), **p(<0.01), ***p(<0.001)} \\
\bottomrule
\end{tabular}
\end{table}
\begin{table}[ht]
\centering
\caption{Wilcoxon signed-rank test results comparing fusion methods (consisting of one or more explainers) to the Baseline explainer (Anchor, LIME or SHAP) for $\bar{{CONF}}$.}
\label{tab:wilcoxon_method_avg_confidence}
\begin{tabular}{lrrrrr}
\toprule
\textbf{Methods} & \(\mathbf{N}\) & \(\mathbf{Median_1}\) & \(\mathbf{Median_2}\) & \(\mathbf{W}\) & \(\mathbf{p}\) \\
\midrule
Best vs Baseline & 27 & 0.97 & 0.85 & 0.00 &< 0.001$^{***}$ \\
Intersection vs Baseline & 27 & 0.83 & 0.85 & 85.00 &0.011$^{*}$ \\
Lasso vs Baseline & 40 & 0.88 & 0.84 & 178.00 &0.001$^{**}$ \\
Lasso global vs Baseline & 43 & 0.68 & 0.85 & 72.00 &< 0.001$^{***}$ \\
Union vs Baseline & 27 & 0.94 & 0.85 & 47.00 &< 0.001$^{***}$ \\
Weighted vs Baseline & 27 & 0.88 & 0.85 & 41.00 &< 0.001$^{***}$ \\
\multicolumn{6}{l}{\scriptsize *p(<0.05), **p(<0.01), ***p(<0.001)} \\
\bottomrule
\end{tabular}
\end{table}
\begin{table}[ht]
\centering
\caption{Wilcoxon signed-rank test results comparing fusion methods (consisting of one or more explainers) to the Baseline explainer (Anchor, LIME or SHAP) for $\bar{{COV}}$.}
\label{tab:wilcoxon_method_avg_coverage}
\begin{tabular}{lrrrrr}
\toprule
\textbf{Methods} & \(\mathbf{N}\) & \(\mathbf{Median_1}\) & \(\mathbf{Median_2}\) & \(\mathbf{W}\) & \(\mathbf{p}\) \\
\midrule
Best vs Baseline & 27 & 0.11 & 0.17 & 38.00 &< 0.001$^{***}$ \\
Intersection vs Baseline & 27 & 0.16 & 0.17 & 178.00 & 0.80 \\
Lasso vs Baseline & 40 & 0.17 & 0.18 & 395.00 & 0.85 \\
Lasso global vs Baseline & 42 & 0.34 & 0.19 & 2.00 &< 0.001$^{***}$ \\
Union vs Baseline & 27 & 0.06 & 0.17 & 16.00 &< 0.001$^{***}$ \\
Weighted vs Baseline & 27 & 0.17 & 0.17 & 172.00 & 0.70 \\
\multicolumn{6}{l}{\scriptsize *p(<0.05), **p(<0.01), ***p(<0.001)} \\
\bottomrule
\end{tabular}
\end{table}
\begin{table}[ht]
\centering
\caption{Wilcoxon signed-rank test results comparing fusion methods (consisting of one or more explainers) to the Baseline explainer (Anchor, LIME or SHAP) for ${{ER}}$.}
\label{tab:wilcoxon_method_explained_ratio}
\begin{tabular}{lrrrrr}
\toprule
\textbf{Methods} & \(\mathbf{N}\) & \(\mathbf{Median_1}\) & \(\mathbf{Median_2}\) & \(\mathbf{W}\) & \(\mathbf{p}\) \\
\midrule
Best vs Baseline & 21 & 100.00 & 86.67 & 0.00 &< 0.001$^{***}$ \\
Intersection vs Baseline & 27 & 70.00 & 90.00 & 65.00 &0.002$^{**}$ \\
Lasso vs Baseline & 25 & 93.02 & 90.00 & 26.00 &< 0.001$^{***}$ \\
Lasso global vs Baseline & 43 & 53.43 & 91.67 & 241.00 &0.004$^{**}$ \\
Union vs Baseline & 21 & 100.00 & 86.67 & 0.00 &< 0.001$^{***}$ \\
Weighted vs Baseline & 26 & 89.60 & 90.39 & 148.00 & 0.50 \\
\multicolumn{6}{l}{\scriptsize *p(<0.05), **p(<0.01), ***p(<0.001)} \\
\bottomrule
\end{tabular}
\end{table}
\begin{table}[ht]
\centering
\caption{Wilcoxon signed-rank test results comparing fusion methods (consisting of one or more explainers) to the Baseline explainer (Anchor, LIME or SHAP) for $\bar{F(n)}$.}
\label{tab:wilcoxon_method_avg_features_count}
\begin{tabular}{lrrrrr}
\toprule
\textbf{Methods} & \(\mathbf{N}\) & \(\mathbf{Median_1}\) & \(\mathbf{Median_2}\) & \(\mathbf{W}\) & \(\mathbf{p}\) \\
\midrule
Best vs Baseline & 27 & 27.50 & 23.59 & 90.00 &0.016$^{*}$ \\
Intersection vs Baseline & 27 & 6.07 & 23.59 & 40.00 &< 0.001$^{***}$ \\
Lasso vs Baseline & 40 & 5.05 & 10.65 & 19.00 &< 0.001$^{***}$ \\
Lasso global vs Baseline & 43 & 1.02 & 9.33 & 46.00 &< 0.001$^{***}$ \\
Union vs Baseline & 27 & 42.61 & 23.59 & 0.00 &< 0.001$^{***}$ \\
Weighted vs Baseline & 27 & 23.90 & 23.59 & 128.00 & 0.15 \\
\multicolumn{6}{l}{\scriptsize *p(<0.05), **p(<0.01), ***p(<0.001)} \\
\bottomrule
\end{tabular}
\end{table}
\begin{table}[ht]
\centering
\caption{Wilcoxon signed-rank test results comparing fusion methods (consisting of one or more explainers) to the Baseline explainer (Anchor, LIME or SHAP) for ${Med}(F(n))$.}
\label{tab:wilcoxon_method_median_features_count}
\begin{tabular}{lrrrrr}
\toprule
\textbf{Methods} & \(\mathbf{N}\) & \(\mathbf{Median_1}\) & \(\mathbf{Median_2}\) & \(\mathbf{W}\) & \(\mathbf{p}\) \\
\midrule
Best vs Baseline & 26 & 26.50 & 21.58 & 61.50 &0.003$^{**}$ \\
Intersection vs Baseline & 27 & 2.25 & 20.50 & 34.00 &< 0.001$^{***}$ \\
Lasso vs Baseline & 28 & 4.83 & 21.58 & 44.00 &< 0.001$^{***}$ \\
Lasso global vs Baseline & 42 & 0.67 & 7.17 & 73.00 &< 0.001$^{***}$ \\
Union vs Baseline & 27 & 42.00 & 20.50 & 0.00 &< 0.001$^{***}$ \\
Weighted vs Baseline & 26 & 17.33 & 21.58 & 132.50 & 0.28 \\
\multicolumn{6}{l}{\scriptsize *p(<0.05), **p(<0.01), ***p(<0.001)} \\
\bottomrule
\end{tabular}
\end{table}

\clearpage \subsubsection{Median Values of Metrics}
\begin{table}[ht]
\centering
\caption{Summary of medians for fusion combinations across all metrics.}
\label{tab:median_summary_rules_combination_all_metrics}
\tiny
\begin{tabular}{lccccccccc}
\toprule
\textbf{Fusion} & N & $\bar{M}$ & $\bar{{CONF}} \cdot {ER}$ & $\bar{{CONF}} \cdot \bar{{COV}} \cdot {ER}$ & $\bar{{CONF}}$ & $\bar{{COV}}$ & ${{ER}}$ & $\bar{F(n)}$ & ${Med}(F(n))$ \\
\midrule
Anchor & 100 & 17.24 & 79.81 & 17.86 & 0.87 & 0.22 & 97.03 & 1.79 & 2.00 \\
Anchor+LIME & 60 & 10.62 & 84.73 & 12.31 & 0.90 & 0.17 & 95.92 & 6.66 & 2.75 \\
Anchor+SHAP & 60 & 10.44 & 90.47 & 12.01 & 0.92 & 0.15 & 97.69 & 4.91 & 2.25 \\
Anchor+LIME+SHAP & 60 & 6.92 & 84.52 & 8.55 & 0.91 & 0.12 & 93.59 & 8.49 & 4.75 \\
SHAP & 108 & 6.86 & 85.89 & 7.82 & 0.94 & 0.13 & 96.46 & 4.69 & 4.50 \\
LIME+SHAP & 162 & 5.51 & 86.03 & 7.47 & 0.94 & 0.10 & 92.91 & 12.54 & 8.50 \\
LIME & 112 & 5.05 & 64.79 & 6.05 & 0.87 & 0.13 & 85.90 & 10.90 & 5.38 \\
\bottomrule
\end{tabular}
\end{table}
\begin{table}[ht]
\centering
\caption{Summary of medians for fusion methods across all metrics.}
\label{tab:median_summary_methods_all_metrics}
\tiny
\begin{tabular}{lccccccccc}
\toprule
\textbf{Method} & N & $\bar{M}$ & $\bar{{CONF}} \cdot {ER}$ & $\bar{{CONF}} \cdot \bar{{COV}} \cdot {ER}$ & $\bar{{CONF}}$ & $\bar{{COV}}$ & ${{ER}}$ & $\bar{F(n)}$ & ${Med}(F(n))$ \\
\midrule
Lasso global & 137 & 6.53 & 22.05 & 7.58 & 0.66 & 0.33 & 46.90 & 0.80 & 0.00 \\
Weighted & 137 & 6.15 & 82.19 & 11.33 & 0.90 & 0.15 & 93.83 & 22.74 & 13.00 \\
Best & 57 & 5.55 & 96.86 & 10.00 & 0.97 & 0.10 & 100.00 & 28.42 & 26.00 \\
Lasso & 137 & 5.28 & 91.05 & 9.35 & 0.94 & 0.12 & 100.00 & 6.69 & 6.00 \\
Baseline & 80 & 4.73 & 76.33 & 8.94 & 0.86 & 0.14 & 93.75 & 22.54 & 20.50 \\
Intersection & 57 & 3.92 & 51.32 & 6.29 & 0.81 & 0.13 & 70.00 & 5.76 & 1.25 \\
Union & 57 & 1.67 & 94.49 & 5.27 & 0.94 & 0.06 & 100.00 & 42.61 & 42.00 \\
\bottomrule
\end{tabular}
\end{table}

\FloatBarrier
\subsection{Additional Figures}\label{sec:figures}
\FloatBarrier \subsubsection{Plots for Baseline (single) Explainers}

\begin{figure}[H]
  \centering

  \subfloat[\(\bar{M}\) (Objective function)\label{fig:single_explainers_plots__obj_fun}]{
    \includegraphics[width=0.45\textwidth]{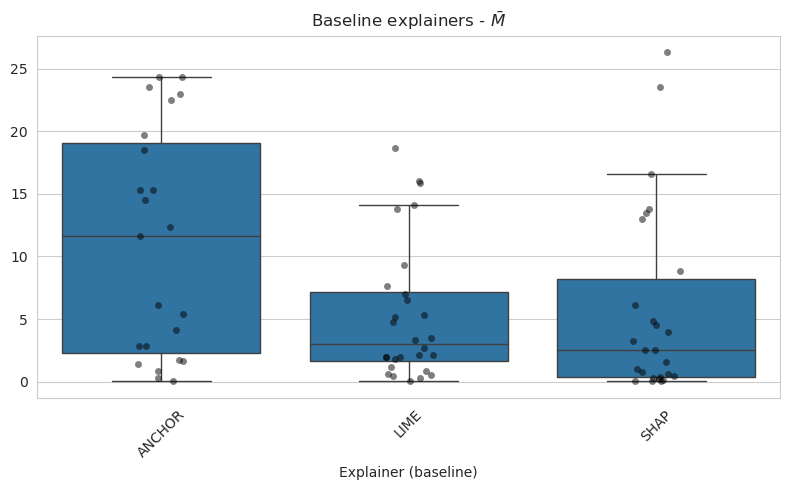}
  }\hfill
  \subfloat[\(\overline{\text{CONF}} \cdot ER\) (Avg. Conf.*Expl. R.)\label{fig:single_explainers_plots__conf_er}]{
    \includegraphics[width=0.45\textwidth]{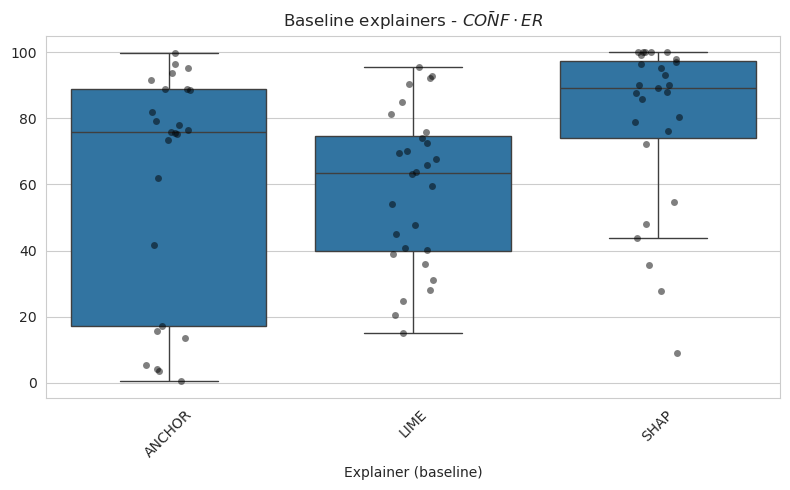}
  }

  \medskip

  \subfloat[\(\overline{\text{CONF}}\) (Average Confidence)\label{fig:single_explainers_plots__conf}]{
    \includegraphics[width=0.45\textwidth]{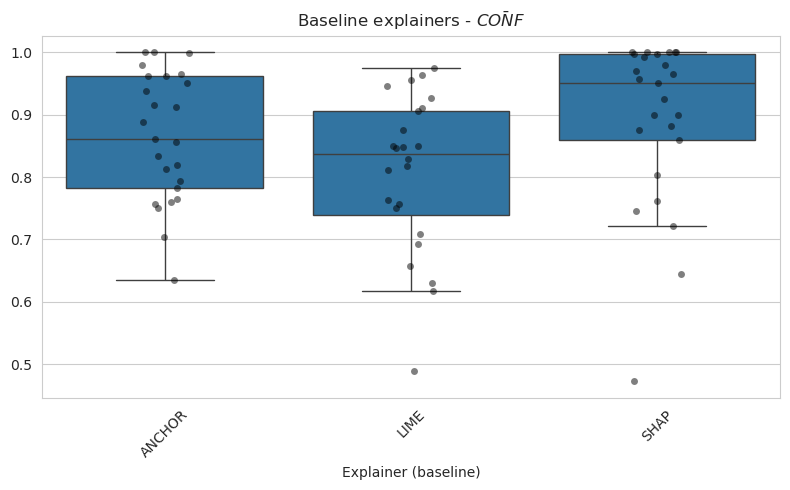}
  }\hfill
  \subfloat[\(\overline{\text{COV}}\) (Average Coverage)\label{fig:single_explainers_plots__cov}]{
    \includegraphics[width=0.45\textwidth]{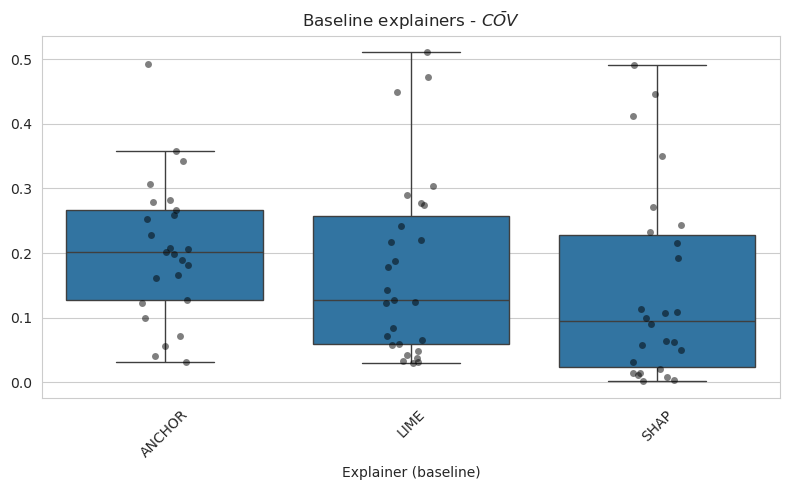}
  }

  \medskip

  \subfloat[\(ER\) (Explained Ratio)\label{fig:single_explainers_plots__er}]{
    \includegraphics[width=0.45\textwidth]{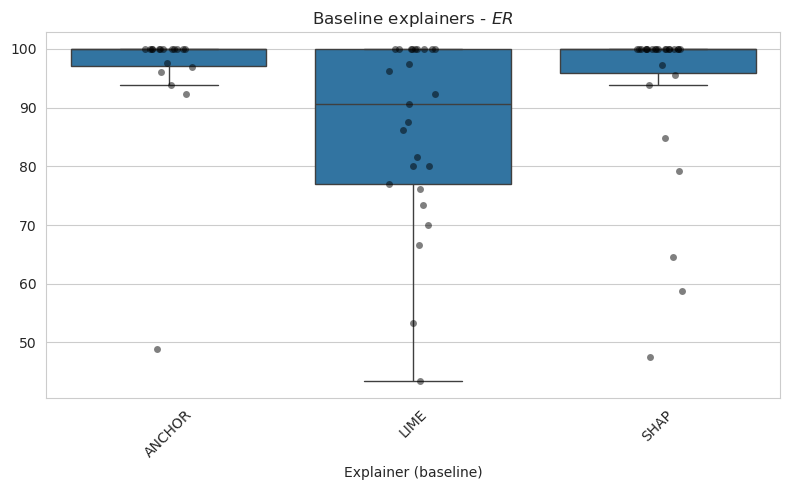}
  }\hfill
  \subfloat[\(\overline{F(n)}\) (Average Feature Count)\label{fig:single_explainers_plots__avg_f_count}]{
    \includegraphics[width=0.45\textwidth]{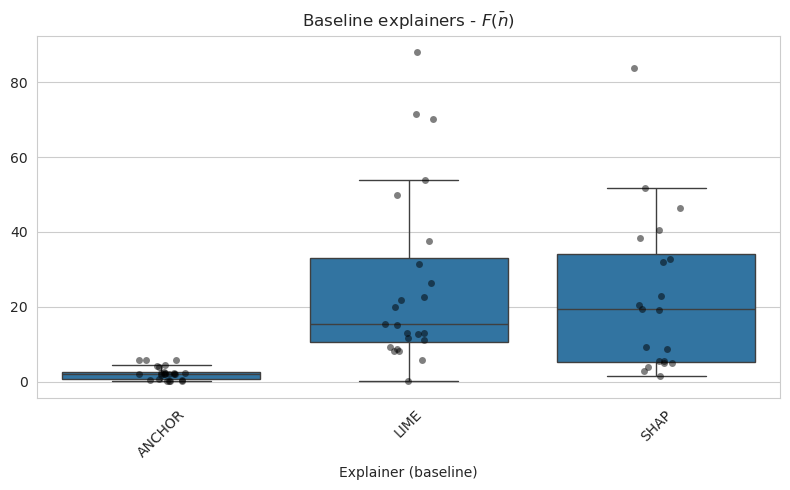}
  }

  \caption{Performance comparison of baseline (single) explainers across key metrics. The metrics displayed are: \(\bar{M}\) (Objective function), \(\overline{\text{CONF}} \cdot ER\) (Average Confidence * Explained Ratio), \(\overline{\text{CONF}}\) (Average Confidence), \(\overline{\text{COV}}\) (Average Coverage), \(ER\) (Explained Ratio), and \(\overline{F(n)}\) (Average Feature Count). Each plot visualizes the metric distribution for Anchor, LIME, and SHAP under the baseline condition.}
  \hypertarget{fig:single_explainers_plots__all}{}
\end{figure}

\FloatBarrier \subsubsection{Comparison of Baselines and Fusion Explainers}
\begin{figure*}[ht]
  \centering

  \subfloat[Objective function performance (\(\bar{M}\)).\label{fig:merged_boxplots_objective_fun}]{
    \includegraphics[width=0.55\textwidth]{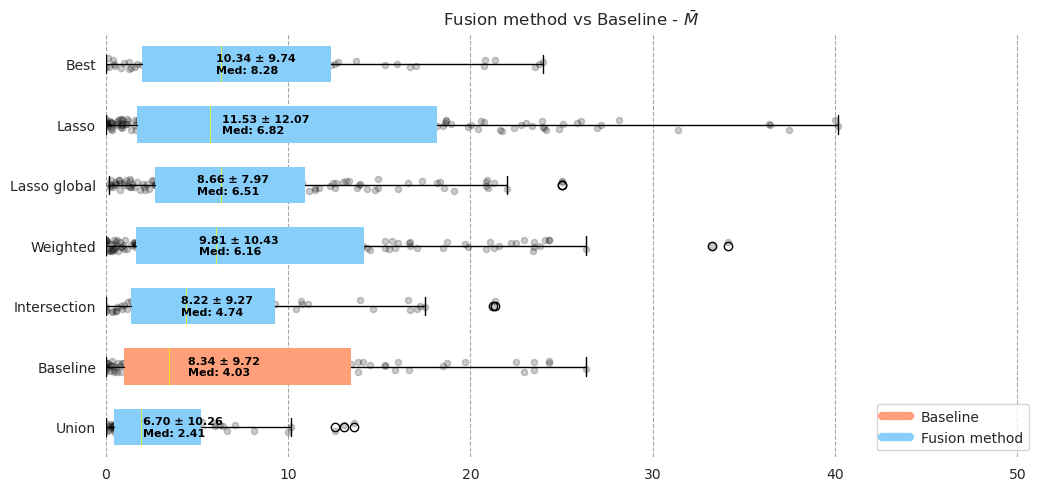}
  }\par\medskip

  \subfloat[Combined metric \(\overline{\text{CONF}} \cdot ER\) (Confidence \(\cdot\) Explained Ratio).\label{fig:merged_boxplots_conf_er}]{
    \includegraphics[width=0.55\textwidth]{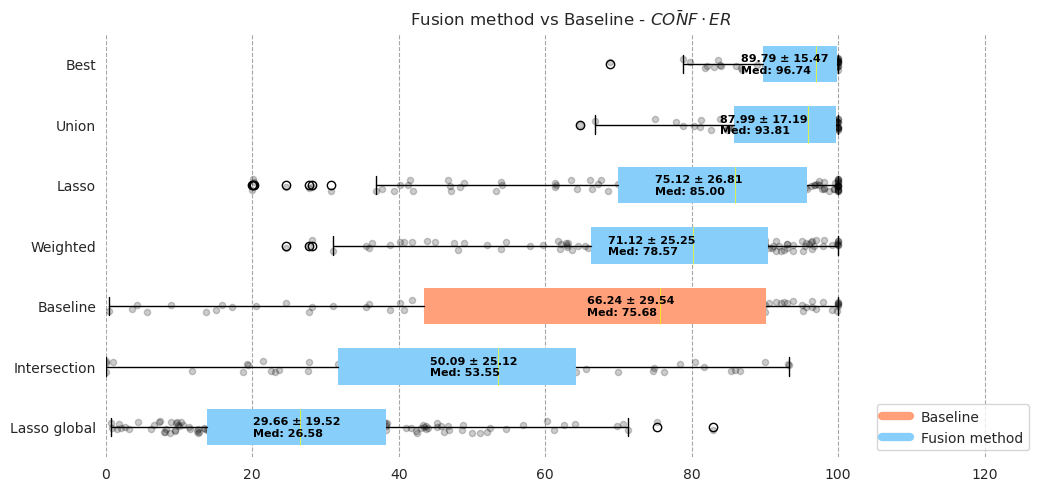}
  }\par\medskip

  \subfloat[Metric \(\overline{\text{COV}}\) (Coverage).\label{fig:merged_boxplots_cov}]{
    \includegraphics[width=0.55\textwidth]{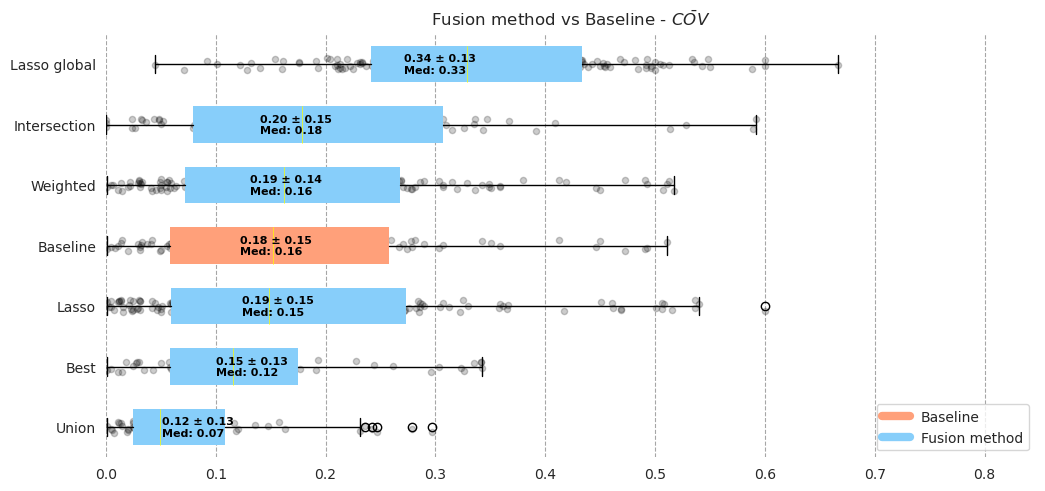}
  }\par\medskip

  \subfloat[Average feature count (\(\bar{F(n)}\)).\label{fig:merged_boxplots_avg_features_count}]{
    \includegraphics[width=0.55\textwidth]{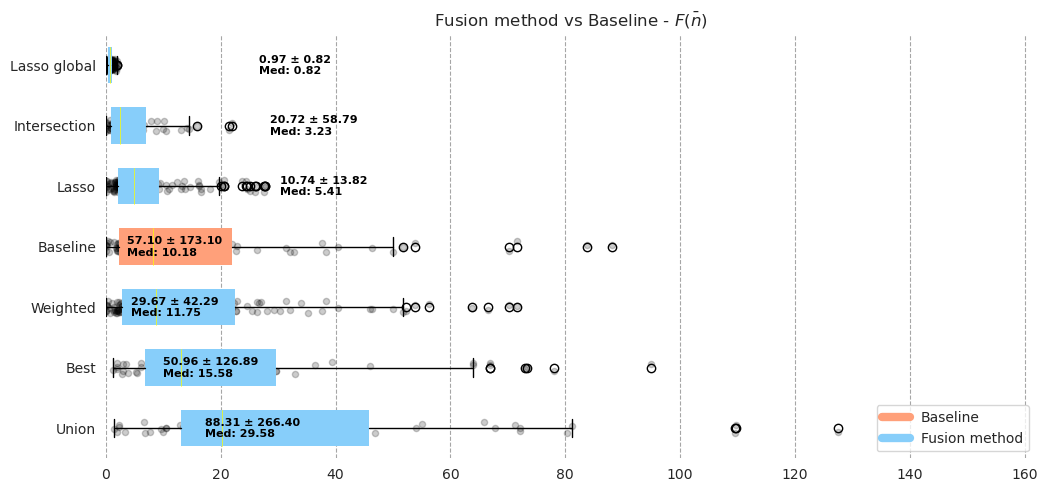}
  }

  \caption{Single-explainer baselines (orange) vs fusion methods (blue). Boxplots are annotated with mean, standard deviation, and median. Circles denote outliers.}
  \hypertarget{fig:merged_boxplots_1}{}
\end{figure*}

\begin{figure*}[ht]
    \centering
    \includegraphics[width=1.00\textwidth]{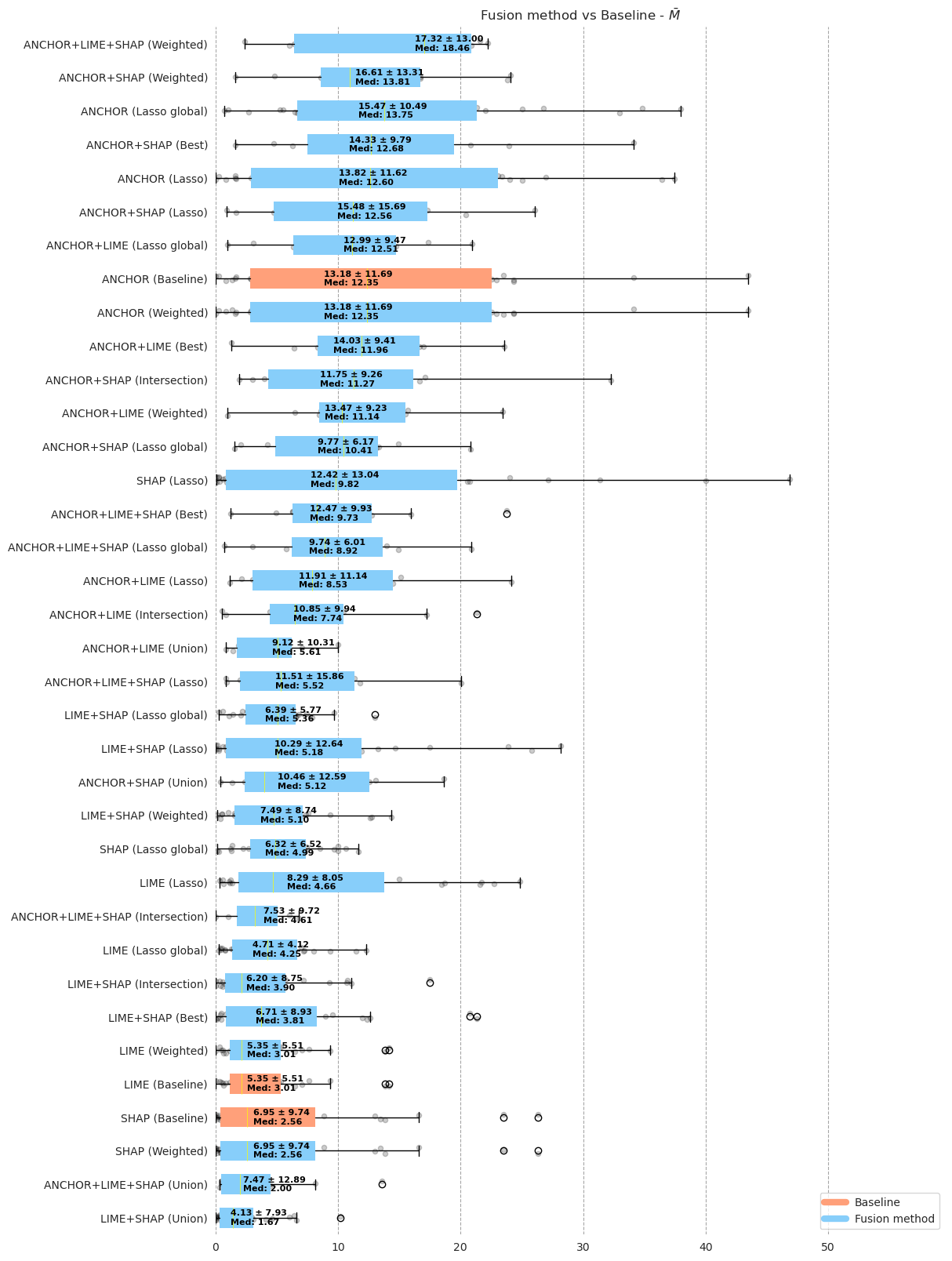}
    \phantomsection
    \caption{Objective function performance ($\bar{M}$): Comparison of single-explainer baselines (orange) and fusion methods (blue). Circles denote outliers.}
    \label{fig:ensemble_vs_single_objective_fun}
\end{figure*}

\begin{figure*}[ht]
    \centering
    \includegraphics[width=1.00\textwidth]{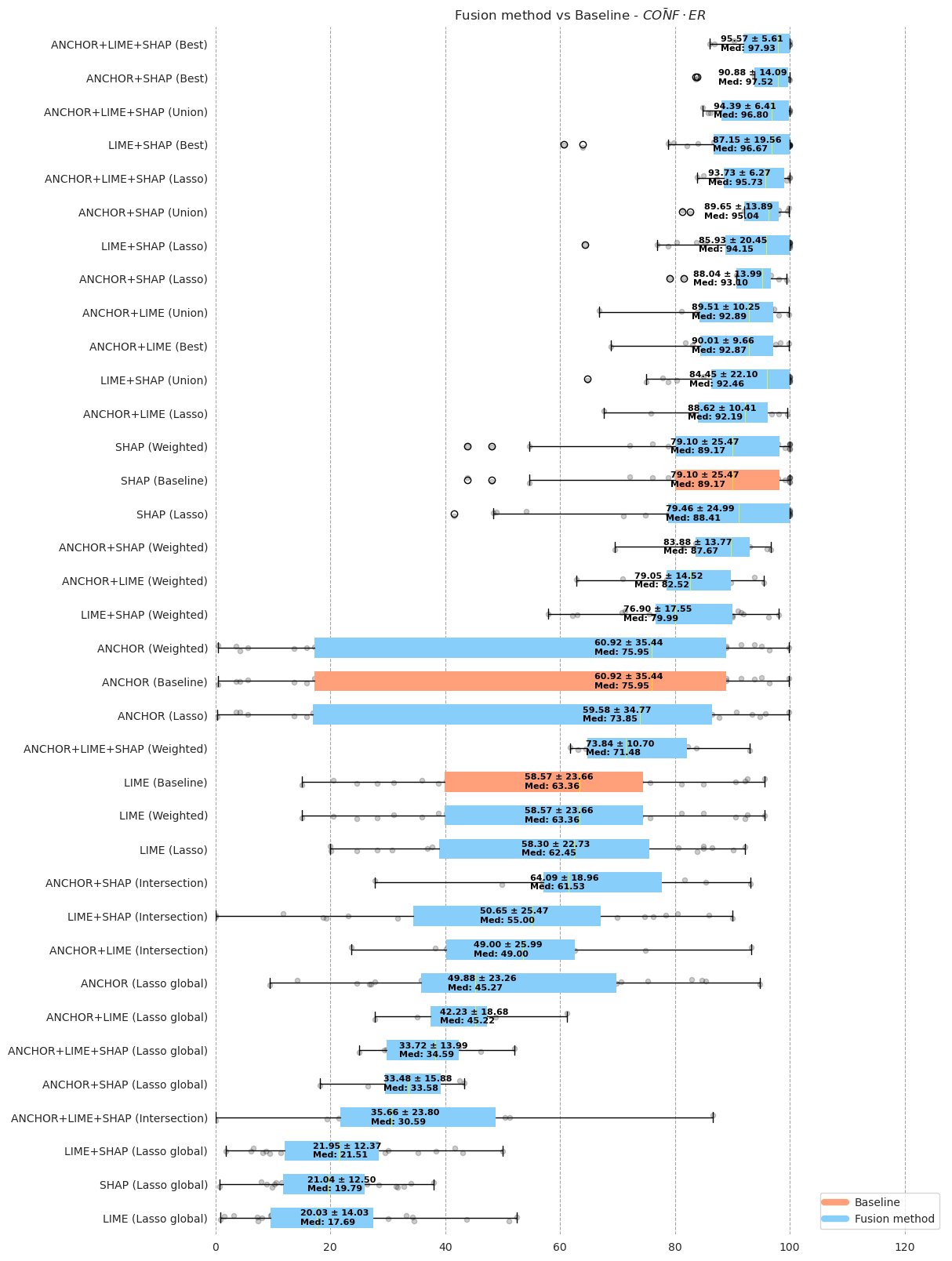}
    \caption{Combined metric $\overline{\text{CONF}} \cdot ER$ (Confidence $\cdot$ Explained Ratio): Comparison of single-explainer baselines (orange) and fusion methods (blue). Circles denote outliers.}
    \hypertarget{fig:ensemble_vs_single_conf_er}{}
\end{figure*}

\begin{figure*}[ht]
    \centering
    \includegraphics[width=1.00\textwidth]{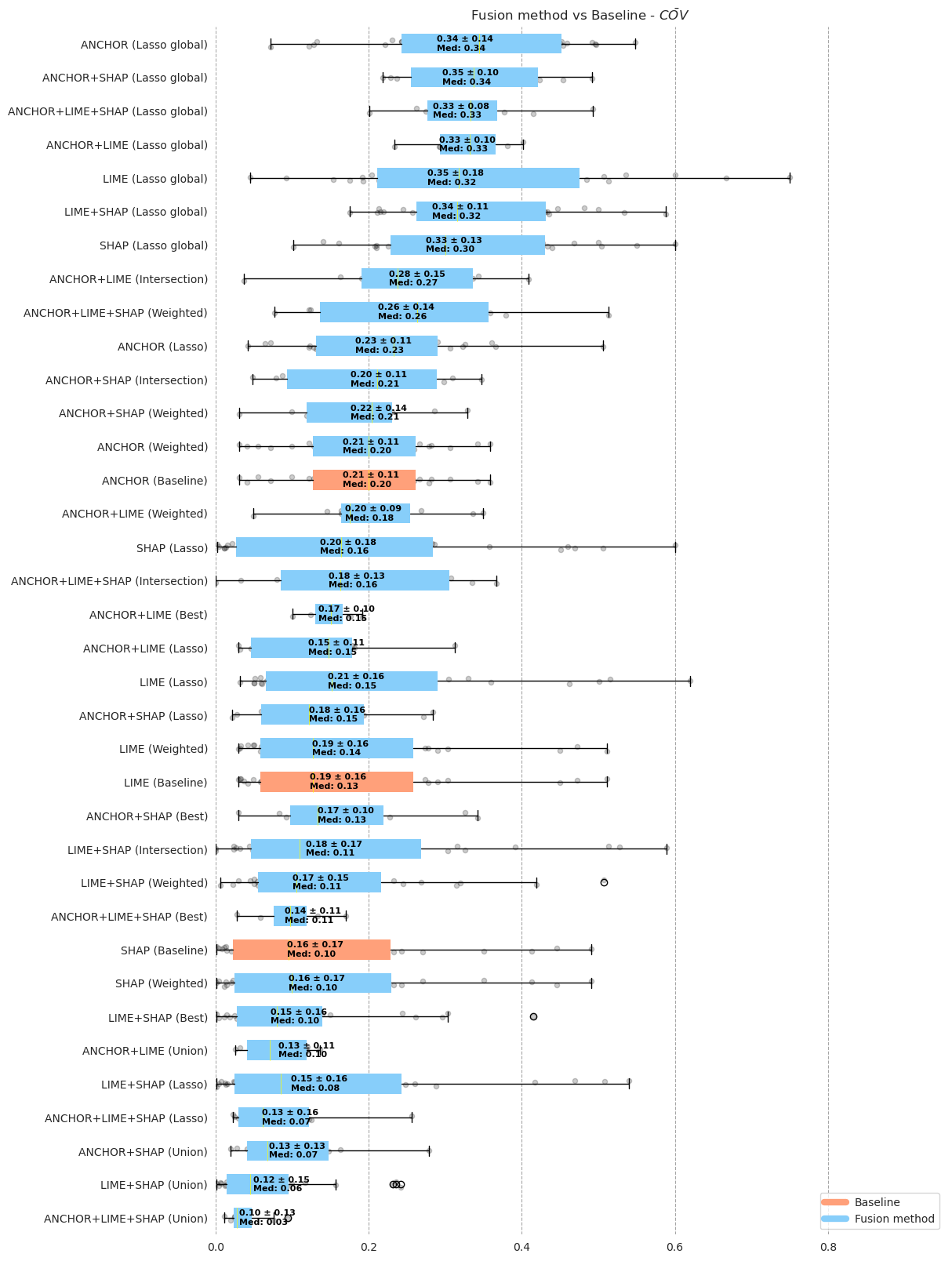}
    \caption{Metric $\overline{\text{COV}}$ (Coverage): Comparison of coverage metrics for single-explainer baselines (orange) and fusion methods (blue). Circles denote outliers.}
    \hypertarget{fig:ensemble_vs_single_cov}{}
\end{figure*}

\begin{figure*}[ht]
    \centering
    \includegraphics[width=1.00\textwidth]{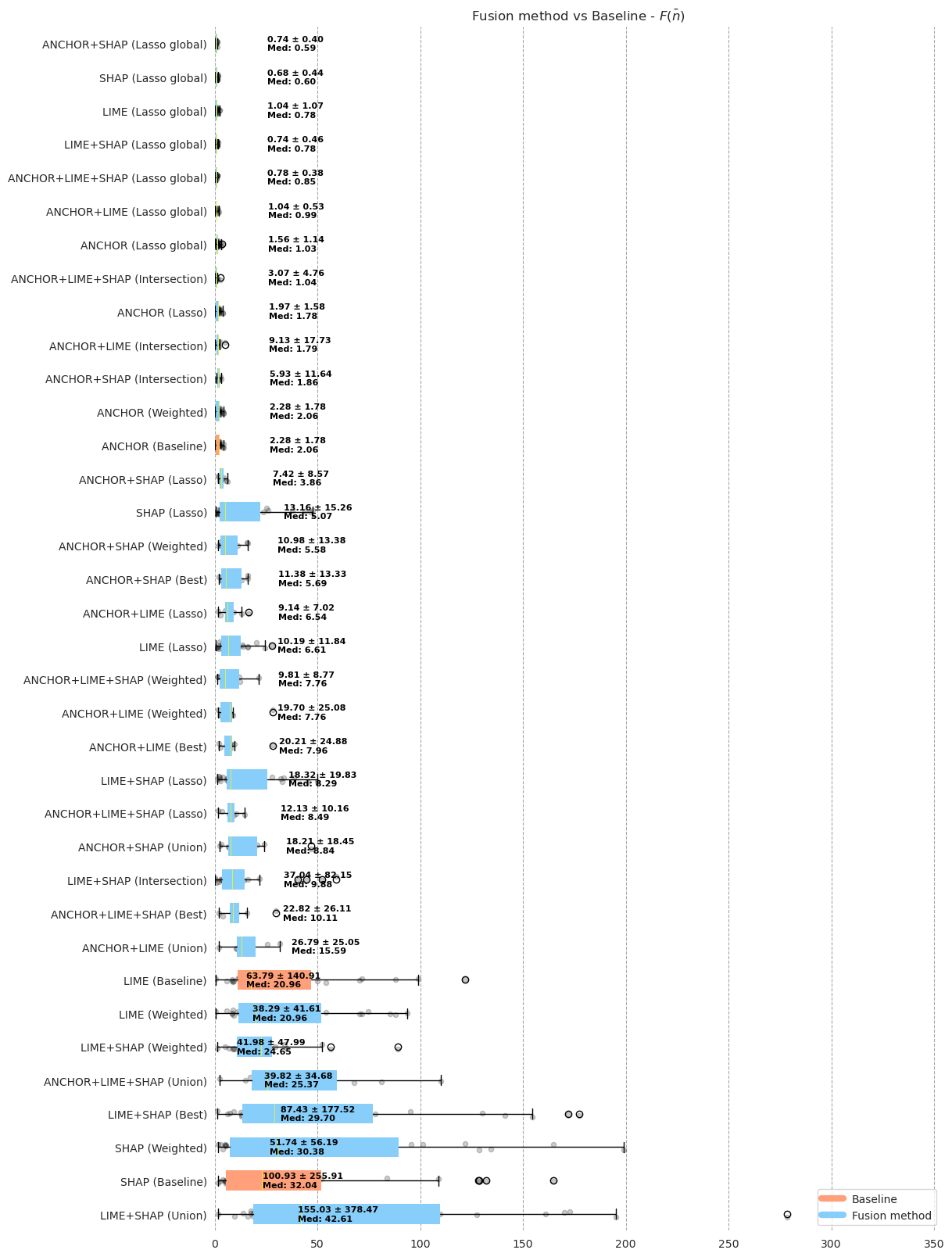}
    \caption{Average feature count ($\bar{F(n)}$): Distribution of feature counts used by single-explainer baselines (orange) and fusion methods (blue). Circles denote outliers.}
    \hypertarget{fig:ensemble_vs_single_avg_features_count}{}
\end{figure*}

\FloatBarrier \subsubsection{Additional Critical Difference Diagrams}

\begin{figure*}[ht]
  \centering

  \subfloat[Critical Difference Diagram for \(\bar{M}\) (Final Metric).\label{fig:cd_final_metric_ens_and_m}]{
    \includegraphics[width=0.75\textwidth]{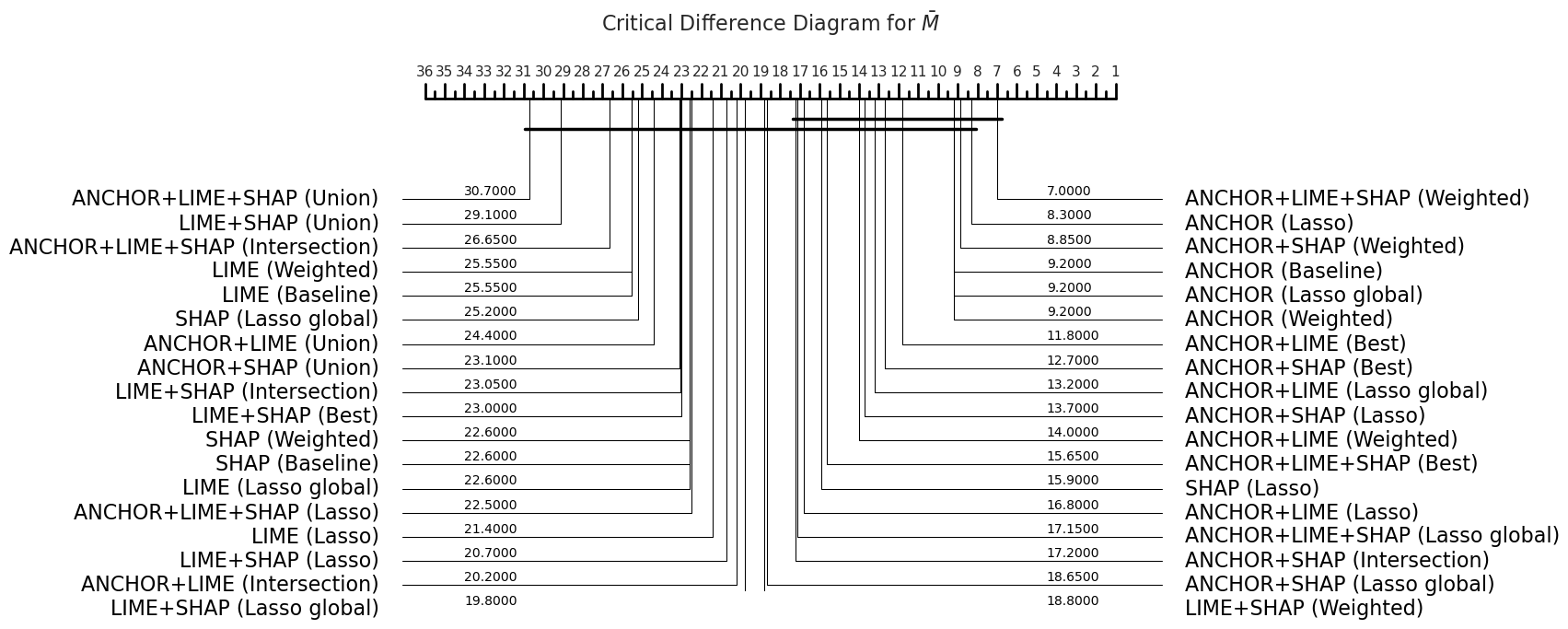}
  }\par\medskip

  \subfloat[Critical Difference Diagram for \(ER\) (Explained Ratio).\label{fig:cd_explained_ratio_ens_and_m}]{
    \includegraphics[width=0.75\textwidth]{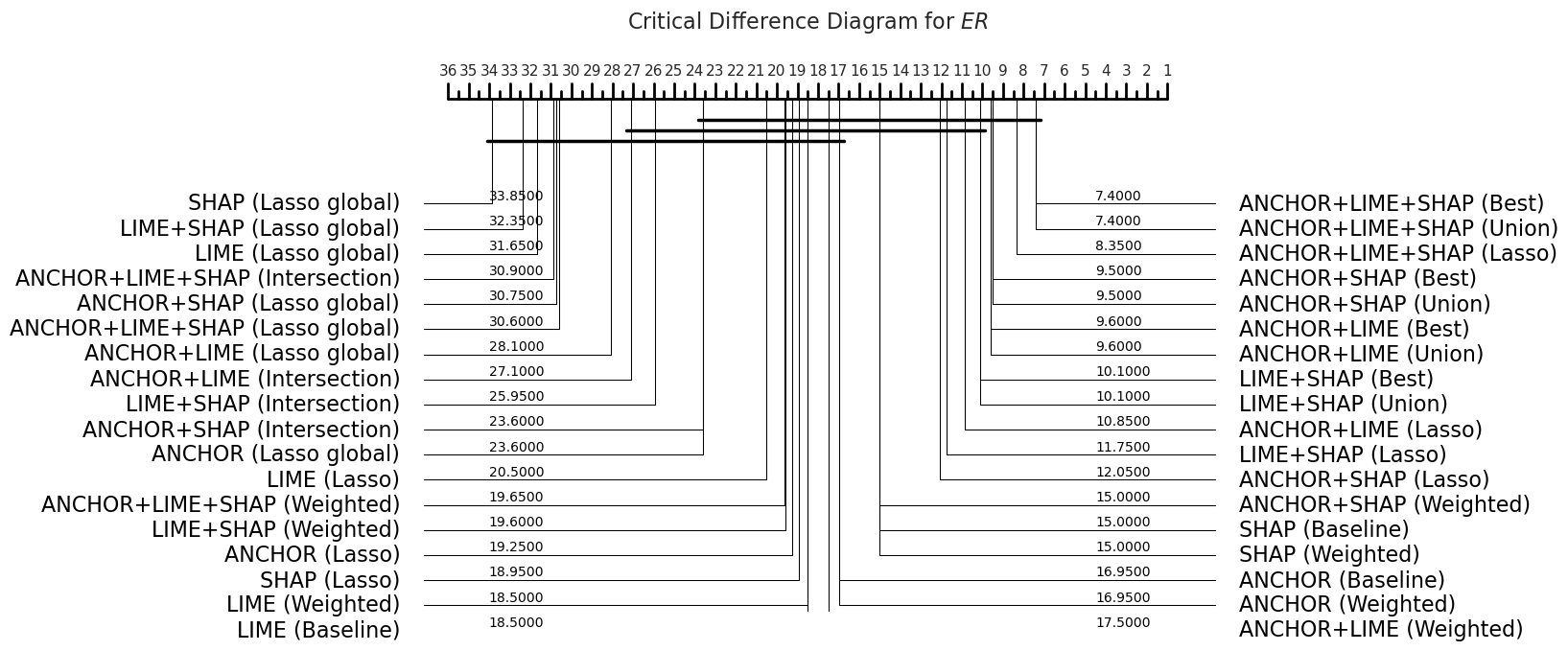}
  }\par\medskip

  \subfloat[Critical Difference Diagram for \(F(\bar{n})\) (Average Feature Count).\label{fig:cd_avg_features_ens_and_m}]{
    \includegraphics[width=0.75\textwidth]{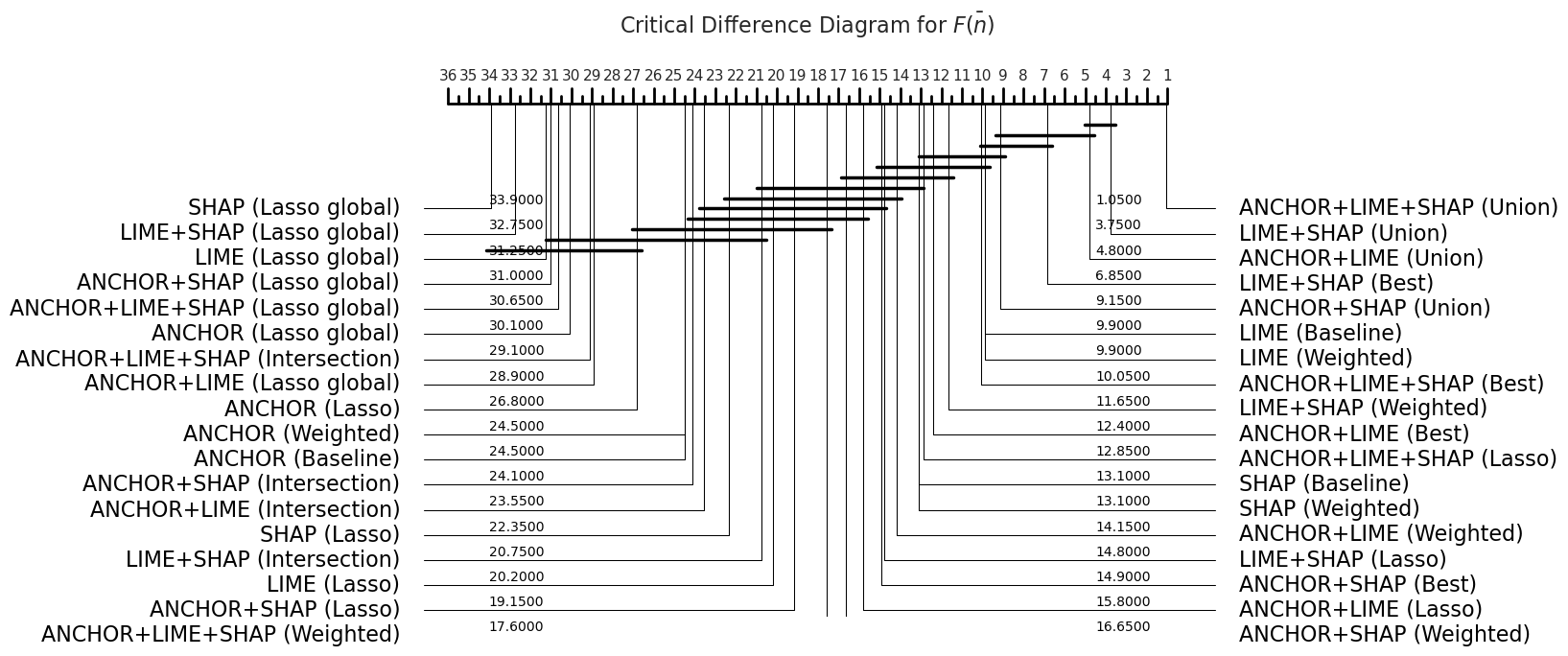}
  }

  \caption{Critical Difference Diagrams for different fusion methods and baselines across metrics. Subfigure (a) shows \(\bar{M}\) (Final Metric), (b) illustrates \(ER\) (Explained Ratio), and (c) presents \(F(\bar{n})\) (Average Feature Count).}
  \label{fig:critical_diff_all}
\end{figure*}
\FloatBarrier 

\subsection{How to read the rules.} \label{app:how_to_read_rule_figures}
Each panel shows the \emph{raw} time series. 
Red highlights mark time points that satisfy a rule. 
Dashed vertical lines indicate the start and end of an interval \([\,\tau_{1},\tau_{2}\,]\). 

For univariate data, a rule is a finite conjunction of point-wise constraints on selected indices \(t_{j_k}\), each constraining the raw value to a half-open interval: \(x_{t_{j_k}} \in (\ell_k, u_k]\). 
For multivariate data, we indicate the feature/channel by a superscript: \(x^{(c_{j_k})}_{t_{j_k}} \in (\ell_k, u_k]\). 
Intervals are obtained by our numeric-to-rules procedure (see Section~\ref{sec:numeric_to_rules}). 
\[
R_{i}:\quad
\left\{
\begin{aligned}
& x_{t_{j_1}} \in (\ell_{1},u_{1}] \\
& x_{t_{j_2}} \in (\ell_{2},u_{2}] \\
& \;\vdots \\
& x_{t_{j_m}} \in (\ell_{m},u_{m}]
\end{aligned}
\right.
\;\Rightarrow\;
\begin{aligned}
\hat y &= \alpha,\\
\mathrm{CONF}(R_{i}) &= \gamma,\\
\mathrm{COV}(R_{i}) &= \kappa.
\end{aligned}
\]

We report \emph{confidence} (precision) and \emph{coverage} for each rule. 
Rules are \emph{human-readable proxies of model behaviour}: they show \emph{where} and \emph{what} the classifier relied on, but they are not causal claims.
Prototypes and class averages, when shown, give shape context. 
The figures are \emph{actionable}: a practitioner can verify the inequality on the highlighted segment or inspect the waveform there. 

\subsection{Dataset-level exemplar visualisations} \label{app:visual_appendix}
\subsubsection{Dataset-level exemplar visualisations for \emph{ECG200}} \label{app:ecg200_visual_catalog}
Visualizations intuitively link numeric feature attribution explanations to \emph{TS} structure, aiding interpretability.  
Our framework transforms numeric explanations into rules consisting of features and intervals of their applicability, which can be visualized on \emph{TS plots} and aid revealing patterns and decision-making.

\begin{figure*}[ht]
  \centering

  \subfloat[ECG200 with Anchor (baseline)\label{fig:ecg200_anchor_bench}]{
    \includegraphics[width=\textwidth]{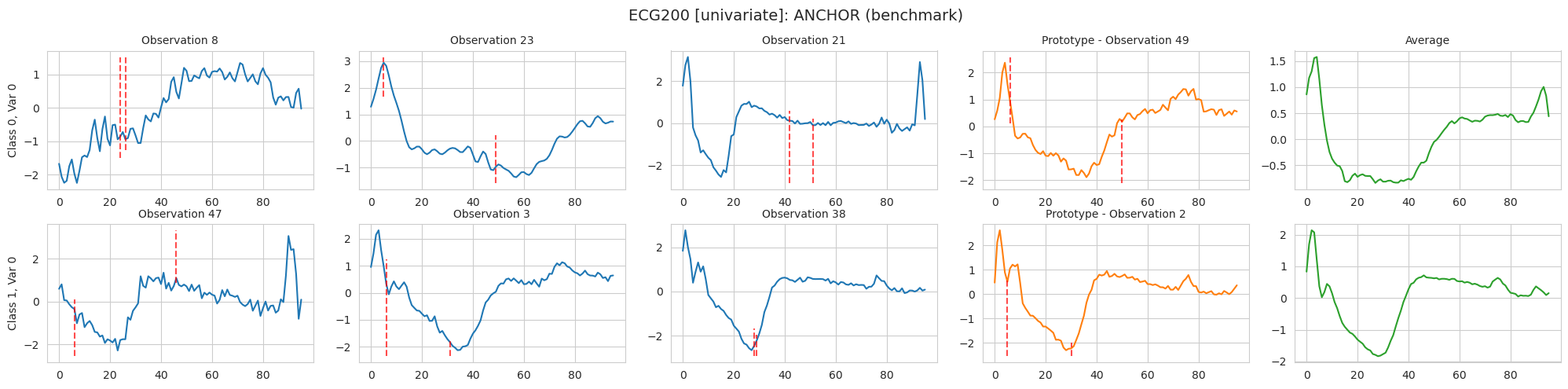}
  }

  \medskip

  \subfloat[ECG200 with LIME (baseline)\label{fig:ecg200_lime_bench}]{
    \includegraphics[width=\textwidth]{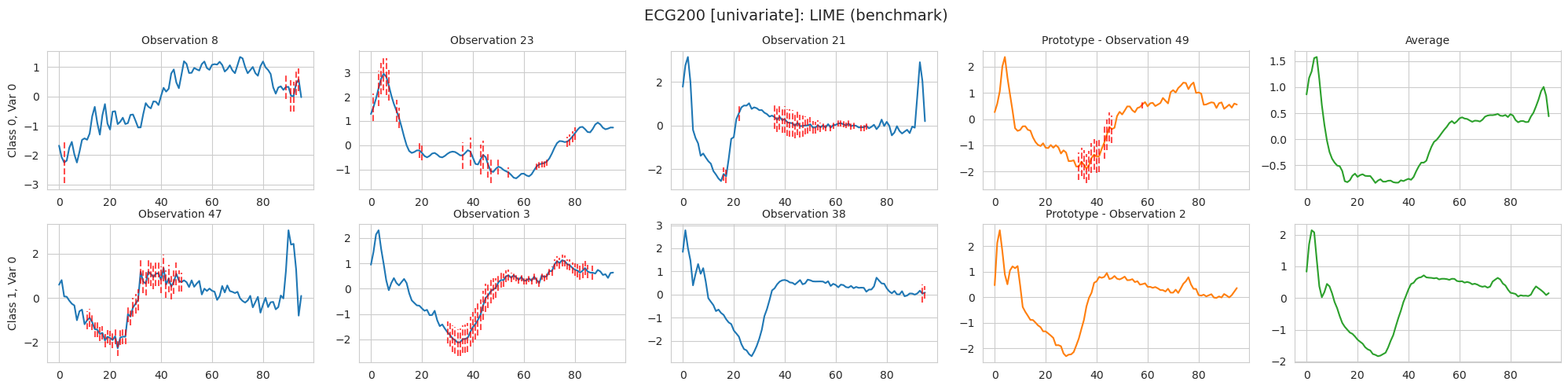}
  }

  \medskip

  \subfloat[ECG200 with SHAP (baseline)\label{fig:ecg200_shap_bench}]{
    \includegraphics[width=\textwidth]{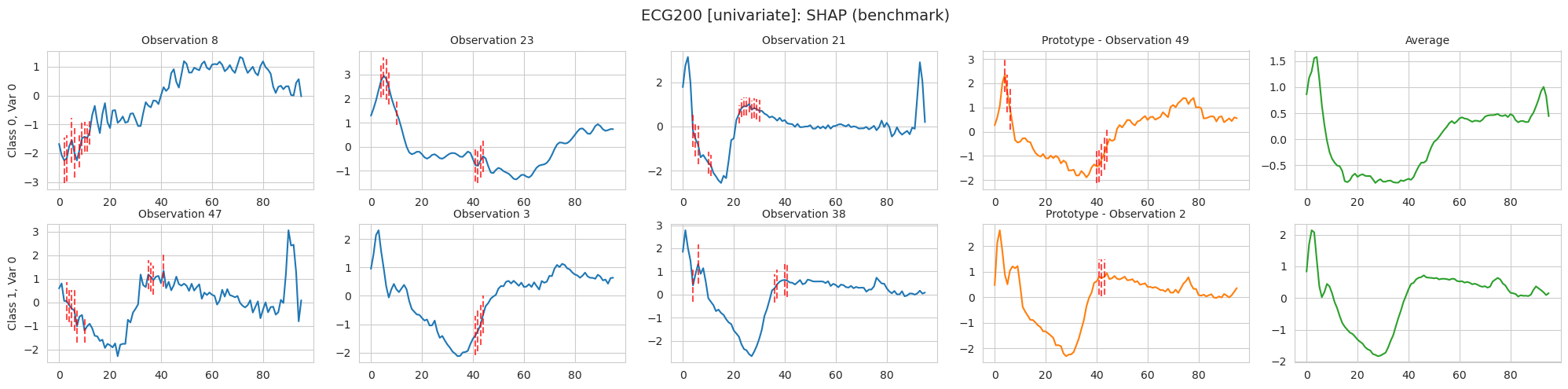}
  }

  \medskip

  \subfloat[ECG200 with Anchor+LIME+SHAP (Lasso fusion)\label{fig:ecg200_anchor_lime_shap_lasso}]{
    \includegraphics[width=\textwidth]{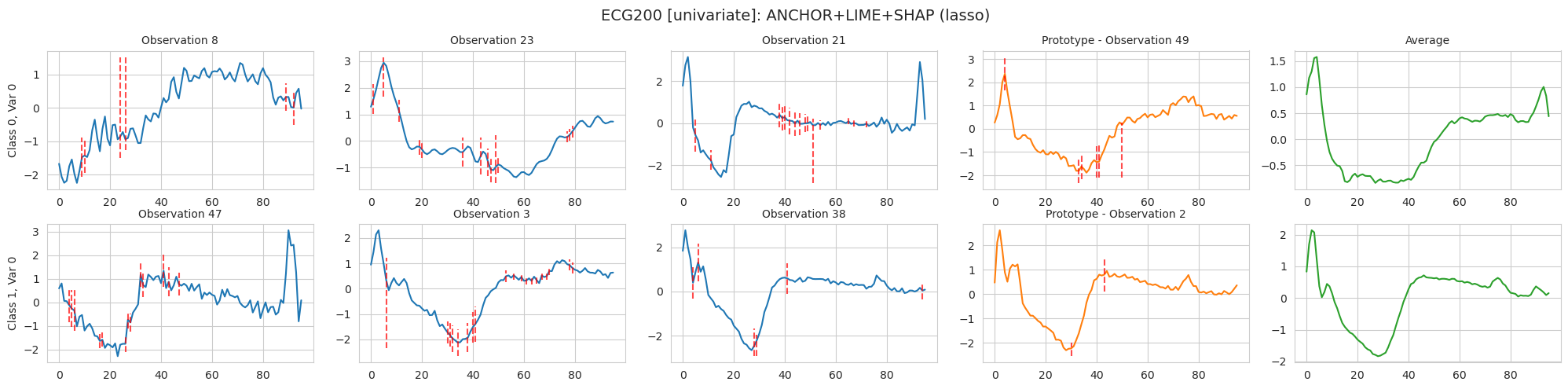}
  }

  \caption{Visual explanations for univariate ECG200 data (two classes).}
  \label{fig:ecg200_rule_explanations}
\end{figure*}

Figure~\ref{fig:ecg200_rule_explanations} shows \emph{LIME} and \emph{SHAP} explanations converted into rule-based intervals for \emph{ECG200} dataset (\emph{univariate}, two classes), as well as \emph{Anchor} (native rule-based).  
Each subplot consists of five plots per row, where each row represents visualizations for one class. 
The next row contains the second class. 
The first three columns display green-plotted instances that are maximally diverse, while the fourth column represents the most average, prototypical observation (orange-plotted). 
The last (green) column shows the class average.
Red vertical lines mark feature intervals where time points satisfy rules. 
\emph{LIME} targets mainly the initial waveform sections, whereas \emph{SHAP} highlights central intervals.
\emph{Anchor+LIME+SHAP Lasso fusion} merges both methods, improving coverage and confidence ratio.  
This slightly increases the median feature count from 4 to 5.  
Notably, \emph{Observation2} in \emph{Class1}, under-unexplained by \emph{LIME}, is covered by \texttt{fusion}.  
Despite added complexity, \texttt{fusion} remains computationally efficient for \emph{TS}.  

\subsubsection{Dataset-level exemplar visualisations for \emph{Plane}} \label{app:plane_visual_catalog}
\begin{figure*}[ht]
    \centering
    \includegraphics[width=1.1\textwidth]{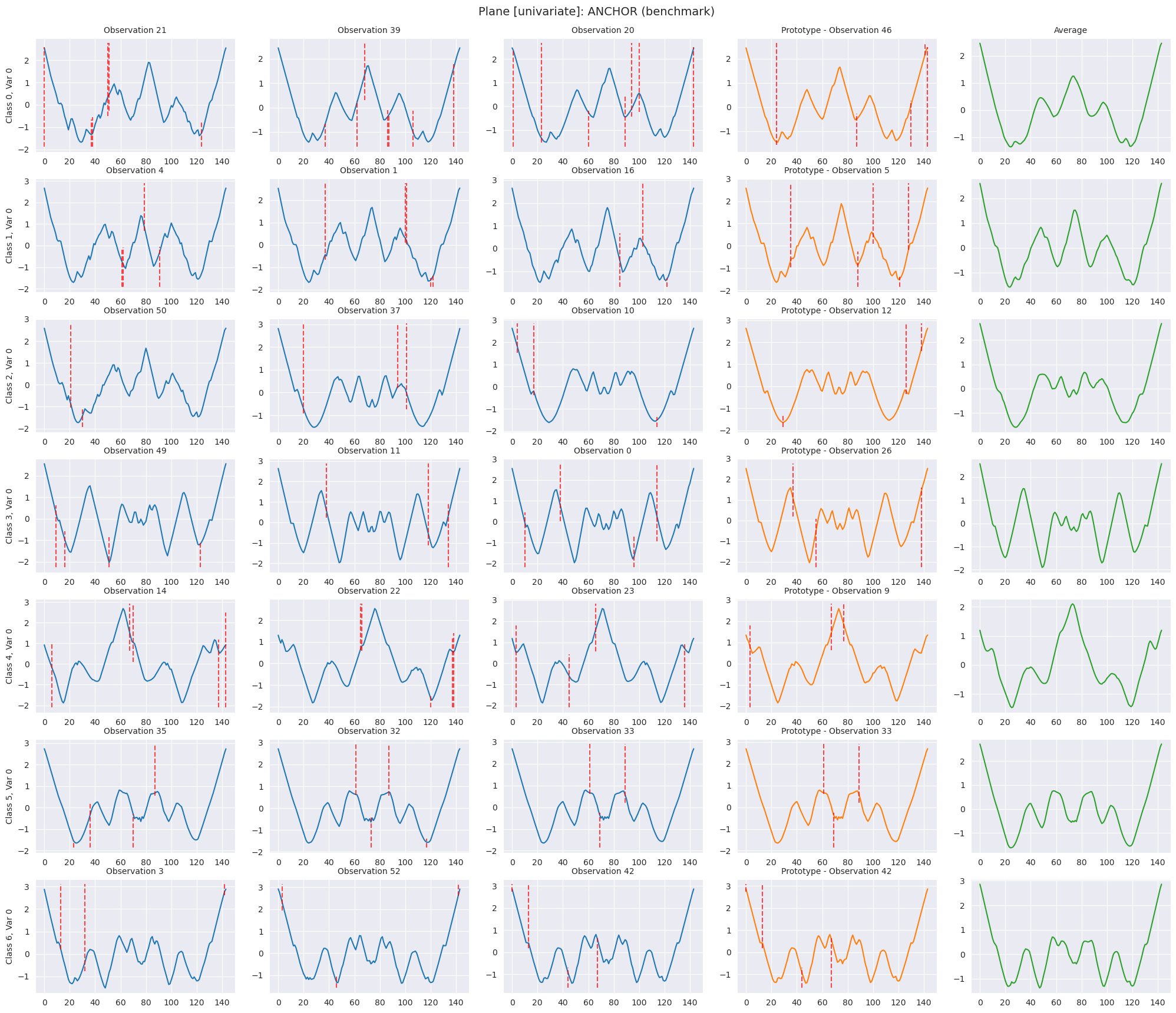}
    \caption{Plane dataset with Anchor (baseline). }
    \label{fig:plane_anchor_benchmark}
\end{figure*}

\begin{figure*}[ht]
    \centering
    \includegraphics[width=1.1\textwidth]{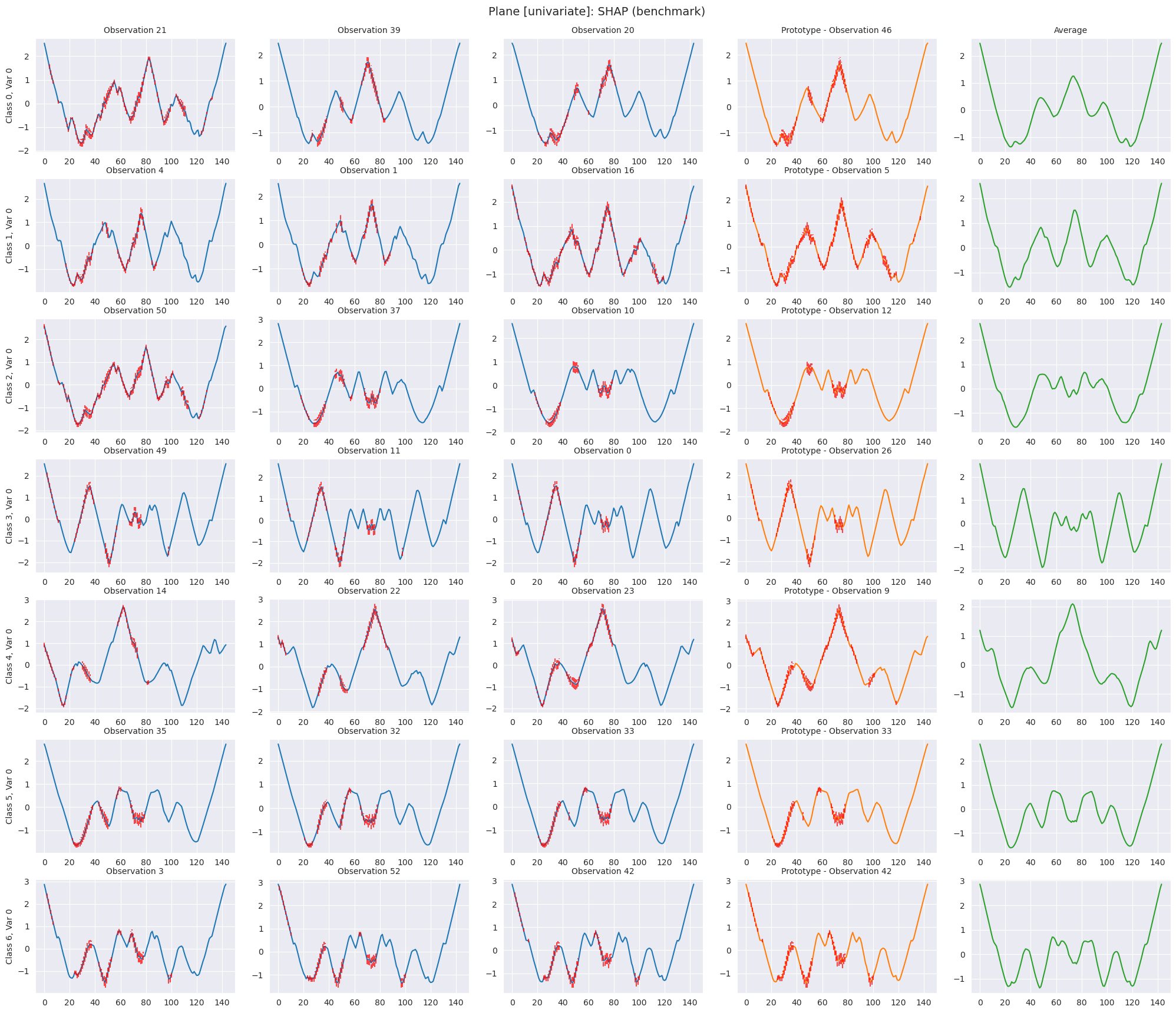}
    \caption{Plane dataset with SHAP (baseline). }
    \label{fig:plane_shap_benchmark}
\end{figure*}

\begin{figure*}[ht]
    \centering
    \includegraphics[width=1.1\textwidth]{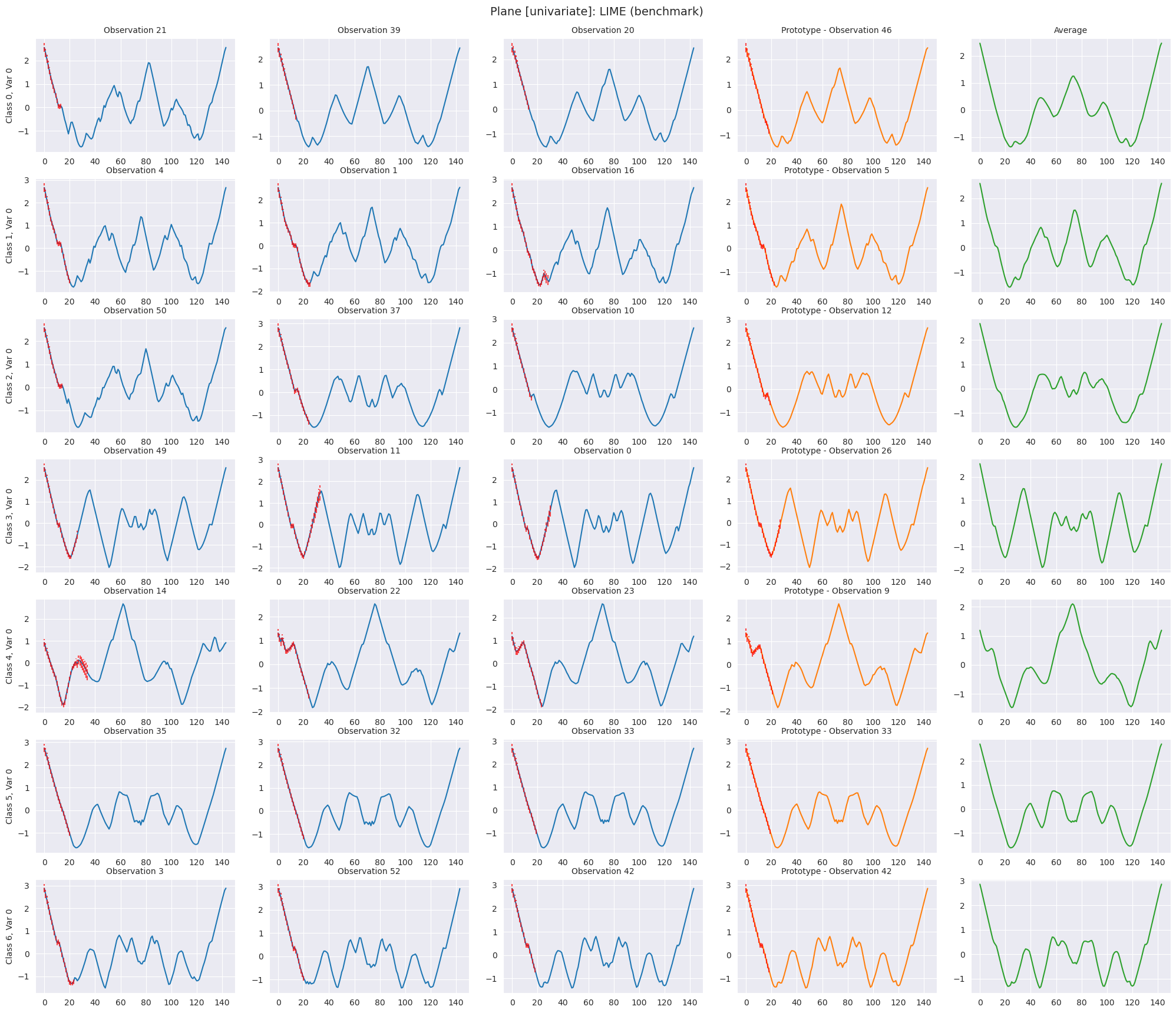}
    \caption{Plane dataset with LIME (baseline). }
    \label{fig:plane_lime_benchmark}
\end{figure*}

\begin{figure*}[ht]
    \centering
    \includegraphics[width=1.1\textwidth]{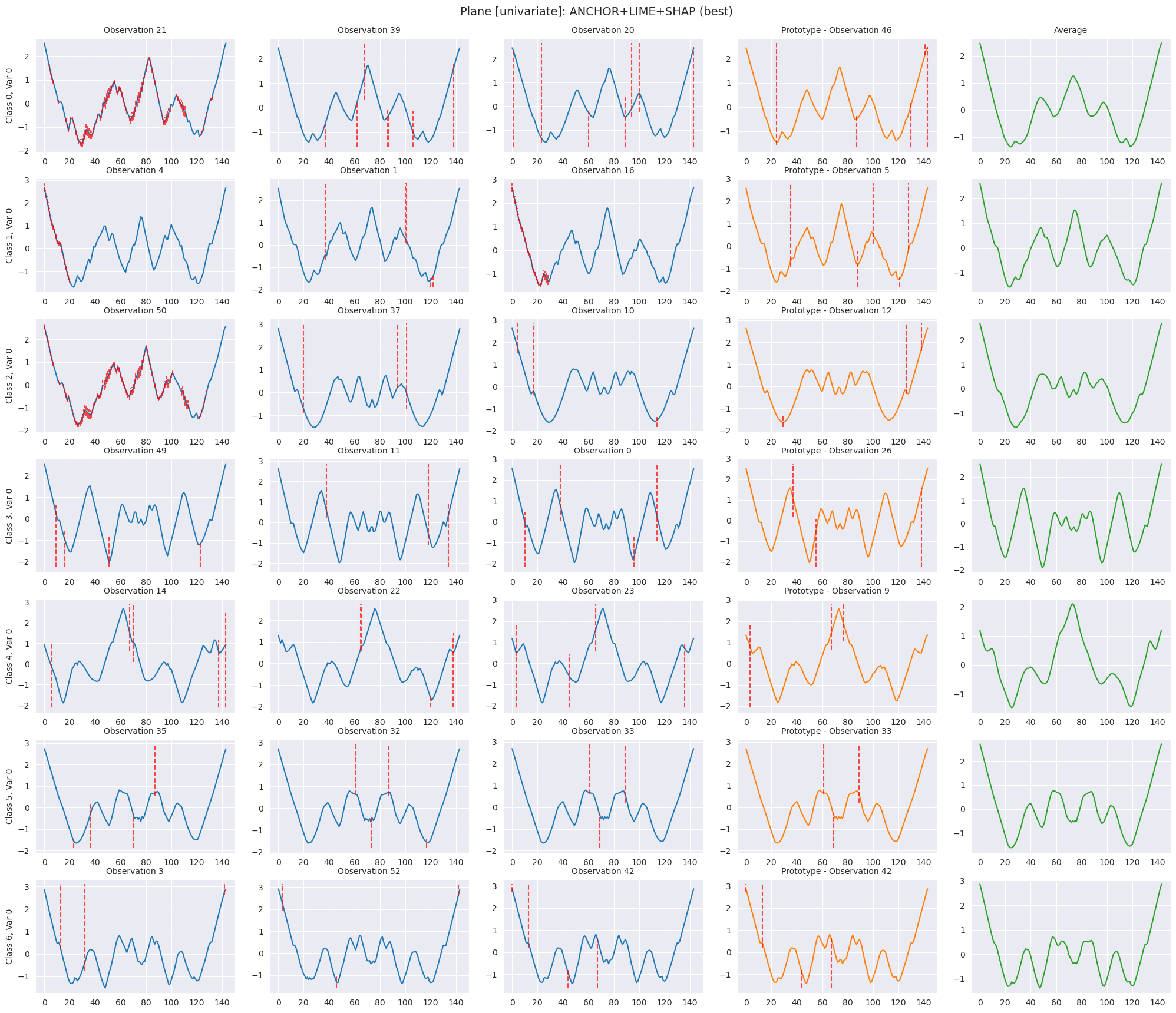}
    \caption{Plane dataset with Anchor+LIME+SHAP (Best Combination). }
    \label{fig:plane_best_combination}
\end{figure*}

The \emph{Plane} dataset (\emph{univariate}, 6 classes) highlights differences in rule intervals from \emph{Anchor}, \emph{SHAP}, \emph{LIME}, and their \texttt{fusion} (Figures~\ref{fig:plane_anchor_benchmark}-\ref{fig:plane_best_combination}, Appendix).  
\emph{Anchor} uses fewer variables, but its red-marked \emph{intervals are often too wide}.  
This reduces informativeness, as broad intervals fail to capture nuances of \emph{TS} shape.  
\emph{SHAP} and \emph{LIME} use more variables, while their interval method (Section~\ref{sec:numeric_to_rules}) generates \emph{narrower ranges} via multidimensional perturbations.  
Narrower intervals may improve interpretability by focusing attention on specific waveform segments. 
However, overly narrow intervals risk overfitting. 
Our fusion step trades off precision, coverage, and simplicity to mitigate this. 
\emph{Fusing} \emph{Anchor}, \emph{SHAP}, and \emph{LIME} with \emph{"best" method} matches \emph{SHAP}'s baseline performance with significantly fewer variables.  

\setstretch{1.0}

\end{appendices}

\FloatBarrier
\setcounter{section}{1}
\renewcommand{\thesection}{\Alph{section}}
\section{Supplementary Materials}\label{sec:supplement}
\textbf{Friedman and Nemenyi:}  
To compare multiple methods statistically, we applied the Friedman test, followed by post-hoc Nemenyi comparisons.  
The Friedman test results provide insights into the statistical significance of differences across methods.  
Table~\ref{tab:friedman_method_all_metrics} reports results at the method level, considering different fusion strategies (e.g. \texttt{Baseline}, \texttt{Best}, \texttt{Intersection}, \texttt{Weighted}).  
Table~\ref{tab:friedman_rules_combination_all_metrics} presents results for all rule sets, accounting for the combinations of explainers they comprise, both single and compound.  
For pairwise comparisons, we employed the Nemenyi test.  
The first set of tables, from Table~\ref{tab:nemenyi_rules_combination_final_metric} to Table~\ref{tab:nemenyi_rules_combination_median_features_count}, evaluates fusion combinations in terms of explainer types.  
The second set, from Table~\ref{tab:nemenyi_method_final_metric} to Table~\ref{tab:nemenyi_method_median_features_count}, focuses on comparisons across fusion methods.  

\subsection{Friedman Tests}
\begin{table}[ht]
\centering
\caption{Friedman test results for all fusion methods (e.g. \texttt{Baseline}, \texttt{Best}, \texttt{Intersection}, ... , \texttt{Weighted}) across metrics.}
\label{tab:friedman_method_all_metrics}
\begin{tabular}{lrr}
\toprule
\textbf{Metric} & \(\chi^2\) & \(p\) \\
\midrule
$\bar{M}$ & 41.46 &< 0.001$^{***}$ \\
$\bar{{CONF}} \cdot {ER}$ & 130.10 &< 0.001$^{***}$ \\
$\bar{{CONF}} \cdot \bar{{COV}} \cdot {ER}$ & 29.49 &< 0.001$^{***}$ \\
$\bar{{CONF}}$ & 95.69 &< 0.001$^{***}$ \\
$\bar{{COV}}$ & 110.02 &< 0.001$^{***}$ \\
${{ER}}$ & 126.00 &< 0.001$^{***}$ \\
$\bar{F(n)}$ & 136.22 &< 0.001$^{***}$ \\
${Med}(F(n))$ & 128.66 &< 0.001$^{***}$ \\
\multicolumn{3}{l}{\scriptsize *p(<0.05), **p(<0.01), ***p(<0.001)} \\
\bottomrule
\end{tabular}
\end{table}
\begin{table}[ht]
\centering
\caption{Friedman test results for all fused rule sets (taking into account explainers combinations they are composed of, e.g. \emph{Anchor}, \emph{Anchor+LIME}, \emph{Anchor+LIME+SHAP}, etc.) across metrics.}
\label{tab:friedman_rules_combination_all_metrics}
\begin{tabular}{lrr}
\toprule
\textbf{Metric} & \(\chi^2\) & \(p\) \\
\midrule
$\bar{M}$ & 36.09 &< 0.001$^{***}$ \\
$\bar{{CONF}} \cdot {ER}$ & 17.31 &0.008$^{**}$ \\
$\bar{{CONF}} \cdot \bar{{COV}} \cdot {ER}$ & 30.64 &< 0.001$^{***}$ \\
$\bar{{CONF}}$ & 19.16 &0.004$^{**}$ \\
$\bar{{COV}}$ & 13.50 &0.036$^{*}$ \\
${{ER}}$ & 13.41 &0.037$^{*}$ \\
$\bar{F(n)}$ & 27.26 &< 0.001$^{***}$ \\
${Med}(F(n))$ & 28.03 &< 0.001$^{***}$ \\
\multicolumn{3}{l}{\scriptsize *p(<0.05), **p(<0.01), ***p(<0.001)} \\
\bottomrule
\end{tabular}
\end{table}

\FloatBarrier

\clearpage \subsection{Nemenyi Tests}
\begin{table}[ht]
\centering
\caption{Nemenyi test results for fusion combinations for $\bar{M}$. P-values for pairwise comparisons.}
\label{tab:nemenyi_rules_combination_final_metric}
\tiny
\begin{tabular}{lrrrrrrr}
\toprule
\textbf{Rules Combination} & Anchor & Anchor+LIME & Anchor+LIME+SHAP & Anchor+SHAP & LIME & LIME+SHAP & SHAP \\
\midrule
Anchor & 1.00 & 0.25 &\textbf{0.023*}& 0.69 &\textbf{0.001**}&\textbf{0.001**}&\textbf{0.002**} \\
Anchor+LIME & 0.25 & 1.00 & 0.90 & 0.90 & 0.13 & 0.31 & 0.63 \\
Anchor+LIME+SHAP &\textbf{0.023*}& 0.90 & 1.00 & 0.63 & 0.63 & 0.87 & 0.90 \\
Anchor+SHAP & 0.69 & 0.90 & 0.63 & 1.00 &\textbf{0.016*}& 0.06 & 0.21 \\
LIME &\textbf{0.001**}& 0.13 & 0.63 &\textbf{0.016*}& 1.00 & 0.90 & 0.90 \\
LIME+SHAP &\textbf{0.001**}& 0.31 & 0.87 & 0.06 & 0.90 & 1.00 & 0.90 \\
SHAP &\textbf{0.002**}& 0.63 & 0.90 & 0.21 & 0.90 & 0.90 & 1.00 \\
\bottomrule
\multicolumn{8}{l}{\scriptsize *p(<0.05), **p(<0.01), ***p(<0.001)} \\
\end{tabular}
\end{table}
\begin{table}[ht]
\centering
\caption{Nemenyi test results for fusion combinations for $\bar{{CONF}} \cdot {ER}$. P-values for pairwise comparisons.}
\label{tab:nemenyi_rules_combination_metric_con_xratio}
\tiny
\begin{tabular}{lrrrrrrr}
\toprule
\textbf{Rules Combination} & Anchor & Anchor+LIME & Anchor+LIME+SHAP & Anchor+SHAP & LIME & LIME+SHAP & SHAP \\
\midrule
Anchor & 1.00 & 0.90 & 0.90 & 0.69 & 0.37 & 0.90 & 0.90 \\
Anchor+LIME & 0.90 & 1.00 & 0.56 & 0.90 &\textbf{0.043*}& 0.90 & 0.69 \\
Anchor+LIME+SHAP & 0.90 & 0.56 & 1.00 & 0.21 & 0.87 & 0.90 & 0.90 \\
Anchor+SHAP & 0.69 & 0.90 & 0.21 & 1.00 &\textbf{0.005**}& 0.87 & 0.31 \\
LIME & 0.37 &\textbf{0.043*}& 0.87 &\textbf{0.005**}& 1.00 & 0.21 & 0.75 \\
LIME+SHAP & 0.90 & 0.90 & 0.90 & 0.87 & 0.21 & 1.00 & 0.90 \\
SHAP & 0.90 & 0.69 & 0.90 & 0.31 & 0.75 & 0.90 & 1.00 \\
\bottomrule
\multicolumn{8}{l}{\scriptsize *p(<0.05), **p(<0.01), ***p(<0.001)} \\
\end{tabular}
\end{table}
\begin{table}[ht]
\centering
\caption{Nemenyi test results for fusion combinations for $\bar{{CONF}} \cdot \bar{{COV}} \cdot {ER}$. P-values for pairwise comparisons.}
\label{tab:nemenyi_rules_combination_metric_con_cov_xratio}
\tiny
\begin{tabular}{lrrrrrrr}
\toprule
\textbf{Rules Combination} & Anchor & Anchor+LIME & Anchor+LIME+SHAP & Anchor+SHAP & LIME & LIME+SHAP & SHAP \\
\midrule
Anchor & 1.00 & 0.87 & 0.06 & 0.75 &\textbf{0.004**}&\textbf{0.001**}&\textbf{0.001**} \\
Anchor+LIME & 0.87 & 1.00 & 0.63 & 0.90 & 0.17 & 0.08 & 0.08 \\
Anchor+LIME+SHAP & 0.06 & 0.63 & 1.00 & 0.75 & 0.90 & 0.90 & 0.90 \\
Anchor+SHAP & 0.75 & 0.90 & 0.75 & 1.00 & 0.25 & 0.13 & 0.13 \\
LIME &\textbf{0.004**}& 0.17 & 0.90 & 0.25 & 1.00 & 0.90 & 0.90 \\
LIME+SHAP &\textbf{0.001**}& 0.08 & 0.90 & 0.13 & 0.90 & 1.00 & 0.90 \\
SHAP &\textbf{0.001**}& 0.08 & 0.90 & 0.13 & 0.90 & 0.90 & 1.00 \\
\bottomrule
\multicolumn{8}{l}{\scriptsize *p(<0.05), **p(<0.01), ***p(<0.001)} \\
\end{tabular}
\end{table}
\begin{table}[ht]
\centering
\caption{Nemenyi test results for fusion combinations for $\bar{{CONF}}$. P-values for pairwise comparisons.}
\label{tab:nemenyi_rules_combination_avg_confidence}
\tiny
\begin{tabular}{lrrrrrrr}
\toprule
\textbf{Rules Combination} & Anchor & Anchor+LIME & Anchor+LIME+SHAP & Anchor+SHAP & LIME & LIME+SHAP & SHAP \\
\midrule
Anchor & 1.00 & 0.90 & 0.90 & 0.50 & 0.31 & 0.90 & 0.90 \\
Anchor+LIME & 0.90 & 1.00 & 0.90 & 0.90 &\textbf{0.043*}& 0.90 & 0.90 \\
Anchor+LIME+SHAP & 0.90 & 0.90 & 1.00 & 0.90 &\textbf{0.043*}& 0.90 & 0.90 \\
Anchor+SHAP & 0.50 & 0.90 & 0.90 & 1.00 &\textbf{0.001**}& 0.90 & 0.50 \\
LIME & 0.31 &\textbf{0.043*}&\textbf{0.043*}&\textbf{0.001**}& 1.00 &\textbf{0.043*}& 0.31 \\
LIME+SHAP & 0.90 & 0.90 & 0.90 & 0.90 &\textbf{0.043*}& 1.00 & 0.90 \\
SHAP & 0.90 & 0.90 & 0.90 & 0.50 & 0.31 & 0.90 & 1.00 \\
\bottomrule
\multicolumn{8}{l}{\scriptsize *p(<0.05), **p(<0.01), ***p(<0.001)} \\
\end{tabular}
\end{table}
\begin{table}[ht]
\centering
\caption{Nemenyi test results for fusion combinations for $\bar{{COV}}$. P-values for pairwise comparisons.}
\label{tab:nemenyi_rules_combination_avg_coverage}
\tiny
\begin{tabular}{lrrrrrrr}
\toprule
\textbf{Rules Combination} & Anchor & Anchor+LIME & Anchor+LIME+SHAP & Anchor+SHAP & LIME & LIME+SHAP & SHAP \\
\midrule
Anchor & 1.00 & 0.44 & 0.13 & 0.31 & 0.17 &\textbf{0.011*}& 0.13 \\
Anchor+LIME & 0.44 & 1.00 & 0.90 & 0.90 & 0.90 & 0.75 & 0.90 \\
Anchor+LIME+SHAP & 0.13 & 0.90 & 1.00 & 0.90 & 0.90 & 0.90 & 0.90 \\
Anchor+SHAP & 0.31 & 0.90 & 0.90 & 1.00 & 0.90 & 0.87 & 0.90 \\
LIME & 0.17 & 0.90 & 0.90 & 0.90 & 1.00 & 0.90 & 0.90 \\
LIME+SHAP &\textbf{0.011*}& 0.75 & 0.90 & 0.87 & 0.90 & 1.00 & 0.90 \\
SHAP & 0.13 & 0.90 & 0.90 & 0.90 & 0.90 & 0.90 & 1.00 \\
\bottomrule
\multicolumn{8}{l}{\scriptsize *p(<0.05), **p(<0.01), ***p(<0.001)} \\
\end{tabular}
\end{table}
\begin{table}[ht]
\centering
\caption{Nemenyi test results for fusion combinations for ${{ER}}$. P-values for pairwise comparisons.}
\label{tab:nemenyi_rules_combination_explained_ratio}
\tiny
\begin{tabular}{lrrrrrrr}
\toprule
\textbf{Rules Combination} & Anchor & Anchor+LIME & Anchor+LIME+SHAP & Anchor+SHAP & LIME & LIME+SHAP & SHAP \\
\midrule
Anchor & 1.00 & 0.90 & 0.50 & 0.90 & 0.31 & 0.90 & 0.63 \\
Anchor+LIME & 0.90 & 1.00 & 0.44 & 0.90 & 0.25 & 0.90 & 0.56 \\
Anchor+LIME+SHAP & 0.50 & 0.44 & 1.00 & 0.25 & 0.90 & 0.87 & 0.90 \\
Anchor+SHAP & 0.90 & 0.90 & 0.25 & 1.00 & 0.13 & 0.90 & 0.37 \\
LIME & 0.31 & 0.25 & 0.90 & 0.13 & 1.00 & 0.69 & 0.90 \\
LIME+SHAP & 0.90 & 0.90 & 0.87 & 0.90 & 0.69 & 1.00 & 0.90 \\
SHAP & 0.63 & 0.56 & 0.90 & 0.37 & 0.90 & 0.90 & 1.00 \\
\bottomrule
\multicolumn{8}{l}{\scriptsize *p(<0.05), **p(<0.01), ***p(<0.001)} \\
\end{tabular}
\end{table}
\begin{table}[ht]
\centering
\caption{Nemenyi test results for fusion combinations for $\bar{F(n)}$. P-values for pairwise comparisons.}
\label{tab:nemenyi_rules_combination_avg_features_count}
\tiny
\begin{tabular}{lrrrrrrr}
\toprule
\textbf{Rules Combination} & Anchor & Anchor+LIME & Anchor+LIME+SHAP & Anchor+SHAP & LIME & LIME+SHAP & SHAP \\
\midrule
Anchor & 1.00 &\textbf{0.023*}&\textbf{0.016*}& 0.56 &\textbf{0.016*}&\textbf{0.001**}& 0.50 \\
Anchor+LIME &\textbf{0.023*}& 1.00 & 0.90 & 0.75 & 0.90 & 0.75 & 0.81 \\
Anchor+LIME+SHAP &\textbf{0.016*}& 0.90 & 1.00 & 0.69 & 0.90 & 0.81 & 0.75 \\
Anchor+SHAP & 0.56 & 0.75 & 0.69 & 1.00 & 0.69 & 0.06 & 0.90 \\
LIME &\textbf{0.016*}& 0.90 & 0.90 & 0.69 & 1.00 & 0.81 & 0.75 \\
LIME+SHAP &\textbf{0.001**}& 0.75 & 0.81 & 0.06 & 0.81 & 1.00 & 0.08 \\
SHAP & 0.50 & 0.81 & 0.75 & 0.90 & 0.75 & 0.08 & 1.00 \\
\bottomrule
\multicolumn{8}{l}{\scriptsize *p(<0.05), **p(<0.01), ***p(<0.001)} \\
\end{tabular}
\end{table}
\begin{table}[ht]
\centering
\caption{Nemenyi test results for fusion combinations for ${Med}(F(n))$. P-values for pairwise comparisons.}
\label{tab:nemenyi_rules_combination_median_features_count}
\tiny
\begin{tabular}{lrrrrrrr}
\toprule
\textbf{Rules Combination} & Anchor & Anchor+LIME & Anchor+LIME+SHAP & Anchor+SHAP & LIME & LIME+SHAP & SHAP \\
\midrule
Anchor & 1.00 & 0.31 &\textbf{0.011*}& 0.78 &\textbf{0.050*}&\textbf{0.001**}& 0.44 \\
Anchor+LIME & 0.31 & 1.00 & 0.87 & 0.90 & 0.90 & 0.13 & 0.90 \\
Anchor+LIME+SHAP &\textbf{0.011*}& 0.87 & 1.00 & 0.40 & 0.90 & 0.81 & 0.75 \\
Anchor+SHAP & 0.78 & 0.90 & 0.40 & 1.00 & 0.69 &\textbf{0.014*}& 0.90 \\
LIME &\textbf{0.050*}& 0.90 & 0.90 & 0.69 & 1.00 & 0.53 & 0.90 \\
LIME+SHAP &\textbf{0.001**}& 0.13 & 0.81 &\textbf{0.014*}& 0.53 & 1.00 & 0.08 \\
SHAP & 0.44 & 0.90 & 0.75 & 0.90 & 0.90 & 0.08 & 1.00 \\
\bottomrule
\multicolumn{8}{l}{\scriptsize *p(<0.05), **p(<0.01), ***p(<0.001)} \\
\end{tabular}
\end{table}

\begin{table}[ht]
\centering
\caption{Nemenyi test results for fusion methods for $\bar{M}$. P-values for pairwise comparisons.}
\label{tab:nemenyi_method_final_metric}
\tiny
\begin{tabular}{lrrrrrrr}
\toprule
\textbf{Method} & Baseline & Best & Intersection & Lasso & Lasso global & Union & Weighted \\
\midrule
Baseline & 1.00 &\textbf{0.015*}& 0.56 & 0.49 & 0.56 & 0.90 & 0.90 \\
Best &\textbf{0.015*}& 1.00 &\textbf{0.001**}&\textbf{0.001**}&\textbf{0.001**}&\textbf{0.001**}&\textbf{0.018*} \\
Intersection & 0.56 &\textbf{0.001**}& 1.00 & 0.90 & 0.90 & 0.90 & 0.53 \\
Lasso & 0.49 &\textbf{0.001**}& 0.90 & 1.00 & 0.90 & 0.90 & 0.45 \\
Lasso global & 0.56 &\textbf{0.001**}& 0.90 & 0.90 & 1.00 & 0.90 & 0.53 \\
Union & 0.90 &\textbf{0.001**}& 0.90 & 0.90 & 0.90 & 1.00 & 0.90 \\
Weighted & 0.90 &\textbf{0.018*}& 0.53 & 0.45 & 0.53 & 0.90 & 1.00 \\
\bottomrule
\multicolumn{8}{l}{\scriptsize *p(<0.05), **p(<0.01), ***p(<0.001)} \\
\end{tabular}
\end{table}
\begin{table}[ht]
\centering
\caption{Nemenyi test results for fusion methods for $\bar{{CONF}} \cdot {ER}$. P-values for pairwise comparisons.}
\label{tab:nemenyi_method_metric_con_xratio}
\tiny
\begin{tabular}{lrrrrrrr}
\toprule
\textbf{Method} & Baseline & Best & Intersection & Lasso & Lasso global & Union & Weighted \\
\midrule
Baseline & 1.00 &\textbf{0.001**}& 0.21 &\textbf{0.001**}& 0.47 &\textbf{0.001**}& 0.80 \\
Best &\textbf{0.001**}& 1.00 &\textbf{0.003**}& 0.33 &\textbf{0.001**}& 0.90 &\textbf{0.001**} \\
Intersection & 0.21 &\textbf{0.003**}& 1.00 & 0.64 &\textbf{0.001**}&\textbf{0.001**}& 0.90 \\
Lasso &\textbf{0.001**}& 0.33 & 0.64 & 1.00 &\textbf{0.001**}&\textbf{0.020*}& 0.11 \\
Lasso global & 0.47 &\textbf{0.001**}&\textbf{0.001**}&\textbf{0.001**}& 1.00 &\textbf{0.001**}&\textbf{0.018*} \\
Union &\textbf{0.001**}& 0.90 &\textbf{0.001**}&\textbf{0.020*}&\textbf{0.001**}& 1.00 &\textbf{0.001**} \\
Weighted & 0.80 &\textbf{0.001**}& 0.90 & 0.11 &\textbf{0.018*}&\textbf{0.001**}& 1.00 \\
\bottomrule
\multicolumn{8}{l}{\scriptsize *p(<0.05), **p(<0.01), ***p(<0.001)} \\
\end{tabular}
\end{table}
\begin{table}[ht]
\centering
\caption{Nemenyi test results for fusion methods for $\bar{{CONF}} \cdot \bar{{COV}} \cdot {ER}$. P-values for pairwise comparisons.}
\label{tab:nemenyi_method_metric_con_cov_xratio}
\tiny
\begin{tabular}{lrrrrrrr}
\toprule
\textbf{Method} & Baseline & Best & Intersection & Lasso & Lasso global & Union & Weighted \\
\midrule
Baseline & 1.00 & 0.90 &\textbf{0.008**}& 0.18 & 0.37 & 0.10 & 0.64 \\
Best & 0.90 & 1.00 &\textbf{0.001**}&\textbf{0.013*}&\textbf{0.044*}&\textbf{0.006**}& 0.14 \\
Intersection &\textbf{0.008**}&\textbf{0.001**}& 1.00 & 0.90 & 0.75 & 0.90 & 0.49 \\
Lasso & 0.18 &\textbf{0.013*}& 0.90 & 1.00 & 0.90 & 0.90 & 0.90 \\
Lasso global & 0.37 &\textbf{0.044*}& 0.75 & 0.90 & 1.00 & 0.90 & 0.90 \\
Union & 0.10 &\textbf{0.006**}& 0.90 & 0.90 & 0.90 & 1.00 & 0.90 \\
Weighted & 0.64 & 0.14 & 0.49 & 0.90 & 0.90 & 0.90 & 1.00 \\
\bottomrule
\multicolumn{8}{l}{\scriptsize *p(<0.05), **p(<0.01), ***p(<0.001)} \\
\end{tabular}
\end{table}
\begin{table}[ht]
\centering
\caption{Nemenyi test results for fusion methods for $\bar{{CONF}}$. P-values for pairwise comparisons.}
\label{tab:nemenyi_method_avg_confidence}
\tiny
\begin{tabular}{lrrrrrrr}
\toprule
\textbf{Method} & Baseline & Best & Intersection & Lasso & Lasso global & Union & Weighted \\
\midrule
Baseline & 1.00 &\textbf{0.001**}& 0.14 &\textbf{0.016*}& 0.90 &\textbf{0.001**}& 0.90 \\
Best &\textbf{0.001**}& 1.00 & 0.14 & 0.54 &\textbf{0.001**}& 0.77 &\textbf{0.001**} \\
Intersection & 0.14 & 0.14 & 1.00 & 0.90 &\textbf{0.004**}&\textbf{0.001**}& 0.35 \\
Lasso &\textbf{0.016*}& 0.54 & 0.90 & 1.00 &\textbf{0.001**}&\textbf{0.022*}& 0.06 \\
Lasso global & 0.90 &\textbf{0.001**}&\textbf{0.004**}&\textbf{0.001**}& 1.00 &\textbf{0.001**}& 0.65 \\
Union &\textbf{0.001**}& 0.77 &\textbf{0.001**}&\textbf{0.022*}&\textbf{0.001**}& 1.00 &\textbf{0.001**} \\
Weighted & 0.90 &\textbf{0.001**}& 0.35 & 0.06 & 0.65 &\textbf{0.001**}& 1.00 \\
\bottomrule
\multicolumn{8}{l}{\scriptsize *p(<0.05), **p(<0.01), ***p(<0.001)} \\
\end{tabular}
\end{table}
\begin{table}[ht]
\centering
\caption{Nemenyi test results for fusion methods for $\bar{{COV}}$. P-values for pairwise comparisons.}
\label{tab:nemenyi_method_avg_coverage}
\tiny
\begin{tabular}{lrrrrrrr}
\toprule
\textbf{Method} & Baseline & Best & Intersection & Lasso & Lasso global & Union & Weighted \\
\midrule
Baseline & 1.00 &\textbf{0.001**}& 0.65 & 0.90 &\textbf{0.001**}& 0.23 & 0.90 \\
Best &\textbf{0.001**}& 1.00 &\textbf{0.001**}&\textbf{0.001**}&\textbf{0.001**}& 0.35 &\textbf{0.001**} \\
Intersection & 0.65 &\textbf{0.001**}& 1.00 & 0.75 &\textbf{0.008**}&\textbf{0.002**}& 0.90 \\
Lasso & 0.90 &\textbf{0.001**}& 0.75 & 1.00 &\textbf{0.001**}& 0.16 & 0.90 \\
Lasso global &\textbf{0.001**}&\textbf{0.001**}&\textbf{0.008**}&\textbf{0.001**}& 1.00 &\textbf{0.001**}&\textbf{0.001**} \\
Union & 0.23 & 0.35 &\textbf{0.002**}& 0.16 &\textbf{0.001**}& 1.00 &\textbf{0.016*} \\
Weighted & 0.90 &\textbf{0.001**}& 0.90 & 0.90 &\textbf{0.001**}&\textbf{0.016*}& 1.00 \\
\bottomrule
\multicolumn{8}{l}{\scriptsize *p(<0.05), **p(<0.01), ***p(<0.001)} \\
\end{tabular}
\end{table}
\begin{table}[ht]
\centering
\caption{Nemenyi test results for fusion methods for ${{ER}}$. P-values for pairwise comparisons.}
\label{tab:nemenyi_method_explained_ratio}
\tiny
\begin{tabular}{lrrrrrrr}
\toprule
\textbf{Method} & Baseline & Best & Intersection & Lasso & Lasso global & Union & Weighted \\
\midrule
Baseline & 1.00 &\textbf{0.001**}& 0.49 &\textbf{0.001**}& 0.69 &\textbf{0.001**}& 0.39 \\
Best &\textbf{0.001**}& 1.00 &\textbf{0.001**}& 0.13 &\textbf{0.001**}& 0.90 &\textbf{0.001**} \\
Intersection & 0.49 &\textbf{0.001**}& 1.00 & 0.29 &\textbf{0.011*}&\textbf{0.001**}& 0.90 \\
Lasso &\textbf{0.001**}& 0.13 & 0.29 & 1.00 &\textbf{0.001**}& 0.13 & 0.39 \\
Lasso global & 0.69 &\textbf{0.001**}&\textbf{0.011*}&\textbf{0.001**}& 1.00 &\textbf{0.001**}&\textbf{0.006**} \\
Union &\textbf{0.001**}& 0.90 &\textbf{0.001**}& 0.13 &\textbf{0.001**}& 1.00 &\textbf{0.001**} \\
Weighted & 0.39 &\textbf{0.001**}& 0.90 & 0.39 &\textbf{0.006**}&\textbf{0.001**}& 1.00 \\
\bottomrule
\multicolumn{8}{l}{\scriptsize *p(<0.05), **p(<0.01), ***p(<0.001)} \\
\end{tabular}
\end{table}
\begin{table}[ht]
\centering
\caption{Nemenyi test results for fusion methods for $\bar{F(n)}$. P-values for pairwise comparisons.}
\label{tab:nemenyi_method_avg_features_count}
\tiny
\begin{tabular}{lrrrrrrr}
\toprule
\textbf{Method} & Baseline & Best & Intersection & Lasso & Lasso global & Union & Weighted \\
\midrule
Baseline & 1.00 &\textbf{0.001**}& 0.10 & 0.90 & 0.10 &\textbf{0.001**}&\textbf{0.027*} \\
Best &\textbf{0.001**}& 1.00 &\textbf{0.001**}&\textbf{0.001**}&\textbf{0.001**}& 0.20 &\textbf{0.001**} \\
Intersection & 0.10 &\textbf{0.001**}& 1.00 & 0.11 &\textbf{0.001**}& 0.29 & 0.90 \\
Lasso & 0.90 &\textbf{0.001**}& 0.11 & 1.00 & 0.08 &\textbf{0.001**}&\textbf{0.033*} \\
Lasso global & 0.10 &\textbf{0.001**}&\textbf{0.001**}& 0.08 & 1.00 &\textbf{0.001**}&\textbf{0.001**} \\
Union &\textbf{0.001**}& 0.20 & 0.29 &\textbf{0.001**}&\textbf{0.001**}& 1.00 & 0.56 \\
Weighted &\textbf{0.027*}&\textbf{0.001**}& 0.90 &\textbf{0.033*}&\textbf{0.001**}& 0.56 & 1.00 \\
\bottomrule
\multicolumn{8}{l}{\scriptsize *p(<0.05), **p(<0.01), ***p(<0.001)} \\
\end{tabular}
\end{table}
\begin{table}[ht]
\centering
\caption{Nemenyi test results for fusion methods for ${Med}(F(n))$. P-values for pairwise comparisons.}
\label{tab:nemenyi_method_median_features_count}
\tiny
\begin{tabular}{lrrrrrrr}
\toprule
\textbf{Method} & Baseline & Best & Intersection & Lasso & Lasso global & Union & Weighted \\
\midrule
Baseline & 1.00 &\textbf{0.001**}& 0.05 & 0.49 & 0.43 &\textbf{0.001**}&\textbf{0.005**} \\
Best &\textbf{0.001**}& 1.00 &\textbf{0.001**}&\textbf{0.001**}&\textbf{0.001**}& 0.20 &\textbf{0.001**} \\
Intersection & 0.05 &\textbf{0.001**}& 1.00 & 0.90 &\textbf{0.001**}& 0.18 & 0.90 \\
Lasso & 0.49 &\textbf{0.001**}& 0.90 & 1.00 &\textbf{0.002**}&\textbf{0.009**}& 0.58 \\
Lasso global & 0.43 &\textbf{0.001**}&\textbf{0.001**}&\textbf{0.002**}& 1.00 &\textbf{0.001**}&\textbf{0.001**} \\
Union &\textbf{0.001**}& 0.20 & 0.18 &\textbf{0.009**}&\textbf{0.001**}& 1.00 & 0.56 \\
Weighted &\textbf{0.005**}&\textbf{0.001**}& 0.90 & 0.58 &\textbf{0.001**}& 0.56 & 1.00 \\
\bottomrule
\multicolumn{8}{l}{\scriptsize *p(<0.05), **p(<0.01), ***p(<0.001)} \\
\end{tabular}
\end{table}

\FloatBarrier

\end{document}